\documentclass[runningheads]{llncs}

\usepackage{eccv}

\usepackage{eccvabbrv}
\usepackage{algorithm}
\usepackage{algpseudocode}
\usepackage{graphicx}
\usepackage{booktabs}
\usepackage{multirow}
\usepackage{makecell}
\usepackage{tabularx}
\usepackage{caption}
\usepackage{enumitem}
\usepackage[accsupp]{axessibility}  
\usepackage{hyperref}

\usepackage{orcidlink}

\begin{document}

\title{Stochastic Optimal Control Sampling for Diffusion Inverse Problems} 

\titlerunning{SOCS for Diffusion Inverse Problems}


\author{Jie Zhang\inst{1} \and
Youmei Qiu\inst{1} \and
Hanling Tian\inst{1} \and
Jingyuan Zhang\inst{1} \and
Xiang Yin\inst{1} \and
Xiaolin Huang\inst{1,2}\thanks{Corresponding Author.}}

\renewcommand{\thefootnote}{\fnsymbol{footnote}}


\authorrunning{Zhang et al.}

\institute{School of Automation and Intelligent Sensing, Shanghai Jiao Tong University, Shanghai, China \and Shanghai Key Laboratory of Flexible Medical Robotics, Tongren Hospital, Institute of Medical Robotics,  Shanghai Jiao Tong University, Shanghai, China
\email{\textbraceleft zhangjie20000411,qiuyoumei,hanlingtian,tonyzhang666,yinxiang,\\xiaolinhuang\textbraceright @sjtu.edu.cn}}

\maketitle

\begin{abstract}
  Benefiting from the strong ability to capture data distributions, diffusion models have become powerful tools for solving image inverse problems. The key is to controllably steer the sampling trajectory toward the measurements while respecting the diffusion prior. In this work, we introduce Stochastic Optimal Control Sampling (SOCS), which models the denoising process as a dynamical system and injects control signals via SOC. Previous SOC-based approach addresses inverse problems by optimizing over the entire trajectory, which is computationally expensive. In contrast, we derive a closed-form control update and apply it at each sampling step, pulling the measurement-consistent clean prediction back onto the denoising flow. In SOCS, we can readily modulate the control strength to align with the diffusion model’s native capabilities and thereby enhance perceptual quality. Our method is compatible with a variety of linear stochastic differential equation backbones. Extensive experiments across a broad spectrum of image inverse tasks demonstrate that SOCS achieves accurate measurement-aligned reconstructions with improved visual fidelity and stronger quantitative performance. Code is available at \url{https://github.com/zjqwq01/SOCS-DIP}.
\end{abstract}

\section{Introduction}
\label{sec:intro} 

Image inverse problems \cite{geman1984stochastic,perona2002scale,freeman2002example} are pervasive in science and engineering, from image super-resolution to more challenging nonlinear degradations. The goal is to reconstruct the ground-truth image from degraded measurements. The central difficulty stems from a degradation operator that is many-to-one and often nonlinear, which renders the solution inherently non-unique. Effective inversion therefore requires strong priors that should be leveraged efficiently. From this perspective, diffusion models \cite{DDPM,EDM,karras2024analyzing,song2019generative,song2020improved,song2021scorebasedgenerativemodelingstochastic} are especially promising for inverse problems. They learn expressive generative priors that capture high-order statistics and geometric structure of natural images, and enable principled posterior sampling by conditioning score-based dynamics on measurements. As a result, diffusion models have become compelling foundations for inverse problems. 

\begin{figure}[t]  
  \centering
  \includegraphics[width=0.8\columnwidth,trim=0 12 0 0,clip]{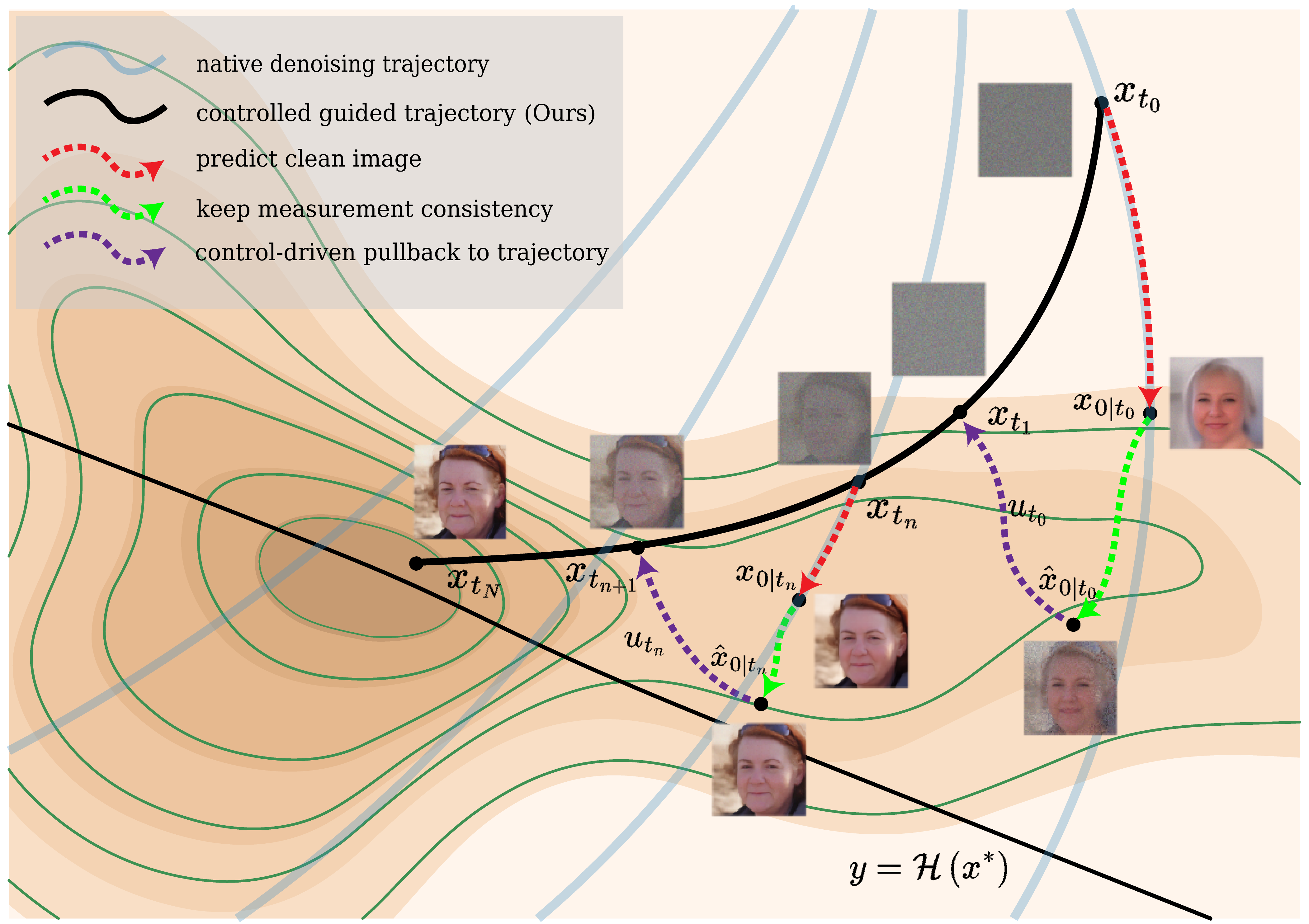}
  \caption{\textbf{Illustration of our method (an example on a random inpainting task).} We incorporate stochastic optimal control into sampling so that, at every step, the optimal control policy pulls the measurement-consistent clean estimate back onto the denoising trajectory, preserving fidelity while honoring the diffusion prior.}
  \label{fig:method_overview}
\end{figure}

To generate images that are both measurement consistent and remarkably faithful, one way is to retrain diffusion models for a specific inverse task \cite{saharia2022image,saharia2022palette,xia2023diffir}, which however requires massive data and incurs heavy computational cost. A more efficient way is 
to inject measurement information into the denoising process of a pretrained diffusion model. The current mainstream of such training-free methods lies within a Bayesian sampling framework, see, e.g., \cite{DPS,2GDM}. They approximate the conditional score by replacing it with a simple function of the noise-free data likelihood and use such approximation to steer denoising toward the measurements. Enforcing measurement consistency along the trajectory naturally calls for controlling the diffusion process to satisfy the measurements. To design such a control signal in this direction, \cite{berner2022optimal} established the foundational theoretical connection between diffusion processes and stochastic optimal control (SOC) and later \cite{solveIPviaDOC} applies SOC to diffusion inverse problems by designing a traditional iLQR controller \cite{iLQR,tassa2012synthesis}, which optimizes the entire diffusion path via iterative rollouts; consequently, it requires second-order Taylor updates, which demand Hessians and lead to high memory usage and long runtimes.

In this work, we introduce a new framework to apply SOC in diffusion inverse problems, termed \textbf{S}tochastic \textbf{O}ptimal \textbf{C}ontrol \textbf{S}ampling (SOCS). Instead of relying on iLQR-style trajectory controllers, we integrate SOC into the stepwise stochastic routine of the diffusion model, which avoids costly higher-order computations and enables the use of a pretrained unconditional diffusion prior without retraining. As shown in \cref{fig:method_overview}, at each iteration we first obtain a preliminary clean estimate by advancing the reverse-diffusion probability-flow ODE with a few Euler steps, second enforce consistency of this estimate with the measurement, and third map the estimate back to the denoising manifold through a closed-form transport. Our method rests on a natural intuition that an uncontrolled diffusion prior already produces high-quality samples, and excessive control will drive the process away from its native manifold, thereby undermining this ability. In light of this, we adopt controllers with bounded amplitude and provide a series of theoretical foundations to justify this design. Accordingly, SOCS not only penalizes the terminal state but also deliberately and gently modulates the amplitude of intermediate information injection, aligning the denoising process with the model’s intrinsic generative capacity. Moreover, our method requires at most first-order Jacobians, yielding wall-clock times comparable to sampling-based baselines while preserving the benefits of SOC guidance.

Empirically, our method delivers substantial gains over existing approaches across a broad spectrum of inverse problems and applies to all linear stochastic differential equation (SDE) formulations \cite{song2019generative}, including the classical variance-exploding SDE (VE-SDE)  and variance-preserving SDE (VP-SDE). Quantitative experiments show that tuning the strength of the injected control markedly improves generative performance. We further extend SOCS to various Latent Diffusion Models (LDMs), including traditional diffusion-based frameworks and state-of-the-art Flow-Matching (FM) variants, demonstrating its compatibility with high-resolution image restoration across different generative paradigms.

\section{Background} \label{sec:Background}

\subsection{Diffusion Models}
Diffusion models \cite{DDPM,EDM,song2019generative,song2020improved,song2021scorebasedgenerativemodelingstochastic} learn a generative prior by inverting a gradual noising process. Let $x_0\sim p_{\mathrm{data}}$ and choose a noise schedule ${\sigma_t}$ that increases from $\sigma_0=0$ to $\sigma_T=\sigma_{\max}$, so the forward marginals $p(x_t)\equiv p(x;\sigma_t)$ are Gaussian–smoothed versions of $p_{\mathrm{data}}$. In the SDE formulation, the forward dynamics can be written as $\mathrm{d}x_t=f_t(x_t)\mathrm{d}t+g_t\mathrm{d}w_t$. With $f_t$ the drift and $g_t$ the diffusion scale (e.g., VP or VE choices), the reverse-time SDE relies on the score $\nabla_x \log p(x_t;\sigma_t)$ to correct the trajectory back to the data distribution. Generation transports $x_T\sim\mathcal{N}(0,\sigma_T^2 I)$ back to data by integrating the reverse dynamics, where the score is approximated by a network $s_\theta(x_t,\sigma_t)\approx\nabla_x\log p(x_t;\sigma_t)$. An equivalent training view predicts the injected noise under $x_t=\sqrt{\bar{\alpha}_t}x_0+\sqrt{1-\bar{\alpha}_t}\varepsilon$, where $\bar{\alpha}_t$ is the cumulative decay factor and $\varepsilon\sim\mathcal N(0,I)$, minimizing $\mathbb{E}_{x_0,\varepsilon,t}\left[w(t)\|\varepsilon-\varepsilon_\theta(x_t,t)\|^2\right]$, which yields $s_\theta(x_t,t)\approx-\varepsilon_\theta(x_t,t)/\sqrt{1-\bar{\alpha}_t}$. Alternative output formulations that predict clean image $x_0$ or the noise $\epsilon$ are algebraically equivalent and map to the same reverse process via simple reparameterizations that are invertible.

\subsection{Bayesian Inverse Problems with Diffusion}
Inverse problems seek to recover a clean signal from degraded measurements. Let the degradation process be modeled as $y = \mathcal{H}(x_0) + n$, where $\mathcal{H}$ is the degradation operator, $x_0$ is the unknown clean data, $y$ is the measurement, and $n \sim \mathcal{N}(0,\beta_y^2 I)$ is additive Gaussian noise. In a Bayesian formulation that treats uncertainty explicitly, the likelihood is $p(y \mid x_0) = \mathcal{N}(\mathcal{H}(x_0), \beta_y^2 I)$, and solving the inverse problem amounts to sampling from the posterior $p(x_0 \mid y)$.

Given access to the measurement $y$, by Bayes’ rule the conditional score can be written as:
$$
\nabla_{x}\log p_t(x_t\mid y)\;=\;\nabla_{x}\log p_t(x_t)\;+\;\nabla_{x}\log p_t(y\mid x_t).
$$
Here, the term $\nabla_{x}\log p_t(x_t)$ is readily available from a pretrained diffusion model, whereas the noise-likelihood term $\nabla_{x}\log p_t(y\mid x_t)$ is generally intractable. A variety of methods \cite{boys2023tweedie,RED-Diff,DAPS,DPS,DiffPIR} have been proposed to estimate or circumvent this term. DPS \cite{DPS} approximates the likelihood via the posterior mean of the clean signal, replacing $p(y\mid x_t)$ with $p\big(y\mid \hat{x}_0\big)$, where $\hat{x}_0 := \mathbb{E}[x_0\mid x_t]$.  Building on this perspective, \cite{RED-Diff} argues that the recursive nature of denoising induces a complex, multi-modal posterior that is difficult to approximate with simple forms, and therefore proposes a variational-inference approach to infer the posterior of the data given observations. DAPS \cite{DAPS} mitigates the step-size dependence of conventional denoising by decoupling the continuous process, which enables later steps to effectively correct errors made earlier. In practice, sampling methods developed under the Bayesian framework commonly employ a guidance term of the form $\nabla_{x_t}\lVert \mathcal{H}(x_t)-y\rVert_2^2$. Without approximating the noise likelihood term, we jointly model this term together with the model prior as a single control signal and modulate it in a unified way. This yields a guidance behavior similar to those approaches and, from the SOC perspective, establishes the necessity of this term.

\subsection{Diffusion with Stochastic Optimal Control}
Integrating stochastic optimal control (SOC) into diffusion-model sampling has emerged as a promising guidance paradigm. DIS \cite{berner2022optimal} first established a foundational link between diffusion processes and SOC, motivating practical formulations. Building on this theory, RB-Modulation \cite{rout2024rbmodulationtrainingfreepersonalizationdiffusion} applies SOC to diffusion-style personalization, framing it as an optimal control problem. Relevant to our setting, UniDB \cite{UniDB} formulates the forward noising of diffusion bridges within an SOC framework on image restoration tasks and introduces a simple controller linked to Doob’s \emph{h}-transform that shows strong detail preservation. However, it requires retraining whenever terminal penalties or other settings change. \cite{solveIPviaDOC} solves inverse problems by formulating reverse diffusion as an optimal-control problem and derives a practical controller that improves stability and data consistency across operators, although computing and storing Hessian matrices impose substantial costs. To address these challenges, we model the denoising process to obviate retraining and integrate SOC at each step, thereby avoiding higher-order computations and substantially reducing runtime.

\section{Method} \label{sec:method}
\subsection{Inverse Problems under an SOC Formulation}
Inverse problems aim to recover the original signal from noisy measurements $y$. With a similar goal, SOCS formulates a stochastic optimal control (SOC) problem whose constraint is a reverse-diffusion linear SDE initialized from random noise, with drift $f(x_t,t)=f_t x_t$, where $f_t$ is some scalar-valued function. To align with standard SOC modeling, we let $t=0$ denote the pure-noise initialization and $t=T$ denote the final clean-image time. The objective steers the reverse denoising trajectory toward a prescribed terminal state by minimizing the squared Euclidean residual $\lVert \mathcal{H}(x_T)-y\rVert_2^2$.

Specifically, our SOC problem with $\mathcal{H}$ under linear SDE can be written as:
\begin{equation}
\begin{aligned}
\min_{\mathbf{u}_{t,\gamma}\in \mathcal{U}}\;&\mathbb{E}\left[ \int_0^T \tfrac{1}{2}\bigl\lVert\mathbf{u}_{t,\gamma}\bigr\lVert_2^2\,\mathrm dt
+ \tfrac{\gamma}{2}\bigl\lVert y-\mathcal{H}(x_T)\bigr\lVert_2^2\right] \\
\text{s.t.}\quad & \mathrm d x_t = (f_t x_t + g_t \mathbf{u}_{t,\gamma})\,\mathrm dt + g_t\,\mathrm d w_t.
\end{aligned}
\label{eq:SOC_construct_SDE}
\end{equation}

\noindent\textbf{Interpretation (prior proximity via control energy).}
By Girsanov's theorem, the control energy upper-bounds the KL deviation of the controlled terminal distribution from the diffusion prior:
\(
D_{\mathrm{KL}}(Q_T\|P_T)\le \frac12\,\mathbb E\!\left[\int_0^T\|u_{t,\gamma}\|_2^2\,\mathrm dt\right].
\)
A detailed proof can be found in Appendix~\ref{app:bounded_energy_prior_proximity}.

\noindent\textbf{Interpretation (posterior sampling).}
The SOC objective in \cref{eq:SOC_construct_SDE} induces a Gibbs reweighting of the diffusion prior on path space,
\(
\frac{\mathrm d Q^*_{\mathrm{path}}}{\mathrm d P_{\mathrm{path}}}\propto \exp\!\big(-\phi(x_T)\big),
\ \phi(x_T)=\frac{\gamma}{2}\|y-\mathcal H(x_T)\|_2^2,
\)
hence \(q_T^*(x)\propto p_T(x)\exp(-\phi(x))\). Under Gaussian noise, \(\phi(x)\) is proportional to the negative log-likelihood, so \(q_T\) admits a posterior-sampling interpretation. More details are provided in Appendix~\ref{app:socs_prior_posterior}.

\noindent\textbf{Theorem 3.1.} Under linear dynamics, additive zero-mean control-independent noise, and a quadratic terminal cost (or a Gauss–Newton linearization of $\mathcal{H}$), the certainty equivalence principle \cite{chen2024generativemodelingphasestochastic,rout2024rbmodulationtrainingfreepersonalizationdiffusion} implies that the optimal controller remains the same. Thus, we obtain the following SOC problem with deterministic ODE condition:
\begin{equation}
\begin{aligned}
\min_{\mathbf{u}_{t,\gamma}\in \mathcal{U}}\;& \int_0^T \tfrac{1}{2}\bigl\lVert\mathbf{u}_{t,\gamma}\bigr\lVert_2^2\,\mathrm dt
+ \tfrac{\gamma}{2}\bigl\lVert y-\mathcal{H}(x_T)\bigr\lVert_2^2 \\
\text{s.t.}\quad & \mathrm d x_t = (f_t x_t + g_t \mathbf{u}_{t,\gamma})\,\mathrm dt.
\end{aligned}
\label{eq:SOC_construct_ODE}
\end{equation}

A key observation is that we can derive a closed-form solution for the SOC problem Eq.~\eqref{eq:SOC_construct_ODE}. Denote $d_{0,\gamma}^{-1} := \bigl(I + \gamma\,e^{2\bar f_T}\,\bar g_T^{\,2}\, J_H(x_T^u)^\top \circ\mathcal{H}\bigr)^{-1},\bar f_{s:t}:= \int_s^t f_z\mathrm dz$ and $\bar g^{2}_{s:t}:= \int_s^t e^{-2\bar f_z}g_z^{2}\mathrm dz$, 
we simplify notation to $\bar f_t$ and $\bar g_t^{2}$ for $\bar f_{0:t}$ and $\bar g^{2}_{0:t}$. Here $J_H(x_T^u)$ is the Jacobian of $\mathcal{H}(x)$ evaluated at $x_T^u$, and the symbol $\circ$ denotes operator composition.
The optimal controller $\mathbf{u}_{t,\gamma}^*$ could be calculated:
\begin{equation}
\mathbf{u}_{t,\gamma}^*=-g_t\gamma e^{\bar{f}_{t:T}} d_{0,\gamma}^{-1} J_H(x_T^u)^\top\bigl(e^{\bar{f}_T}\mathcal{H}(x_0) - y\bigr).
\label{eq:SOC_u_t}
\end{equation}
Therefore, transition of $x_t$ from $x_0$ is:
\begin{equation}
\begin{aligned}
x_t &= e^{\bar{f}_t}x_0 + \gamma e^{\bar{f}_t} e^{\bar{f}_T} \bar{g}_t^2d_{0,\gamma}^{-1} J_H(x_T^u)^\top \bigl(y-e^{\bar{f}_T}\mathcal{H}(x_0)\bigr).
\end{aligned}
\label{eq:SOC_x_t}
\end{equation}

The detailed derivations of Eq.~\eqref{eq:SOC_u_t} and Eq.~\eqref{eq:SOC_x_t} are provided in Appendix \ref{sec:Proof of Theorem 3.1}. The above discussions hold for all linear SDEs, covering both the VE–SDE and VP–SDE. 
Existing Bayesian sampling methods \cite{DPS,DAPS} inherently assume an infinite terminal penalty with $\gamma$ driven to infinity. In practice, treating $\gamma$ as a tunable hyperparameter can be more effective. In the next section, using the VP–SDE as a running case, we examine how different approximations to $\mathcal{H}$ and the choice of the terminal penalty affect Eq.~\eqref{eq:SOC_x_t} and present two main variants of the method.

\subsection{Case Study: SOCS-VP-SDE}
\label{sec:A Case: SOCS-VP-SDE}
In this section, we apply our posterior-sampling pipeline to VP–SDE. In the VP–SDE, the forward noising dynamics are $\mathrm{d}x_t = -\tfrac{1}{2}\beta(t)x_t\mathrm{d}t + \sqrt{\beta(t)}\mathrm{d}w_t$, where the drift term $-\tfrac{1}{2}\beta(t)x_t$ linearly contracts the state toward the origin, and the diffusion term $\sqrt{\beta(t)}$ injects Gaussian noise under a time-varying schedule. We take $\beta(t)=g(t)^2$.

\noindent\textbf{SOCS-VP-Nonlinear-$\gamma$.} Eq.~\eqref{eq:SOC_x_t} makes clear that $x_t$ consists of a baseline term $e^{\bar f_t}x_0$ together with an affine data-consistency correction. The latter is formed by back-projecting the measurement residual from measurement space to image space via the Jacobian adjoint $J_{\mathcal H}(x_T^u)^{\top}$, and modulating it with a time-varying gain. Using $\bar g_t^{\,2}=e^{-2\bar f_t}-1$, Eq.~\eqref{eq:SOC_x_t} can be equivalently recast as:
\begin{equation}
\begin{aligned}
x_t
&= e^{\bar{f}_t}x_0
+ \gamma e^{\bar{f}_T}\bigl(e^{-\bar{f}_t}-e^{\bar{f}_t}\bigr)\hat{x}_T, \\
\hat{x}_T
&\triangleq
\Bigl(I + \gamma(1-e^{2\bar{f}_T})\,J_\mathcal{H}(x_T^u)^\top\circ \mathcal{H}\Bigr)^{-1}
\circ J_\mathcal{H}(x_T^u)^\top\bigl(y-e^{\bar{f}_T}\mathcal{H}(x_0)\bigr).
\end{aligned}
\label{eq:VP-SDE-nonlinear-gamma}
\end{equation}
In practice, the Gauss–Newton inverse of the regularized normal matrix defined in Eq.~\eqref{eq:VP-SDE-nonlinear-gamma} is typically challenging to handle. We use $\hat{x}_T$ to denote the Gauss–Newton update , and compute it via an implicit solve instead of an explicit matrix inverse. Equivalently, we solve $(I + \gamma(1-e^{2\bar{f}_T})J_\mathcal{H}(x_T^u)^\top\circ\mathcal{H})(\hat{x}_T)=J_\mathcal{H}(x_T^u)^\top(y-e^{\bar{f}_T}\mathcal{H}(x_0))$. Leveraging Langevin dynamics \cite{welling2011bayesian} sampling, we refine \(\hat x_T\) via a stochastic-gradient step on the corresponding data-consistency energy:
\begin{equation}
\begin{aligned}
\hat{x}_T^{j+1} =& \hat{x}_T^{j} - \eta \nabla_{\hat{x}_T^{j}} \big\lVert\big(I + \gamma(1-e^{2\bar{f}_T}) J_\mathcal{H}(x_T^u)^\top\big)\circ\mathcal{H}(\hat{x}_T^{j})\\
&- J_\mathcal{H}(x_T^u)^\top\big(y-e^{\bar{f}_T}\mathcal{H}(x_0)\big)\big\rVert^2 + \sqrt{2\eta} \epsilon_j,
\end{aligned}
\label{eq:MCMC-sampling-gamma}
\end{equation}
where \(\eta>0\) is the step size and $\epsilon_j\sim\mathcal{N}(0,I)$. Unlike previous methods \cite{DAPS,MCG,DPS-MO,DPS}, which motivate the update from a Bayesian posterior estimation and invoke the gradient of the quadratic data-fidelity loss, here an analogous quadratic guidance term arises naturally from the SOC formulation. Since we do not approximate the degradation operator and explicitly retain the terminal penalty coefficient $\gamma$, we refer to this sampling scheme as SOCS-VP-Nonlinear-$\gamma$.

If we disregard the running cost in the SOC formulation by taking $\gamma \to \infty$, then, since $\lim_{\gamma \to \infty} \gamma d_{0,\gamma}^{-1} = \frac{1}{e^{2\bar{f}_T}\bar{g}_T^2}\bigl(J_\mathcal{H}(x_T^u)^\top\circ\mathcal{H}\bigr)^{-1}$. Eq.~\eqref{eq:SOC_x_t} simplifies:
\begin{equation}
x_t = e^{\bar{f}_t}\bigl(1-\dfrac{\bar{g}_t^2}{\bar{g}_T^2}\bigr)x_0 + \dfrac{e^{\bar{f}_t}\bar{g}_t^2}{e^{\bar{f}_T}\bar{g}_T^2}\bigl(J_\mathcal{H}(x_T^u)^\top\circ\mathcal{H}\bigr)^{-1}\circ J_\mathcal{H}(x_T^u)^\top (y).
\label{eq:VP-SDE_nonlinear}
\end{equation}
Following the SOCS-VP-Nonlinear-$\gamma$ spirit, we set $\hat{x}_T=\bigl(J_\mathcal{H}(x_T^u)^{\top}\circ \mathcal{H}\bigr)^{-1}\circ J_\mathcal{H}(x_T^u)^{\top}(y)$. This amounts to solving \(J_\mathcal{H}(x_T^u)^{\top}\circ \mathcal{H}(\hat{x}_T)=J_\mathcal{H}(x_T^u)^{\top} (y)\), we enforce \(\mathcal{H}(\hat{x}_T)=y\) as a sufficient measurement-consistency condition that satisfies the preceding equation. We then refine $\hat{x}_T$ using Langevin dynamics:
\begin{equation}
\hat{x}_T^{j+1} = \hat{x}_T^{j} - \eta \nabla_{\hat{x}_T^{j}} \bigl\lVert\mathcal{H}(\hat{x}_T^{j}) - y\bigr\rVert^2 + \sqrt{2\eta} \epsilon_j.
\label{eq:MCMC-sampling}
\end{equation}
According to \cite{DAPS}, as $\eta\to 0$ and $j\to\infty$, the iterate $\hat{x}_T^{j}$ approximately follows $p(\hat{x}_T\mid x_t,y)$. We view this optimization as a special instance of Eq.~\eqref{eq:MCMC-sampling-gamma}, which we denote as SOCS-VP-Nonlinear-$\infty$.

\noindent \textbf{SOCS-VP-Linear.} Furthermore, if we take the regime $\gamma \to \infty$ and adopt a linear approximation of the degradation operator, Eq.~\eqref{eq:SOC_x_t} reduces to:
\begin{equation}
x_t = e^{\bar{f}_t}\bigl(1-\dfrac{\bar{g}_t^2}{\bar{g}_T^2}\bigr)x_0 + \dfrac{e^{\bar{f}_t}\bar{g}_t^2}{e^{\bar{f}_T}\bar{g}_T^2}\underbrace{\bigl(\mathcal{H}^\top\mathcal{H}\bigr)^{-1}\mathcal{H}^\top}_{\mathcal{H}^{\dagger}(\cdot)}(y),
\label{eq:VP-SDE_linear}
\end{equation} 
where the coefficients in front of $x_0$ and $\mathcal{H}^{\dagger}(y)$ are time-dependent constants that can be precomputed offline. Because $\mathcal{H}^{\dagger}(y)$ assumes exact access to the pseudoinverse, which is rarely available in practice, we adopt a pragmatic two-step scheme. Within the EDM framework, we first produce a coarse prediction $x_T(x_t)$ using a few-step Euler ODE solver, however, this prediction may not fully align with the low-quality input in the degenerate subspace. Following \cite{DDRM,DDNM}, we enforce measurement consistency by projecting onto the affine subspace $\{x\in\mathbb{R}^n : \mathcal{H}(x)=y\}$:
\begin{equation}
\hat{x}_T(x_t)=\bigl(I-\mathcal{H}^{\dagger}\mathcal{H}\bigr)x_T(x_t) + \mathcal{H}^{\dagger}(y).
\label{eq:force_consistency}
\end{equation}
Details for $\mathcal{H}(\cdot)$ and $\mathcal{H}^{\dagger}(\cdot)$ are given in the Appendix~\ref{sec:inverse_problem_setup}. 

In all cases above, once we obtain the projected estimate $\hat{x}_T(x_t)$ from Eq.~\eqref{eq:force_consistency} in the linear setting, or the MCMC-refined estimate from Eq.~\eqref{eq:MCMC-sampling-gamma} and Eq.~\eqref{eq:MCMC-sampling} in the nonlinear or finite-$\gamma$ setting, we apply SOC to pull the resulting $\hat{x}_T$ back onto the denoising manifold at time $t$ by substituting it into the corresponding closed-form expression for $x_t$. This continues the reverse-time sampling with measurement consistency and SOC guidance. 
Under a locally linear and well-conditioned observation model, \emph{SOCS-Linear} regime yielding a stable closed-form affine projection onto $\{x:\mathcal H(x)=y\}$. The \emph{SOCS-Nonlinear} regime retains the full $J_{\mathcal H}(\cdot)$ with a finite $\gamma$, preserving the geometry of the prior manifold under strong degeneracy or model mismatch and enabling multi step likelihood-guided refinement. 
As emphasized in \cite{DAPS}, to enhance robustness, we decouple consecutive states $x_{t+\Delta t}$ and $x_t$ along the entire sampling trajectory so that the sampler can correct global errors that arise early in denoising, rather than restricting the distance. Algorithm~\ref{algo:SOCS_algorithmic} provides the detailed procedure.
\begin{algorithm}[t]
\caption{SOCS~(Stochastic Optimal Control Sampling)}\label{algo:SOCS_algorithmic}
\begin{algorithmic}[1] 
\Require Score model $s_\theta$, measurement $\mathbf y$, operator $\mathcal H$, time grid $(t_i)_{i=0}^{N}$
\State Sample $\mathbf x_{t_0} \sim \mathcal N(\mathbf 0, I)$
\For{$i = 0$ \textbf{to} $N-1$}
  \State \begingroup
    \thinmuskip=2mu \medmuskip=2mu plus 1mu minus 1mu
    \thickmuskip=3mu plus 1mu minus 1mu
    $\mathbf{x}_{T\mid t_i}\!\gets\!\mathbf{x}^{(T)}_{t_i}\!\bigl(\mathbf{x}_{t_i}\bigr)$
    \Comment{few step ODE sampling with $s_\theta$}
    \endgroup
  \If{$\mathrm{mode} = \text{Linear}$}
     \State $\hat{\mathbf x}_{T|t_i} \gets (I-\mathcal H^{\dagger}\mathcal H)\,\mathbf x_{T|t_i} + \mathcal H^{\dagger}\mathbf y$   \Comment{Eq.~\eqref{eq:force_consistency}}
  \ElsIf{$\mathrm{mode} = \text{Nonlinear-}\gamma$}
     \For{$j = 0$ \textbf{to} $N_{\text{inner}}-1$}
           \State $\hat{\mathbf x}_{T|t_i}^{(j+1)} \gets \textsc{Refine}(\hat{\mathbf x}_{T|t_i}^{(j)})$   \Comment{\cref{eq:MCMC-sampling} for $\gamma\to\infty$/\cref{eq:MCMC-sampling-gamma} for general cases.}
     \EndFor
     \State $\hat{\mathbf x}_{T|t_i} \gets \hat{\mathbf x}_{T|t_i}^{(N_{\text{inner}})}$
  \EndIf
  \State Project $\hat{\mathbf x}_{T|t_i}$ to SOC manifold, get $\mathbf x_{t_{i+1}}$     \Comment{Eq.~\eqref{eq:SOC_x_t}}
\EndFor
\State \Return $\mathbf x_N \gets \hat{\mathbf x}_N$  
\end{algorithmic}
\end{algorithm}

As shown in \cref{tab:sampling_time}, the computational overhead of the Langevin refinements is minor relative to diffusion model evaluations, since the dominant cost typically stems from the neural network rather than the measurement operator $\mathcal{H}$. For completeness, we also present the linear-SDE case instantiated as a VE--SDE in Appendix~\ref{sec:SOCS-VE-SDE}. Moreover, we extend SOCS to latent diffusion models (LDMs) and flow-matching models (represented by SD3), details are provided in Appendix~\ref{app:SOCS with LDMs} and Appendix~\ref{app:socs_fm}, respectively.
\begin{figure*}[!t]
  \centering
  \includegraphics[width=\textwidth]{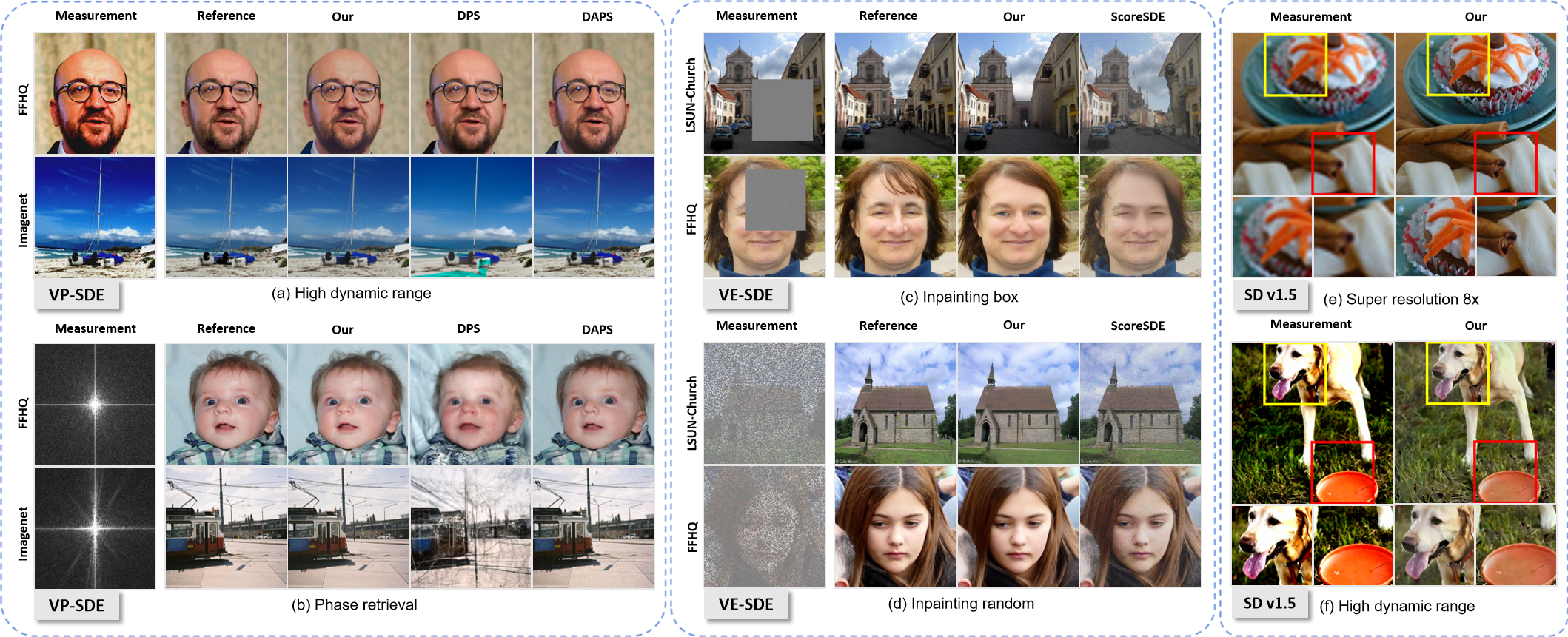}
  \caption{\textbf{Representative samples of SOCS}. Our method leverages stochastic optimal control (SOC) to provide an efficient, theoretically grounded framework for diffusion inverse problems. In (a)(b), we present VP-SDE results on FFHQ and ImageNet-256, respectively; in (c)(d), we showcase VE-SDE results on LSUN Church and FFHQ; in (e)(f), we show natural images at a resolution of 512.}
  \label{fig:first_page}
\end{figure*}

\subsection{Comparison with posterior sampling methods}
\begin{figure}[t]
\centering
\begin{minipage}[t]{0.60\linewidth}
  \centering
  \includegraphics[width=\linewidth]{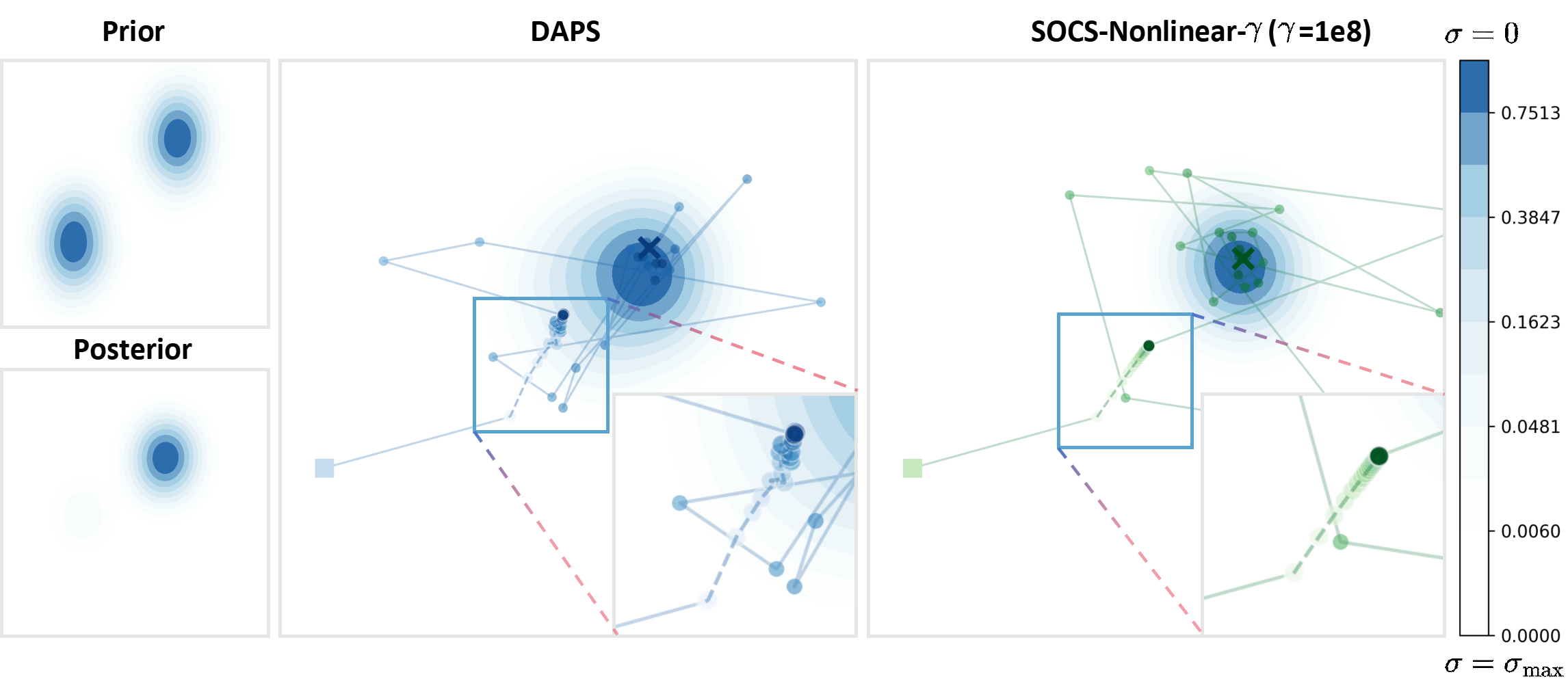}
  \captionof{figure}{SOCS vs DAPS on 2D synthetic data. SOCS exhibits a more direct refinement trajectory.}
  \label{fig:SOCS_toy_example_sec3}
\end{minipage}\hfill
\begin{minipage}[t]{0.38\linewidth}
  \centering
  \raisebox{-2mm}{\includegraphics[width=0.98\linewidth]{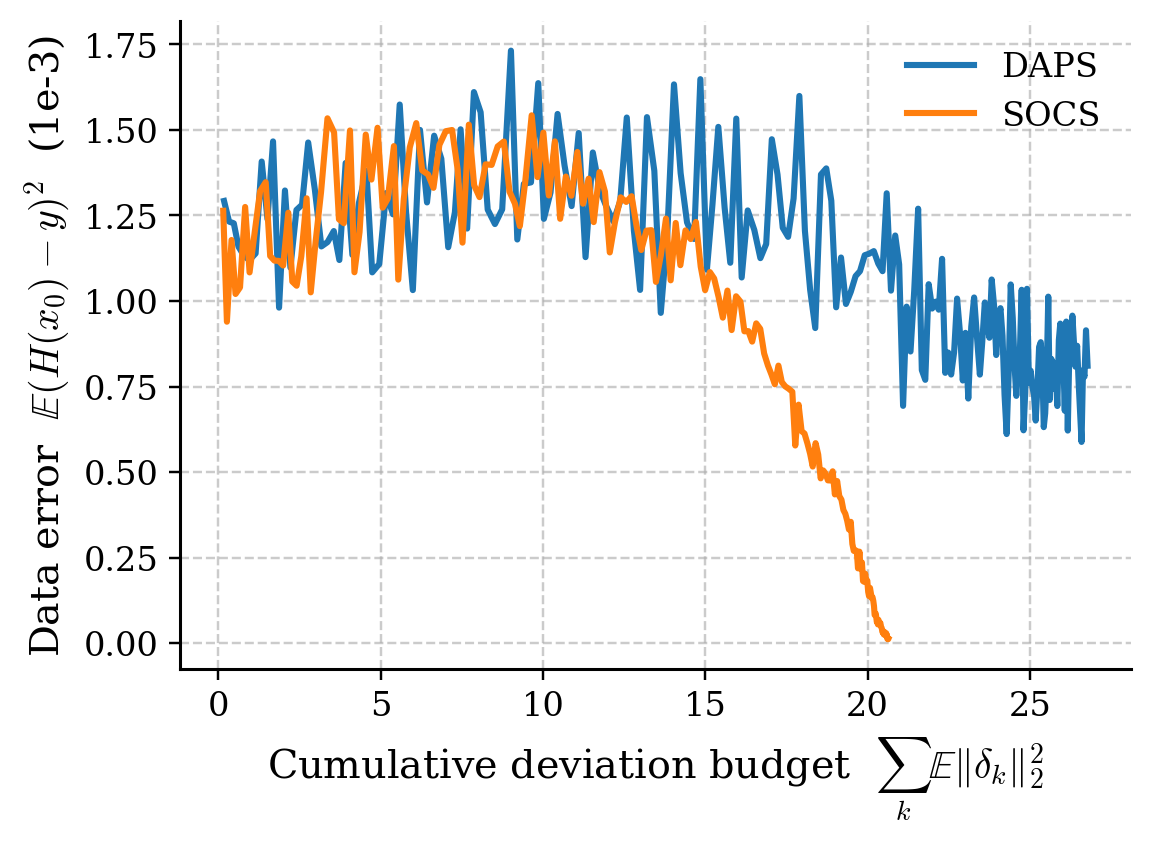}}
  \setlength{\abovecaptionskip}{0pt}
  \captionof{figure}{Data error versus cumulative deviation budget.}
  \label{fig:control_pareto_sec3}
\end{minipage}
\end{figure}

Existing posterior sampling methods enforce data consistency by steering the reverse denoising process with measurement guidance.
This includes approximating $p_t(y\mid x_t)$ with an isotropic Gaussian \cite{DPS,2GDM}, projection-based corrections \cite{DPS,MCG}, and subsequent optimization refinements \cite{PnP-DM,DiffPIR,DAPS,DPS-MO,peng2024improving}.
Although implemented differently, these approaches ultimately rely on the same core term $\nabla_{x_t}\lVert \mathcal{H}(x_t)-y\rVert_2^2$, which injects observational information into prior-driven dynamics. We provide the first stochastic optimal control justification of this term. In SOCS-Nonlinear, the standard gradient guidance emerges as a special case in the limit $\gamma\to\infty$, whereas for finite $\gamma$ our framework yields an optimal feedback that minimizes control energy under a terminal measurement penalty. Both SOCS and DAPS \cite{DAPS} benefit from trajectory decoupling, Concretely, at each outer step $k$, they first predict a clean image $x_{T,k}$ from the current state $x_k$, then run gradient-based Langevin refinement to improve data consistency and obtain $\hat{x}_{T,k}$, and finally project back to the denoising trajectory to produce $x_{k+1}$. 
\cref{fig:SOCS_toy_example_sec3} compares SOCS and DAPS on a two-dimensional nonlinear inverse problem with a Gaussian mixture prior. We visualize the sampling trajectories $\{x_t\}_{t\in[0,T]}$ for both methods, and within the first step we additionally show the inner refinement path $x_{T,0}\!\to\!\hat{x}_{T,0}$, where intermediate Langevin iterates are connected by dashed segments. 
SOCS exhibits a more direct refinement trajectory. 
In \cref{fig:control_pareto_sec3}, we further quantify the budget--consistency trade-off using $\delta_k$, which characterizes the optimization trajectory from $x_{T,k} \to \hat{x}_{T,k}$. 
Benefiting from the optimal control objective, SOCS consistently attains lower data error than DAPS under a smaller cumulative deviation budget $\sum_k \mathbb{E}\|\delta_k\|^2_2$, which cannot be achieved by Langevin refinement driven solely by $\nabla_{x_t}\lVert \mathcal{H}(x_t)-y\rVert_2^2$. 
We provide additional discussions in Appendix~\cref{sec:Toy_example}.

\section{Experiments}
\subsection{Experimental Setup}
We evaluate our three sampling variants using the FFHQ model pretrained by \cite{DPS} and the ImageNet pretrained model provided by \cite{dhariwal2021diffusionmodelsbeatgans}. As discussed in \cref{sec:method}, we adopt the EDM time discretization and noise schedule. For each time $t$, we estimate $x_T(x_t)$ with a few-step Euler ODE solver.

For all linear tasks, we use 4 ODE-solver NFEs and 250 reverse-denoising steps and for nonlinear tasks, we use 8 ODE-solver NFEs and 500 reverse-denoising steps. Detailed settings for SOCS-Linear, SOCS-Nonlinear, and SOCS-Nonlinear-$\gamma$, including model configurations, samplers, denoising scheduler and other hyperparameters are provided in the Appendix~\ref{sec:Experimental_details}. We also report an ablation on the terminal-penalty coefficient $\gamma$ to investigate its impact on reconstruction quality, along with other ablation studies.

\textbf{Datasets and metrics.} Following convention, we evaluate on FFHQ $256\times256$ \cite{FFHQ} and ImageNet $256\times256$ \cite{ImageNet}. For assessment, we use 100 images from the validation split of each dataset. We include Peak Signal-to-Noise Ratio (PSNR), Structural Similarity Index (SSIM), Learned Perceptual Image Patch Similarity (LPIPS)\cite{LPIPS} and Fréchet Inception Distance (FID)\cite{FID} as evaluation metrics.

\textbf{Inverse problems.} We evaluate our method on a suite of linear and nonlinear tasks. For linear inverse problems, we include: (i) $\times 4$ bicubic super-resolution; (ii) Gaussian deblur with a $61\times 61$ kernel and standard deviation 3.0; (iii) motion blur generated using $61\times 61$ kernels with blur strength $0.5$; (iv) inpainting, where box inpainting masks a random $128\times 128$ square and random inpainting masks $70\%$ of pixels; (v) super-resolution + inpainting, which first applies a random $30\%-70\%$ pixel mask followed by $\times 4$ bicubic downsampling.

We further consider three nonlinear inverse problems: (i) phase retrieval, where we apply the Fourier transform to each image and keep only the magnitude as the measurement. Given the task’s inherent difficulty, following prior work \cite{DPS,DAPS}, we use an oversampling rate of $2.0$ which is a standard choice to mitigate the ill-posedness and report the best result over four independent runs; (ii) High dynamic range (HDR) reconstruction, which aims to recover a higher–dynamic-range image from a low–dynamic–range input, with the expansion factor set to 2; (iii) For nonlinear deblurring, we follow the default configuration described in \cite{nonlin_blur_setting} under the same protocol and parameter settings.

For both linear and nonlinear tasks, we report the mean performance (PSNR, SSIM and LPIPS) across 100 validation images. The FID scores are evaluated with the same 100 validation images. To conform with the noisy inverse-problem setting, we corrupt all measurements, both linear and nonlinear, with additive white Gaussian noise of standard deviation $\beta_y = 0.05$. For experiments based on SD-v1.5, we instead use $\beta_y=0.01$, while for SD3 we set $\beta_y=0.03$. Complete task specifications and hyperparameter are provided in the Appendix~\ref{sec:inverse_problem_setup}.
\begin{table}[htbp]
\centering
\caption{\textbf{Quantitative evaluation on FFHQ and ImageNet across 6 linear tasks.} We compare methods on FFHQ (left) and ImageNet (right). The best and second-best results are indicated by bold and underlined marks, respectively.}
\label{tab:quant_eval_linear}
\renewcommand{\arraystretch}{0.87}
\resizebox{\textwidth}{!}{%
\begin{tabular}{@{}l |l|cccc|cccc@{}}
\toprule
\multicolumn{1}{c|}{\multirow{2}{*}{Task}} &
\multicolumn{1}{c|}{\multirow{2}{*}{Method}} &
\multicolumn{4}{c|}{FFHQ (256$\times$256)} &
\multicolumn{4}{c}{ImageNet (256$\times$256)} \\
\cmidrule(lr){3-6}\cmidrule(l){7-10}
& & PSNR ($\uparrow$) & SSIM ($\uparrow$) & LPIPS ($\downarrow$) & FID ($\downarrow$)
  & PSNR ($\uparrow$) & SSIM ($\uparrow$) & LPIPS ($\downarrow$) & FID ($\downarrow$) \\
\midrule
\multirow{12}{*}{\makecell[l]{Super \\Resolution 4$\times$}}
  & SOCS-Linear         &28.51 &0.797 &0.165 &\bf{72.99} &\bf{26.42} &\bf{0.733} &0.340 &\bf{93.77} \\
  & SOCS-Nonlinear      &28.39 &0.772 &\underline{0.150} &74.82 &26.10 &0.672 &\bf{0.277} &102.51 \\
  & SOCS-Nonlinear-$\gamma$ &\underline{28.66} &0.800 &0.166 &76.86 &\underline{26.36} &\underline{0.708} &\underline{0.283} &\underline{97.54} \\
    \cmidrule(lr){2-10}
  & DAPS\cite{DAPS}           &28.31 &0.768 &\bf{0.149} &74.75 &25.95 &0.667 &0.335 &112.60 \\
  & DPS\cite{DPS}             &24.25 &0.677 &0.190 &84.33 &22.56 &0.574 &0.387 &155.01 \\
  & DDRM\cite{DDRM}           &27.53 &0.796 &0.180 &92.19 &24.90 &0.673 &0.308 &183.50 \\
  & DDNM\cite{DDNM}           &27.93 &\underline{0.808} &0.170 &91.15 &25.02 &0.694 &0.306 &158.35 \\
  & DCDP\cite{DCDP}           &26.06 &0.629 &0.322 &118.09 &23.83 &0.548 &0.511 &162.58 \\
  & FPS\,-\,SMC\cite{FPS-SMC} &\bf{29.02} &\bf{0.828} &0.234 &106.56 &24.64 &0.671 &0.340 &174.98 \\
  & DiffPIR\cite{DiffPIR}     &26.48 &0.733 &0.272 &\underline{74.62} &24.04 &0.616 &0.308 &97.72 \\
  & RED-diff\cite{RED-Diff}   &25.46 &0.591 &0.391 &130.71 &23.12 &0.504 &0.546 &163.79 \\
  & DOC\cite{solveIPviaDOC}   &27.15 &0.790 &0.173 &90.96 &- &- &- &- \\
\midrule
\multirow{10}{*}{Inpaint (Box)}
  & SOCS-Linear           &24.22 &0.794 &\underline{0.109} &53.21 &20.05 &0.741 &0.263 &160.89 \\
  & SOCS-Nonlinear        &24.27 &0.742 &0.135 &55.26 &\underline{20.40} &0.702 &0.248 &117.96 \\
  & SOCS-Nonlinear-$\gamma$   &24.45 &0.826 &\bf{0.092} &\bf{43.79} &\bf{20.51} &\bf{0.772} &\bf{0.209} &\bf{113.67} \\
    \cmidrule(lr){2-10}
  & DAPS                      &24.41 &0.753 &0.117 &51.51 &20.05 &0.702 &0.246 &130.05 \\
  & DPS                       &23.49 &0.804 &0.134 &62.18 &18.97 &0.717 &0.269 &124.78 \\
  & DDRM                      &22.18 &0.818 &0.126 &62.87 &18.72 &0.739 &0.235 &125.26 \\
  & DDNM                      &24.14 &\bf{0.837} &0.117 &48.96 &18.46 &\underline{0.752} &0.238 &\underline{117.25} \\
  & DCDP                      &\underline{24.49} &0.776 &0.206 &47.17 &17.44 &0.656 &0.261 &167.07 \\
  & FPS\,-\,SMC               &\bf{24.70} &\underline{0.828} &0.126 &\underline{45.29} &19.83 &0.743 &\underline{0.209} &125.73 \\
\midrule
\multirow{10}{*}{Inpaint (Random)}
  & SOCS-Linear           &29.31 &0.834 &0.105 &77.09 &28.44 &0.790 &0.175 &91.92 \\
  & SOCS-Nonlinear        &\underline{30.13} &0.820 &0.093 &55.42 &28.05 &0.754 &\bf{0.102} &55.44 \\
  & SOCS-Nonlinear-$\gamma$   &\bf{30.63} &\bf{0.872} &\bf{0.065} &\bf{52.73} &\underline{28.75} &\underline{0.808} &\underline{0.108} &\bf{53.02} \\
    \cmidrule(lr){2-10}
  & DAPS                      &29.66 &0.800 &\underline{0.090} &\underline{54.57} &\bf{28.76} &0.789 &0.145 &\underline{54.41} \\
  & DPS                       &28.30 &0.812 &0.132 &71.74 &26.59 &0.736 &0.254 &98.27 \\
  & DDRM                      &25.80 &0.774 &0.166 &103.58 &22.63 &0.742 &0.271 &219.72 \\
  & DDNM                      &29.17 &\underline{0.850} &0.092 &64.65 &26.55 &\bf{0.868} &0.131 &99.36 \\
  & DCDP                      &30.12 &0.835 &0.105 &62.09 &26.41 &0.731 &0.167 &68.22 \\
  & FPS\,-\,SMC               &27.68 &0.824 &0.213 &104.22 &26.50 &0.707 &0.233 &68.96 \\
  & DOC                       &27.70 &0.795 &0.132 &100.51 &- &- &- &- \\
\midrule
\multirow{5}{*}{Inpaint\&SR}
  & SOCS-Linear           &27.21 &\underline{0.762} &0.195 &86.36 &\underline{25.53} &\underline{0.697} &0.384 &142.82 \\
  & SOCS-Nonlinear        &26.75 &0.714 &0.191 &94.79 &24.65 &0.569 &0.346 &134.14 \\
  & SOCS-Nonlinear-$\gamma$   &\bf{27.29} &\bf{0.780} &\underline{0.180} &\bf{82.30} &\bf{25.73} &\bf{0.708} &\underline{0.344} &\underline{123.85} \\
    \cmidrule(lr){2-10}
  & DAPS                      &\underline{27.22} &0.757 &\bf{0.166} &\underline{82.51} &25.48 &0.686 &\bf{0.309} &\bf{118.82} \\
  & DPS                       &23.00 &0.640 &0.213 &87.87 &21.45 &0.533 &0.425 &166.25 \\
\midrule
\multirow{12}{*}{\makecell[l]{Gaussian \\Deblurring}}
  & SOCS-Linear           &27.21 &0.785 &0.190 &98.13 &25.38 &\underline{0.720} &\bf{0.275} &111.49 \\
  & SOCS-Nonlinear        &\underline{28.52} &0.788 &0.179 &76.12 &26.32 &0.708 &0.313 &\bf{106.00} \\
  & SOCS-Nonlinear-$\gamma$   &\bf{28.79} &\bf{0.821} &\bf{0.162} &\bf{70.36} &\underline{26.67} &\bf{0.740} &\underline{0.282} &110.69 \\
    \cmidrule(lr){2-10}
  & DAPS                      &28.46 &\underline{0.809} &\underline{0.169} &73.54 &26.37 &0.697 &0.297 &109.61 \\
  & DPS                       &25.30 &0.706 &0.215 &74.87 &22.84 &0.587 &0.312 &115.02 \\
  & DDRM                      &27.68 &0.756 &0.224 &95.25 &\bf{27.81} &0.639 &0.333 &124.55 \\
  & DDNM                      &28.25 &0.801 &0.220 &72.23 &23.82 &0.657 &0.323 &211.84 \\
  & DCDP                      &27.73 &0.784 &0.218 &\underline{71.74} &25.12 &0.690 &0.429 &142.46 \\
  & FPS\,-\,SMC               &27.05 &0.770 &0.284 &122.98 &23.09 &0.588 &0.417 &220.59 \\
  & DiffPIR                   &25.42 &0.714 &0.345 &119.64 &23.25 &0.646 &0.340 &195.75 \\
  & RED-diff                  &26.34 &0.592 &0.240 &72.09 &24.24 &0.589 &0.292 &\underline{108.00} \\
\midrule
\multirow{9}{*}{\makecell[l]{Motion \\Deblurring}}
  & SOCS-Linear           &26.53 &0.724 &0.196 &93.62 &25.38 &0.720 &0.275 &111.48 \\
  & SOCS-Nonlinear        &\underline{29.74} &0.790 &0.128 &70.28 &\underline{28.36} &0.749 &\underline{0.203} &\underline{72.83} \\
  & SOCS-Nonlinear-$\gamma$   &\bf{30.39} &\underline{0.828} &\bf{0.104} &72.32 &\bf{28.91} &\bf{0.773} &0.204 &90.51 \\
    \cmidrule(lr){2-10}
  & DAPS                      &29.49 &\bf{0.829} &\underline{0.110} &\bf{55.66} &27.76 &\underline{0.765} &\bf{0.182} &\bf{64.79} \\
  & DPS                       &26.48 &0.750 &0.137 &70.14 &25.12 &0.682 &0.272 &100.31 \\
  & DCDP                      &25.85 &0.736 &0.174 &87.73 &23.71 &0.648 &0.414 &177.72 \\
  & FPS\,-\,SMC               &28.66 &0.820 &0.139 &\underline{67.92} &26.66 &0.680 &0.339 &87.92 \\
  & DiffPIR                   &25.77 &0.725 &0.356 &131.46 &24.70 &0.678 &0.369 &102.73 \\
  & DOC                       &25.87 &0.759 &0.199 &92.76 &- &- &- &- \\
\bottomrule
\end{tabular}%
}
\end{table}

\begin{table}[htbp]
\centering
\caption{\textbf{Quantitative evaluation on FFHQ and ImageNet across 3 nonlinear tasks.} We compare methods on FFHQ (left) and ImageNet (right). The best and second-best results are indicated by bold and underlined marks, respectively.}
\label{tab:quant_eval_nonlinear}
\renewcommand{\arraystretch}{0.87}
\resizebox{\textwidth}{!}{%
\begin{tabular}{@{}l| l|cccc|cccc@{}}
\toprule
\multicolumn{1}{c|}{\multirow{2}{*}{Task}} &
\multicolumn{1}{c|}{\multirow{2}{*}{Method}} &
\multicolumn{4}{c|}{FFHQ (256$\times$256)} &
\multicolumn{4}{c}{ImageNet (256$\times$256)} \\
\cmidrule(lr){3-6}\cmidrule(l){7-10}
& & PSNR ($\uparrow$) & SSIM ($\uparrow$) & LPIPS ($\downarrow$) & FID ($\downarrow$)
  & PSNR ($\uparrow$) & SSIM ($\uparrow$) & LPIPS ($\downarrow$) & FID ($\downarrow$) \\
\midrule
\multirow{7}{*}{Phase Retrieval}
  & SOCS-Linear           &- &- &- &- &- &- &- &- \\
  & SOCS-Nonlinear        &$28.73_{\pm 3.01}$ &$0.759_{\pm 0.073}$ &$0.119_{\pm 0.070}$ &52.43 &$24.74_{\pm 6.46}$ &$\underline{0.710}_{\pm 0.212}$ &$0.263_{\pm 0.200}$ &96.70 \\
  & SOCS-Nonlinear-$\gamma$   &$\underline{29.59}_{\pm 2.88}$ &$\textbf{0.808}_{\pm 0.050}$ &$\underline{0.072}_{\pm 0.029}$ &$\textbf{44.86}$ &$\underline{24.90}_{\pm 6.58}$ &$\textbf{0.716}_{\pm 0.221}$ &$\underline{0.256}_{\pm 0.207}$ &$\textbf{89.42}$ \\
  \cmidrule(lr){2-10}
  & DAPS                      &$\textbf{29.88}_{\pm 2.38}$ &$\underline{0.807}_{\pm 0.041}$ &$\textbf{0.069}_{\pm 0.030}$ &$\underline{45.33}$ &$\textbf{25.05}_{\pm 7.11}$ &$0.687_{\pm 0.230}$ &$\textbf{0.253}_{\pm 0.214}$ &\underline{95.92} \\
  & DPS                       &$16.29_{\pm 5.45}$ &$0.442_{\pm 0.192}$ &$0.388_{\pm 0.180}$ &137.81 &$15.89_{\pm 3.86}$ &$0.445_{\pm 0.159}$ &$0.413_{\pm 0.127}$ &201.93 \\
  & DCDP                      &$28.06_{\pm 8.64}$ &$0.754_{\pm 0.351}$ &$0.257_{\pm 0.272}$ &69.56 &$23.05_{\pm 6.54}$ &$0.624_{\pm 0.257}$ &$0.290_{\pm 0.232}$ &119.47 \\
  & RED-diff                  &$16.83_{\pm 7.84}$ &$0.521_{\pm 0.202}$ &$0.579_{\pm 0.200}$ &209.75 &$14.36_{\pm 5.38}$ &$0.398_{\pm 0.129}$ &$0.579_{\pm 0.129}$ &220.99 \\
\midrule
\multirow{7}{*}{Nonlinear Deblur}
  & SOCS-Linear           &$21.73_{\pm 2.36}$ &$0.647_{\pm 0.008}$ &$0.286_{\pm 0.081}$ &113.90 &$20.70_{\pm 3.06}$ &$0.564_{\pm 0.150}$ &$0.479_{\pm 0.150}$ &193.74 \\
  & SOCS-Nonlinear        &$28.29_{\pm 1.58}$ &$0.769_{\pm 0.035}$ &$0.169_{\pm 0.051}$ &73.36 &$26.11_{\pm 2.85}$ &$0.704_{\pm 0.080}$ &$0.262_{\pm 0.102}$ &111.88 \\
  & SOCS-Nonlinear-$\gamma$   &$\underline{28.42}_{\pm 1.63}$ &$\textbf{0.786}_{\pm 0.037}$ &$0.157_{\pm 0.050}$ &68.20 &$\underline{26.81}_{\pm 2.51}$ &$\textbf{0.740}_{\pm 0.069}$ &$\textbf{0.201}_{\pm 0.106}$ &80.51 \\
    \cmidrule(lr){2-10}
  & DAPS                      &$28.10_{\pm 1.62}$ &$0.762_{\pm 0.030}$ &$\underline{0.153}_{\pm 0.033}$ &69.45 &$\textbf{27.06}_{\pm 2.81}$ &$\underline{0.733}_{\pm 0.066}$ &$\underline{0.205}_{\pm 0.088}$ &83.81 \\
  & DPS                       &$23.05_{\pm 2.24}$ &$0.637_{\pm 0.087}$ &$0.226_{\pm 0.076}$ &90.69 &$21.98_{\pm 2.62}$ &$0.551_{\pm 0.144}$ &$0.443_{\pm 0.149}$ &207.94 \\
  & DCDP                      &$27.44_{\pm 2.45}$ &$0.770_{\pm 0.068}$ &$0.186_{\pm 0.054}$ &\textbf{58.41} &$26.07_{\pm 0.15}$ &$0.620_{\pm 0.113}$ &$0.216_{\pm 0.113}$ &\textbf{53.71} \\
  & RED-diff                  &$\textbf{29.16}_{\pm 0.08}$ &$\underline{0.783}_{\pm 0.051}$ &$\textbf{0.146}_{\pm 0.055}$ &\underline{63.22} &$25.08_{\pm 0.68}$ &$0.584_{\pm 0.130}$ &$0.244_{\pm 0.046}$ &\underline{55.75} \\
\midrule
\multirow{6}{*}{\makecell[l]{High Dynamic\\Range}}
  & SOCS-Linear           &$25.96_{\pm 1.84}$ &$0.767_{\pm 0.058}$ &$\underline{0.112}_{\pm 0.033}$ &63.87 &$22.51_{\pm 3.67}$ &$0.744_{\pm 0.109}$ &$0.179_{\pm 0.082}$ &72.35 \\
  & SOCS-Nonlinear        &$\underline{26.30}_{\pm 3.24}$ &$\underline{0.800}_{\pm 0.079}$ &$0.118_{\pm 0.061}$ &$\underline{61.54}$ &$\underline{24.24}_{\pm 4.29}$ &$\underline{0.757}_{\pm 0.117}$ &$\underline{0.152}_{\pm 0.094}$ &$\textbf{60.72}$ \\
  & SOCS-Nonlinear-$\gamma$   &$24.31_{\pm 3.11}$ &$\textbf{0.825}_{\pm 0.088}$ &$\textbf{0.108}_{\pm 0.073}$ &69.05 &$21.88_{\pm 4.51}$ &$\textbf{0.768}_{\pm 0.121}$ &$\textbf{0.148}_{\pm 0.088}$ &87.64 \\
    \cmidrule(lr){2-10}
  & DAPS                      &$\textbf{26.92}_{\pm 3.35}$ &$0.792_{\pm 0.084}$ &$0.145_{\pm 0.066}$ &$\textbf{55.00}$ &$\textbf{25.21}_{\pm 4.05}$ &$0.714_{\pm 0.116}$ &$0.167_{\pm 0.096}$ &\underline{62.37} \\
  & DPS                       &$24.53_{\pm 5.39}$ &$0.660_{\pm 0.180}$ &$0.270_{\pm 0.021}$ &79.48 &$19.83_{\pm 6.46}$ &$0.603_{\pm 0.280}$ &$0.538_{\pm 0.248}$ &179.90 \\
  & RED-diff                  &$21.26_{\pm 2.55}$ &$0.617_{\pm 0.080}$ &$0.269_{\pm 0.081}$ &97.15 &$22.85_{\pm 3.31}$ &$0.602_{\pm 0.128}$ &$0.279_{\pm 0.138}$ &87.19 \\
\bottomrule
\end{tabular}%
}
\end{table}

\textbf{Baselines.} For VP-SDE, we compare against DPS \cite{DPS}, DDRM \cite{DDRM}, DDNM \cite{DDNM}, DCDP \cite{DCDP}, FPS-SMC \cite{FPS-SMC}, DiffPIR \cite{DiffPIR}, DAPS \cite{DAPS} and DOC \cite{solveIPviaDOC} which also use SOC. Note that DDRM, DDNM, and FPS-SMC are limited to linear inverse problems and do not handle nonlinear cases. We especially include RED-diff \cite{RED-Diff} for nonlinear experiments. For VE–SDE, we benchmark against MCG \cite{MCG} and ScoreSDE \cite{song2021scorebasedgenerativemodelingstochastic}, focusing on the inpainting tasks. In addition, for SD v1.5 based latent diffusion experiments, we report results for LatentDAPS \cite{DAPS} and PSLD \cite{rout2023solving}. For flow-matching (FM) models, we include PSLD, LatentDAPS, FlowChef \cite{Flowchef}, and FlowDPS \cite{Flowdps} as additional baselines.
\begin{figure}[!t]
  \centering
  \begin{subfigure}{0.45\textwidth}
    \centering
    \includegraphics[width=\linewidth]{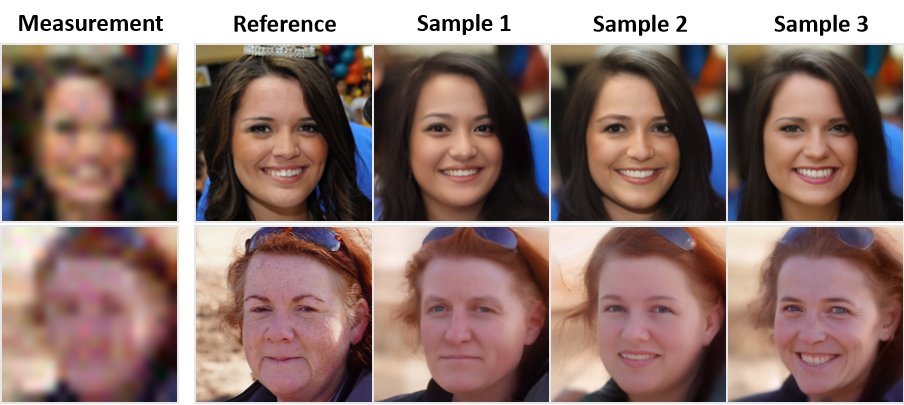}
    \caption{Super resolution by a factor 16} 
    \label{fig:sr}
  \end{subfigure}
  \begin{subfigure}{0.45\textwidth}
    \centering
    \includegraphics[width=\linewidth]{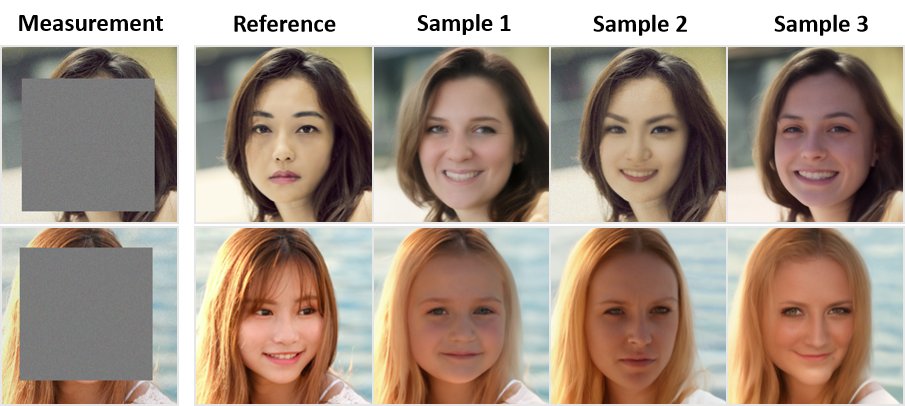}
    \caption{$192\times192$ box inpainting} 
    \label{fig:inpaint}
  \end{subfigure}
    \caption{\textbf{Sample diversity}. We present several diverse samples generated by the SOCS under two sparse measurements. SOCS produces a variety of samples with distinct features, including differences in expression, wearings, and hairstyles.}
  \label{fig:diversity}
\end{figure}
\subsection{Main Results} \label{sec:main_results}
We report quantitative results for the linear experiments on the FFHQ and ImageNet datasets in \cref{tab:quant_eval_linear}, and for the nonlinear experiments in \cref{tab:quant_eval_nonlinear}. Together with the qualitative comparisons in \cref{fig:first_page}, our method achieves comparable or superior performance across all tasks.

The observed preference between the linear and nonlinear variants follows the conditioning and local linearity of the observation model. When the forward map behaves linearly and remains well-conditioned, the linear variant approximates $J_{\mathcal H}\approx\mathcal H$ with $\gamma\to\infty$ and performs a stable affine projection onto the measurement-consistent set, which cooperates with the diffusion prior and avoids unnecessary iterative refinement. In the presence of strong degeneracy, large null spaces or spatially varying operators, or noticeable model mismatch, the nonlinear variant retains $J_{\mathcal H}(\cdot)$ with a finite $\gamma$ and conducts multi step Langevin guidance that respects the geometry of prior manifold and better resolves ambiguity without introducing projection artifacts.

Nonlinear forward models further accentuate this preference. Nonlinear deblurring involves state-dependent point spread functions that invalidate a global linear approximation, and high dynamic range formation includes saturation and tone-curve effects that break linearity. SOCS-Nonlinear addresses these cases via Jacobian-aware refinement and controlled guidance, enforcing data consistency while maintaining trajectory stability. Phase retrieval is magnitude based and discards phase, so no bounded linear pseudoinverse $\mathcal H^{\dagger}$ exists to map amplitude-only measurements back to the image domain. This makes SOCS-Linear inapplicable, whereas SOCS-Nonlinear remains effective through likelihood-driven updates. Empirically, as conditioning worsens, SOCS-Nonlinear consistently surpasses SOCS-Linear despite higher cost. We therefore recommend SOCS-Linear for super-resolution and SOCS-Nonlinear for the remaining tasks.

Moreover, our method produces a wider range of plausible samples when the measurements contain less information. For demonstration, we use a larger downsampling factor of 16 for the super resolution task and a larger 192×192 mask for the box inpainting task. The examples in \cref{fig:diversity} show that our method generates diverse samples while remaining faithful to the measurements.

\noindent\textbf{Further results on LDMs.} To demonstrate the broad versatile applicability of SOCS across various Latent Diffusion Model (LDM) frameworks, we evaluate our method on both the traditional epsilon-prediction based SD~v1.5 \cite{rombach2022high} and the advanced Flow-Matching (FM) based SD3. 
By operating within the lightweight VAE \cite{VAE} latent space, these models facilitate high-fidelity image synthesis under complex prompt guidance. 
For SD~v1.5, we implement the method described in \cref{sec:method} to validate SOCS on multiple inverse problems. 
Representative samples on FFHQ $256\times256$ for three linear and one non-linear task are shown in \cref{fig:SD1.5}, with $512\times512$ results presented in \cref{fig:first_page}. 
Furthermore, we extend SOCS to FM-based models like SD3-medium, which leverage stable ODE-based sampling trajectories and offer superior scalability for large-scale generative modeling.
Under an idealized straight-line interpretation, a flow-matching trajectory can be approximated as a linear ODE with near-constant velocity, enabling a seamless integration of SOCS. 
Quantitative results in \cref{tab:ffhq_768_sd3} for two $\times 12$ super-resolution settings ($768\times768$) demonstrate that our method achieves state-of-the-art performance across most metrics. 
Visual results for SOCS-SD3 are provided in \cref{fig:Denoising_processes_SD3}. 
Detailed prompts, theoretical derivations, and implementation specifics for both LDM paradigms are provided in Appendix~\ref{app:SOCS with LDMs} and \ref{app:socs_fm}, respectively.
\begin{table}[t]
\centering
\caption{\textbf{Quantitative evaluation on FFHQ at $768\times768$ on two super-resolution tasks with SD3-medium.} The best and second-best results within each type of task are indicated by bold and underlined marks, respectively.}
\label{tab:ffhq_768_sd3}
\renewcommand{\arraystretch}{1.05}
\setlength{\tabcolsep}{4pt}
\small
\resizebox{0.9\linewidth}{!}{%
\begin{tabular}{l|cccc|cccc}
\toprule
& \multicolumn{8}{c}{FFHQ($768\times768$)} \\
\cmidrule(lr){2-9}
\multirow{2}{*}{Method}
& \multicolumn{4}{c|}{Super-resolution $\times12$ (Avgpool)}
& \multicolumn{4}{c}{Super-resolution $\times12$ (Bicubic)} \\
\cmidrule(lr){2-5}\cmidrule(lr){6-9}
& PSNR$\uparrow$ & SSIM$\uparrow$ & LPIPS$\downarrow$ & FID$\downarrow$ 
& PSNR$\uparrow$ & SSIM$\uparrow$ & LPIPS$\downarrow$ & FID$\downarrow$   \\
\midrule
PSLD\cite{rout2023solving}
& 25.58 & 0.547 & 0.481 & 148.16  
& 25.60 & 0.572 & 0.517 & 154.78   \\
LatentDAPS\cite{DAPS}
& 28.94 & 0.823 & \underline{0.296} & 74.43 
& 29.65 & 0.819 & \underline{0.276} & 65.41  \\
FlowChef\cite{Flowchef}
& 29.21 & \textbf{0.852} & 0.341 & 153.37 
& 29.56 & \textbf{0.846} & 0.346 & 152.00 \\
FlowDPS\cite{Flowdps}
& \underline{29.41} & 0.777 & 0.482 & \underline{65.00} 
& \underline{29.73} & 0.775 & 0.482 & \underline{61.24}  \\
SOCS-SD3(Ours)
& \textbf{30.14} & \underline{0.827} & \textbf{0.259} & \textbf{53.61} 
& \textbf{30.45} & \underline{0.842} & \textbf{0.248} & \textbf{45.72}  \\
\bottomrule
\end{tabular}%
}
\end{table}

\subsection{Additional Discussion}
\label{sec:Additional_Discussion}
\begin{figure*}[!t]
  \centering
  \includegraphics[width=0.90\textwidth,trim=0 18 0 0,clip]{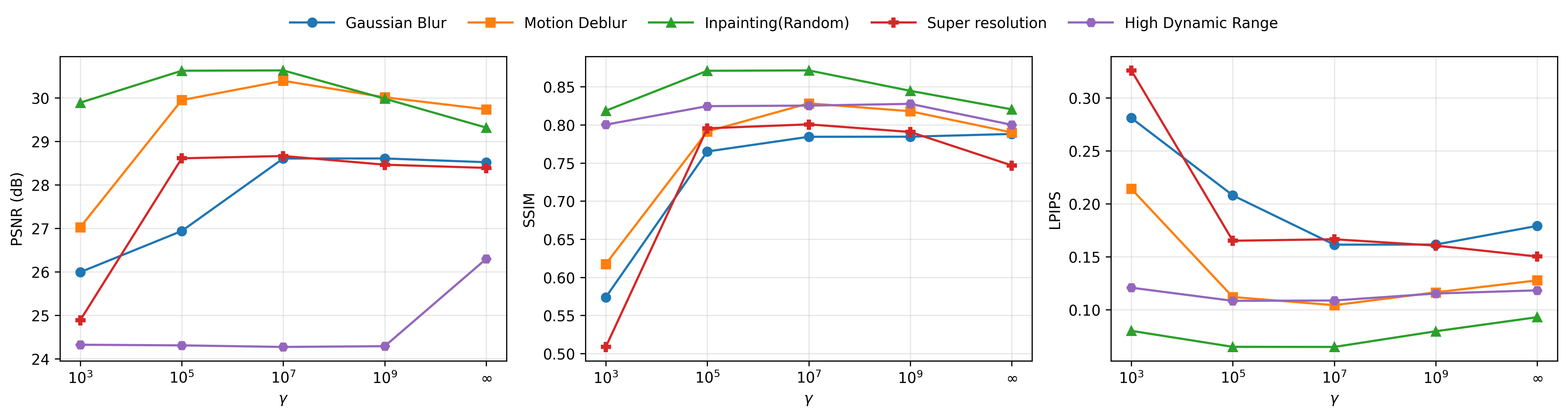}
  \caption{\textbf{Quantitative evaluation of image quality metrics as a function of the terminal penalty coefficient $\gamma$.}}
  \label{fig:metrics_vs_gamma}
\end{figure*}
\noindent \textbf{Terminal penalty coefficient of SOCS.} 
\cref{fig:metrics_vs_gamma} reports quantitative performance as a function of $\gamma$ over five tasks, and \cref{fig:gamma_performance} provides qualitative comparisons. We observe a clear trade-off: as $\gamma\to\infty$, the update becomes dominated by the reconstruction term, which can pull samples away from the generative prior and yield oversmoothed outputs; conversely, overly small $\gamma$ (e.g., $\sim10^{3}$) under-injects guidance, leading to noisy super-resolution and weaker measurement consistency in phase retrieval. From the posterior-sampling perspective in App.~\ref{app:socs_prior_posterior}, $\gamma$ acts as an effective noise precision balancing the diffusion prior and the measurement likelihood, with $\gamma=1/\sigma^2$ recovering the exact Gaussian posterior in canonical measurement space. In practice, preconditioning and the $r_t$-scaling of the data-fidelity term in the Langevin refinement rescale this trade-off, which motivates using numerically larger $\gamma$. Importantly, across operators and tasks, performance exhibits a broad plateau for finite $\gamma$, where metrics and visual quality vary little. This robustness stems from how $\gamma$ enters SOCS. In \cref{eq:MCMC-sampling-gamma}, it appears through a regularized inverse $(I+\gamma A)^{-1}$, so the effective guidance weight $\gamma(I+\gamma A)^{-1}$ saturates once $\gamma$ is sufficiently large. Consequently, SOCS does not require fine-grained tuning; we recommend $\gamma\in[10^{5},10^{9}]$ for a favorable balance between measurement fidelity and the generative prior.

\noindent \textbf{Computational efficiency of SOCS.} We run SOCS with various configurations to test performance under different computing budgets. Specifically, we evaluate with the number of function evaluations (NFE) of the diffusion models ranging from 400 to 4K, with configurations specified in Appendix~\ref{sec:Additional_results}.

\noindent \textbf{More ablation studies.} We discuss more additional ablation studies in Appendix~\ref{sec:Additional_results}, covering the number of function evaluations, the number of ODE steps, and the impact of the noise schedule on model performance.

\section{Conclusion}
In summary, we present SOCS for solving image inverse problems. Our method models the denoising dynamics through the lens of SOC and integrates them into sampling process, enforcing measurement consistency while gently tuning information injection to match the model’s native generative capacity. Empirically, SOCS delivers competitive reconstruction quality across a range of challenging inverse problems compared with existing methods.

\clearpage  

\section*{Acknowledgements}
The authors thank the anonymous reviewers for their valuable comments and constructive feedback, which have significantly enhanced the quality of this work. This research was supported by the AI for Science Program of the Shanghai Municipal Commission of Economy and Informatization (Grant No. 2025-GZL-RGZN-BTBX-02026) and the National Natural Science Foundation of China (Grant No. 62376155).

%
%
\bibliographystyle{splncs04}
\bibliography{main}

@String(ICML  = {Int. Conf. Mach. Learn.})

@String(ICLR  = {Int. Conf. Learn. Represent.})

@String(ICML  = {ICML})

@String(ICLR  = {ICLR})

@article{UniDB,
  title={UniDB: A Unified Diffusion Bridge Framework via Stochastic Optimal Control},
  author={Zhu, Kaizhen and Pan, Mokai and Ma, Yuexin and Fu, Yanwei and Yu, Jingyi and Wang, Jingya and Shi, Ye},
  journal={arXiv preprint arXiv:2502.05749},
  year={2025}
}

@inproceedings{DAPS,
  title={Improving diffusion inverse problem solving with decoupled noise annealing},
  author={Zhang, Bingliang and Chu, Wenda and Berner, Julius and Meng, Chenlin and Anandkumar, Anima and Song, Yang},
  booktitle={Proceedings of the Computer Vision and Pattern Recognition Conference},
  pages={20895--20905},
  year={2025}
}

@article{DPS,
  title={Diffusion posterior sampling for general noisy inverse problems},
  author={Chung, Hyungjin and Kim, Jeongsol and Mccann, Michael T and Klasky, Marc L and Ye, Jong Chul},
  journal={arXiv preprint arXiv:2209.14687},
  year={2022}
}

@article{solveIPviaDOC,
  title={Solving inverse problems via diffusion optimal control},
  author={Li, Henry and Pereira, Marcus},
  journal={Advances in Neural Information Processing Systems},
  volume={37},
  pages={73549--73571},
  year={2024}
}

@article{chen2024generativemodelingphasestochastic,
  title={Generative modeling with phase stochastic bridges},
  author={Chen, Tianrong and Gu, Jiatao and Dinh, Laurent and Theodorou, Evangelos A and Susskind, Joshua and Zhai, Shuangfei},
  journal={arXiv preprint arXiv:2310.07805},
  year={2023}
}

@article{rout2024rbmodulationtrainingfreepersonalizationdiffusion,
  title={Rb-modulation: Training-free personalization of diffusion models using stochastic optimal control},
  author={Rout, Litu and Chen, Yujia and Ruiz, Nataniel and Kumar, Abhishek and Caramanis, Constantine and Shakkottai, Sanjay and Chu, Wen-Sheng},
  journal={arXiv preprint arXiv:2405.17401},
  year={2024}
}

@inproceedings{FFHQ,
  title={A style-based generator architecture for generative adversarial networks},
  author={Karras, Tero and Laine, Samuli and Aila, Timo},
  booktitle={Proceedings of the IEEE/CVF conference on computer vision and pattern recognition},
  pages={4401--4410},
  year={2019}
}

@inproceedings{ImageNet,
  title={Imagenet: A large-scale hierarchical image database},
  author={Deng, Jia and Dong, Wei and Socher, Richard and Li, Li-Jia and Li, Kai and Fei-Fei, Li},
  booktitle={2009 IEEE conference on computer vision and pattern recognition},
  pages={248--255},
  year={2009},
  organization={Ieee}
}

@inproceedings{LPIPS,
  title={The unreasonable effectiveness of deep features as a perceptual metric},
  author={Zhang, Richard and Isola, Phillip and Efros, Alexei A and Shechtman, Eli and Wang, Oliver},
  booktitle={Proceedings of the IEEE conference on computer vision and pattern recognition},
  pages={586--595},
  year={2018}
}

@article{FID,
  title={Gans trained by a two time-scale update rule converge to a local nash equilibrium},
  author={Heusel, Martin and Ramsauer, Hubert and Unterthiner, Thomas and Nessler, Bernhard and Hochreiter, Sepp},
  journal={Advances in neural information processing systems},
  volume={30},
  year={2017}
}

@inproceedings{nonlin_blur_setting,
  title={Explore image deblurring via encoded blur kernel space},
  author={Tran, Phong and Tran, Anh Tuan and Phung, Quynh and Hoai, Minh},
  booktitle={Proceedings of the IEEE/CVF conference on computer vision and pattern recognition},
  pages={11956--11965},
  year={2021}
}

@article{dhariwal2021diffusionmodelsbeatgans,
  title={Diffusion models beat gans on image synthesis},
  author={Dhariwal, Prafulla and Nichol, Alexander},
  journal={Advances in neural information processing systems},
  volume={34},
  pages={8780--8794},
  year={2021}
}

@inproceedings{DDNM,
  title     = {Zero-Shot Image Restoration Using Denoising Diffusion Null-Space Model},
  author    = {Wang, Yinhuai and Yu, Jiwen and Zhang, Jian},
  booktitle = {ICLR},
  year      = {2023},
  eprint    = {2212.00490},
  archivePrefix = {arXiv},
  primaryClass  = {cs.CV}
}

@article{DDRM,
  title={Denoising diffusion restoration models},
  author={Kawar, Bahjat and Elad, Michael and Ermon, Stefano and Song, Jiaming},
  journal={Advances in neural information processing systems},
  volume={35},
  pages={23593--23606},
  year={2022}
}

@inproceedings{DiffPIR,
  title={Denoising diffusion models for plug-and-play image restoration},
  author={Zhu, Yuanzhi and Zhang, Kai and Liang, Jingyun and Cao, Jiezhang and Wen, Bihan and Timofte, Radu and Van Gool, Luc},
  booktitle={Proceedings of the IEEE/CVF conference on computer vision and pattern recognition},
  pages={1219--1229},
  year={2023}
}

@article{DCDP,
  title={Decoupled data consistency with diffusion purification for image restoration},
  author={Li, Xiang and Kwon, Soo Min and Liang, Shijun and Alkhouri, Ismail R and Ravishankar, Saiprasad and Qu, Qing},
  journal={arXiv preprint arXiv:2403.06054},
  year={2024}
}

@inproceedings{RED-Diff,
  title        = {A Variational Perspective on Solving Inverse Problems with Diffusion Models},
  author       = {Mardani, Morteza and Song, Jiaming and Kautz, Jan and Vahdat, Arash},
  booktitle    = {Proceedings of the 12th International Conference on Learning Representations (ICLR)},
  year         = {2024},
  url          = {https://openreview.net/forum?id=umG1nU1wZg},
  note         = {ICLR 2024}
}

@inproceedings{FPS-SMC,
  title={Diffusion posterior sampling for linear inverse problem solving: A filtering perspective},
  author={Dou, Zehao and Song, Yang},
  booktitle={The Twelfth International Conference on Learning Representations},
  year={2024}
}

@article{MCG,
  title={Improving diffusion models for inverse problems using manifold constraints},
  author={Chung, Hyungjin and Sim, Byeongsu and Ryu, Dohoon and Ye, Jong Chul},
  journal={Advances in Neural Information Processing Systems},
  volume={35},
  pages={25683--25696},
  year={2022}
}

@article{DPS-MO,
  title={Enhancing and accelerating diffusion-based inverse problem solving through measurements optimization},
  author={Chen, Tianyu and Wang, Zhendong and Zhou, Mingyuan},
  journal={arXiv preprint arXiv:2412.03941},
  year={2024}
}

@ARTICLE{1100008,
  author={Levine, W.},
  journal={IEEE Transactions on Automatic Control}, 
  title={Optimal control theory: An introduction}, 
  year={1972},
  volume={17},
  number={3},
  pages={423-423},
  doi={10.1109/TAC.1972.1100008}}

@inproceedings{2GDM,
  title={Pseudoinverse-guided diffusion models for inverse problems},
  author={Song, Jiaming and Vahdat, Arash and Mardani, Morteza and Kautz, Jan},
  booktitle={International Conference on Learning Representations},
  year={2023}
}

@inproceedings{peng2024improving,
  title        = {Improving Diffusion Models for Inverse Problems using Optimal Posterior Covariance},
  author       = {Peng, Xinyu and Zheng, Ziyang and Dai, Wenrui and Xiao, Nuoqian and Li, Chenglin and Zou, Junni and Xiong, Hongkai},
  booktitle    = {Proceedings of the 41st International Conference on Machine Learning (ICML)},
  series       = {Proceedings of Machine Learning Research},
  volume       = {235},
  pages        = {—},
  year         = {2024},
  publisher    = {PMLR},
  url          = {https://proceedings.mlr.press/v235/peng24a.html}
}

@article{PnP-DM,
  title={Principled probabilistic imaging using diffusion models as plug-and-play priors},
  author={Wu, Zihui and Sun, Yu and Chen, Yifan and Zhang, Bingliang and Yue, Yisong and Bouman, Katherine},
  journal={Advances in Neural Information Processing Systems},
  volume={37},
  pages={118389--118427},
  year={2024}
}

@article{EDM,
  title={Elucidating the design space of diffusion-based generative models},
  author={Karras, Tero and Aittala, Miika and Aila, Timo and Laine, Samuli},
  journal={Advances in neural information processing systems},
  volume={35},
  pages={26565--26577},
  year={2022}
}

@article{song2021scorebasedgenerativemodelingstochastic,
  title={Score-based generative modeling through stochastic differential equations},
  author={Song, Yang and Sohl-Dickstein, Jascha and Kingma, Diederik P and Kumar, Abhishek and Ermon, Stefano and Poole, Ben},
  journal={arXiv preprint arXiv:2011.13456},
  year={2020}
}

@inproceedings{rombach2022high,
  title={High-resolution image synthesis with latent diffusion models},
  author={Rombach, Robin and Blattmann, Andreas and Lorenz, Dominik and Esser, Patrick and Ommer, Bj{\"o}rn},
  booktitle={Proceedings of the IEEE/CVF conference on computer vision and pattern recognition},
  pages={10684--10695},
  year={2022}
}

@inproceedings{iLQR,
  title={Iterative linear quadratic regulator design for nonlinear biological movement systems},
  author={Li, Weiwei and Todorov, Emanuel},
  booktitle={First International Conference on Informatics in Control, Automation and Robotics},
  volume={2},
  pages={222--229},
  year={2004},
  organization={SciTePress}
}

@article{saharia2022image,
  title={Image super-resolution via iterative refinement},
  author={Saharia, Chitwan and Ho, Jonathan and Chan, William and Salimans, Tim and Fleet, David J and Norouzi, Mohammad},
  journal={IEEE transactions on pattern analysis and machine intelligence},
  volume={45},
  number={4},
  pages={4713--4726},
  year={2022},
  publisher={IEEE}
}

@inproceedings{saharia2022palette,
  title={Palette: Image-to-image diffusion models},
  author={Saharia, Chitwan and Chan, William and Chang, Huiwen and Lee, Chris and Ho, Jonathan and Salimans, Tim and Fleet, David and Norouzi, Mohammad},
  booktitle={ACM SIGGRAPH 2022 conference proceedings},
  pages={1--10},
  year={2022}
}

@inproceedings{xia2023diffir,
  title={Diffir: Efficient diffusion model for image restoration},
  author={Xia, Bin and Zhang, Yulun and Wang, Shiyin and Wang, Yitong and Wu, Xinglong and Tian, Yapeng and Yang, Wenming and Van Gool, Luc},
  booktitle={Proceedings of the IEEE/CVF international conference on computer vision},
  pages={13095--13105},
  year={2023}
}

@article{DDPM,
  title={Denoising diffusion probabilistic models},
  author={Ho, Jonathan and Jain, Ajay and Abbeel, Pieter},
  journal={Advances in neural information processing systems},
  volume={33},
  pages={6840--6851},
  year={2020}
}

@inproceedings{karras2024analyzing,
  title={Analyzing and improving the training dynamics of diffusion models},
  author={Karras, Tero and Aittala, Miika and Lehtinen, Jaakko and Hellsten, Janne and Aila, Timo and Laine, Samuli},
  booktitle={In Proceedings of the IEEE/CVF Conference on Computer Vision and Pattern Recognition},
  pages={24174--24184},
  year={2024}
}

@article{song2019generative,
  title={Generative modeling by estimating gradients of the data distribution},
  author={Song, Yang and Ermon, Stefano},
  journal={Advances in neural information processing systems},
  volume={32},
  year={2019}
}

@article{song2020improved,
  title={Improved techniques for training score-based generative models},
  author={Song, Yang and Ermon, Stefano},
  journal={Advances in neural information processing systems},
  volume={33},
  pages={12438--12448},
  year={2020}
}

@article{boys2023tweedie,
  title={Tweedie moment projected diffusions for inverse problems},
  author={Boys, Benjamin and Girolami, Mark and Pidstrigach, Jakiw and Reich, Sebastian and Mosca, Alan and Akyildiz, O Deniz},
  journal={arXiv preprint arXiv:2310.06721},
  year={2023}
}

@article{VAE,
  title={Auto-encoding variational bayes},
  author={Kingma, Diederik P and Welling, Max},
  journal={arXiv preprint arXiv:1312.6114},
  year={2013}
}

@inproceedings{tassa2012synthesis,
  title={Synthesis and stabilization of complex behaviors through online trajectory optimization},
  author={Tassa, Yuval and Erez, Tom and Todorov, Emanuel},
  booktitle={2012 IEEE/RSJ International Conference on Intelligent Robots and Systems},
  pages={4906--4913},
  year={2012},
  organization={IEEE}
}

@article{geman1984stochastic,
  title={Stochastic relaxation, Gibbs distributions, and the Bayesian restoration of images},
  author={Geman, Stuart and Geman, Donald},
  journal={IEEE Transactions on pattern analysis and machine intelligence},
  number={6},
  pages={721--741},
  year={1984},
  publisher={IEEE}
}

@article{perona2002scale,
  title={Scale-space and edge detection using anisotropic diffusion},
  author={Perona, Pietro and Malik, Jitendra},
  journal={IEEE Transactions on pattern analysis and machine intelligence},
  volume={12},
  number={7},
  pages={629--639},
  year={2002},
  publisher={IEEE}
}

@article{freeman2002example,
  title={Example-based super-resolution},
  author={Freeman, William T and Jones, Thouis R and Pasztor, Egon C},
  journal={IEEE Computer graphics and Applications},
  volume={22},
  number={2},
  pages={56--65},
  year={2002},
  publisher={IEEE}
}

@article{rout2023solving,
  title={Solving linear inverse problems provably via posterior sampling with latent diffusion models},
  author={Rout, Litu and Raoof, Negin and Daras, Giannis and Caramanis, Constantine and Dimakis, Alex and Shakkottai, Sanjay},
  journal={Advances in Neural Information Processing Systems},
  volume={36},
  pages={49960--49990},
  year={2023}
}

@article{berner2022optimal,
  title={An optimal control perspective on diffusion-based generative modeling},
  author={Berner, Julius and Richter, Lorenz and Ullrich, Karen},
  journal={arXiv preprint arXiv:2211.01364},
  year={2022}
}

@inproceedings{todorov2005generalized,
  title={A generalized iterative LQG method for locally-optimal feedback control of constrained nonlinear stochastic systems},
  author={Todorov, Emanuel and Li, Weiwei},
  booktitle={Proceedings of the 2005, American Control Conference, 2005.},
  pages={300--306},
  year={2005},
  organization={IEEE}
}

@incollection{merton1975optimum,
  title={Optimum consumption and portfolio rules in a continuous-time model},
  author={Merton, Robert C},
  booktitle={Stochastic optimization models in finance},
  pages={621--661},
  year={1975},
  publisher={Elsevier}
}

@article{behncke2000optimal,
  title={Optimal control of deterministic epidemics},
  author={Behncke, Horst},
  journal={Optimal control applications and methods},
  volume={21},
  number={6},
  pages={269--285},
  year={2000},
  publisher={Wiley Online Library}
}

@book{neely2010stochastic,
  title={Stochastic network optimization with application to communication and queueing systems},
  author={Neely, Michael},
  year={2010},
  publisher={Morgan \& Claypool Publishers}
}

@inproceedings{welling2011bayesian,
  title={Bayesian learning via stochastic gradient Langevin dynamics},
  author={Welling, Max and Teh, Yee W},
  booktitle={Proceedings of the 28th international conference on machine learning (ICML-11)},
  pages={681--688},
  year={2011}
}

@article{Flowchef,
  title={Steering rectified flow models in the vector field for controlled image generation},
  author={Patel, Maitreya and Wen, Song and Metaxas, Dimitris N and Yang, Yezhou},
  journal={arXiv preprint arXiv:2412.00100},
  year={2024}
}

@inproceedings{Flowdps,
  title={Flowdps: Flow-driven posterior sampling for inverse problems},
  author={Kim, Jeongsol and Kim, Bryan Sangwoo and Ye, Jong Chul},
  booktitle={Proceedings of the IEEE/CVF International Conference on Computer Vision},
  pages={12328--12337},
  year={2025}
}

\clearpage
\appendix
\section{Proof}
\label{sec:rationale}
\subsection{Proof of Theorem 3.1}
\label{sec:Proof of Theorem 3.1}
Consider the SOC problem Eq.~\eqref{eq:SOC_construct_ODE}. Denote $d_{0,\gamma}^{-1} := \bigl(I + \gamma\,e^{2\bar f_T}\,\bar g_T^{\,2} J_\mathcal{H}(x_T^u)^\top \circ \mathcal{H}\bigr)^{-1},\bar f_{s:t} := \int_s^t f_z\,\mathrm dz$ and $\bar g^{\,2}_{s:t} := \int_s^t e^{-2\bar f_z}g_z^{\,2}\,\mathrm dz$, for simplicity, write \(\bar f_t := \bar f_{0:t}\) and \(\bar g_t^{\,2} := \bar g^{\,2}_{0:t}\). Here $J_\mathcal{H}(x_T^u)$ is the Jacobian of $\mathcal{H}(x)$ evaluated at $x_T^u$, and the symbol $\circ$ denotes operator composition. Then the closed-form optimal controller $\mathbf{u}_{t,\gamma}^\star$ is:
\begin{equation}
\mathbf{u}_{t,\gamma}^*=-g_t\gamma e^{\bar{f}_{t:T}}d_{0,\gamma}^{-1} J_\mathcal{H}(x_T^u)^\top\bigl(e^{\bar{f}_T}\mathcal{H}(x_0) - y\bigr),
\label{eq:SOC_u_t_appendix}
\end{equation}

and the transition of $x_t$ from $x_0$ is:
\begin{equation}
\begin{aligned}
x_t &= e^{\bar{f}_t}x_0 + \bigl(\gamma e^{\bar{f}_t} e^{\bar{f}_T} \bar{g}_t^2d_{0,\gamma}^{-1} J_\mathcal{H}(x_T^u)^\top \bigr)\bigl(y-e^{\bar{f}_T}\mathcal{H}(x_0)\bigr).
\end{aligned}
\label{eq:SOC_x_t_proof}
\end{equation}
Proof. According to Pontryagin Maximum Principle~\cite{1100008} recipe, one can construct the Hamiltonian:
\begin{equation}
    H(t,x_t,\mathbf{u}_{t,\gamma},\mathbf{P}_t)=\dfrac{1}{2}||\mathbf{u}_{t,\gamma}||_2^2 + \mathbf{P}_t^\top(f_tx_t+g_t\mathbf{u}_{t,\gamma}).
    \label{eq:Hamiltonian}
\end{equation}
By setting:
\begin{equation}
    \dfrac{\partial H}{\partial \mathbf{u}_{t,\gamma}} = 0 \Rightarrow \mathbf{u}_{t,\gamma}^* = -g_t\mathbf{P}_t.
\end{equation}
Then the value function becomes:
\begin{equation}
\begin{aligned}
    V^* &= \min_{\mathbf{u}_{t,\gamma}} H(t,x_t,\mathbf{u}_{t,\gamma},\mathbf{P}_t)=H(t,x_t,\mathbf{u}_{t,\gamma}^*,\mathbf{P}_t)\\
    &= -\dfrac{1}{2}g_t^2 ||\mathbf{P}_t||_2^2 + f_tx_t.
\end{aligned}
\end{equation}
Now, according to minimum principle theorem to obtain the following set of differential equations:
\begin{gather}
\frac{\mathrm d x_t}{\mathrm d t}
  = \nabla_{\mathbf P_t}\, H\!\left(t,x_t,\mathbf u_{t,\gamma}^\ast,\mathbf P_t\right)
  = -\,g_t^{2}\,\mathbf P_t + f_t\,x_t, \label{eq:dx}\\[2pt]
\frac{\mathrm d \mathbf P_t}{\mathrm d t}
  = -\,\nabla_{x_t}\, H\!\left(t,x_t,\mathbf u_{t,\gamma}^\ast,\mathbf P_t\right)
  = -\,\mathbf P_t\,f_t, \label{eq:dP}\\[2pt]
x_0^{u} = x_0, \label{eq:x_0^u}\\[2pt]
\mathbf P_T
  = -\,\gamma\, J_{\mathcal H}\!\left(x_T^{u}\right)^{\top}\!
      \Bigl(y - \mathcal H(x_T)\Bigr). \label{eq:P_T}
\end{gather}
Solving the Eq.~\eqref{eq:dP}, we have:
\begin{equation}
\begin{gathered}
\mathbf{P}_t = P_0e^{-\bar{f}_t},\\
\mathbf{P}_T = P_0e^{-\bar{f}_T},
\end{gathered}
\label{eq:P_t&P_T}
\end{equation}
Solve the Eq.~\eqref{eq:dx}:
\begin{equation}
\begin{gathered}
\dfrac{\mathrm dx_t}{dt}=-g_t^2 \mathbf{P}_t + f_tx_t,\\
\Rightarrow \dfrac{\mathrm d(e^{-\bar{f}_t}x_t)}{dt} = -e^{-\bar{f}_t} g_t^2 \mathbf{P}_t.\\
\end{gathered}
\end{equation}
Hence, we can get:
\begin{equation}
    x_T = e^{\bar{f}_T}x_0 - \mathbf{P}_Te^{2\bar{f}_T}\bar{g}_T^2, \label{eq:x_T}
\end{equation}
and:
\begin{equation}
    x_t = e^{\bar{f}_t}x_0-e^{\bar{f}_t}P_0 \bar{g}_t^2=e^{\bar{f}_t}x_0 - \mathbf{P}_Te^{\bar{f}_t}e^{\bar{f}_T}\bar{g}_t^2. \label{eq:x_t}
\end{equation}
Take the Eq.~\eqref{eq:x_T} into the Eq.~\eqref{eq:P_T} and solve $\mathbf{P}_T$,
\begin{equation}
    \begin{aligned}
        \mathbf{P}_T =& -\gamma J_\mathcal{H}(x_T^u)^\top\bigl(y-e^{\bar{f}_T}\mathcal{H}(x_0) +  \mathbf{P}_Te^{2\bar{f}_T}\bar{g}_T^2\mathcal{H}\bigr), \label{eq:P_T_l1} \\
       \Rightarrow \mathbf{P}_T =& \gamma\underbrace{\bigl(I + \gamma e^{2\bar{f}_T}\bar{g}_T^2 J_\mathcal{H}(x_T^u)^\top \circ\mathcal{H}\bigr)^{-1}}_{d_{0,\gamma}^{-1}}\\ 
       &\circ J_\mathcal{H}(x_T^u)^\top \bigl(e^{\bar{f}_T}\mathcal{H}(x_0) - y\bigr).
    \end{aligned} 
\end{equation}
Also, take the Eq.~\eqref{eq:P_T_l1} into the Eq.~\eqref{eq:x_t},
\begin{equation}
\begin{aligned}
x_t &= e^{\bar{f}_t}x_0 + \bigl(\gamma e^{\bar{f}_t} e^{\bar{f}_T} \bar{g}_t^2d_{0,\gamma}^{-1} J_\mathcal{H}(x_T^u)^\top \bigr)\bigl(y-e^{\bar{f}_T}\mathcal{H}(x_0)\bigr).
\end{aligned} \label{eq:final_x_t}
\end{equation}
Preserve $\gamma$,
\begin{equation}
    \begin{aligned}
        \mathbf{u}_{t,\gamma}^* &= -g_t\mathbf{P}_T\\
        &=-g_t\mathbf{P}_Te^{\bar{f}_{t:T}}\\
        &=-g_t\gamma e^{\bar{f}_{t:T}}d_{0,\gamma}^{-1} J_\mathcal{H}(x_T^u)^\top \bigl(e^{\bar{f}_T}\mathcal{H}(x_0) - y\bigr),
    \end{aligned}\label{eq:final_u_t}
\end{equation}
which concludes the proof of the Eq.~\eqref{eq:SOC_u_t} and Eq.~\eqref{eq:SOC_x_t}.

\subsection{Proof of Optimality Inequality}\label{sec:Proof_Proposition 3.2}
Consider the SOC problem Eq.~\eqref{eq:SOC_construct_ODE}, denote $\mathcal{J}(\mathbf{u}_{t,\gamma},\gamma) \triangleq \int_0^T \tfrac{1}{2}\bigl\lVert\mathbf{u}_{t,\gamma}\bigr\lVert_2^2\,\mathrm dt+ \tfrac{\gamma}{2}\bigl\lVert y-\mathcal{H}(x_T)\bigr\lVert_2^2$ as the overall cost of the system, $\mathbf{u}_{t,\gamma}^*$ as the optimal controller \eqref{eq:SOC_u_t_appendix}, then
\begin{equation}
    \mathcal{J}(\mathbf{u}_{t,\gamma}^*,\gamma) \le \mathcal{J}(\mathbf{u}_{t,\infty}^*,\infty).
    \label{eq:finite_gamma_better}
\end{equation}
Proof. Throughout the proof we set $J:=J_\mathcal{H}(x_T^u)$ and adopt the Gauss-Newton linearized model at $x_T^u$. Define $C:=e^{2\bar f_T}\bar g_T^{2}J^\top J$, $m_t:= g_te^{\bar{f}_{t:T}}$, $d_{0,\gamma}^{-1}:=(I + \gamma C)^{-1}$, $a:= y-e^{\bar{f}_T}\mathcal{H}(x_0)$, $v:=J^\top a$. 
Take $t=T$ in Eq.~\eqref{eq:SOC_x_t},
\begin{equation}
x_T^u = e^{\bar{f}_T}x_0 + \gamma e^{2\bar f_T}\bar{g}_T^{\,2} d_{0,\gamma}^{-1} J^\top a.
\end{equation}
We obtain
\begin{equation}
\begin{aligned}
    &y - \mathcal{H}(x_T) =a - e^{2\bar f_T}\bar{g}_T^{\,2} J\gamma d_{0,\gamma}^{-1} J^\top a, \\
\Rightarrow &J^\top\bigl(y - \mathcal{H}(x_T)\bigr) \\
&=J^\top a-e^{2\bar f_T}\bar{g}_T^{\,2}J^\top J\gamma d_{0,\gamma}^{-1}J^\top a \\
&=\bigl(I - \gamma C(I + \gamma C)^{-1}\bigr)J^\top a.
\end{aligned}
\end{equation}
Using the identity  $I - \gamma C(I + \gamma C)^{-1}=(I + \gamma C)^{-1}=d_{0,\gamma}^{-1}$, it follows that
\begin{equation}
J^\top\bigl(y-\mathcal{H}(x_T)\bigr) = d_{0,\gamma}^{-1}J^\top a.
\end{equation}
If $J$ has full row rank in the measurement dimension, then
\begin{equation}
\begin{aligned}
\bigl\lVert y-\mathcal{H}(x_T^u)\bigr\lVert_2 &\le \dfrac{1}{\sigma_{min}(J_\mathcal{H})}\bigl\lVert(I + \gamma C)^{-1}v\bigr\lVert_2 \\
&\le \dfrac{\bigl\lVert v\bigr\lVert_2}{\sigma_{min}(J_\mathcal{H})(I + \gamma \lambda_{min}(C))}.
\end{aligned}
\end{equation}
Hence
\begin{equation}
    \lim_{\gamma \to \infty}\dfrac{\gamma}{2}\bigl\lVert y-\mathcal{H}(x_T)\bigr\lVert_2^2=0.
\end{equation}
Furthermore, according to Eq.~\eqref{eq:SOC_u_t_appendix}, 
\begin{equation}
\begin{aligned}
    &\mathbf{u}_{t,\gamma}^*=m_t \gamma d_{0,\gamma}^{-1} J^\top a, \\
\Rightarrow &\bigl\lVert\mathbf{u}_{t,\gamma}^*\bigr\lVert_2^2=m_t^2 \bigl\lVert\gamma d_{0,\gamma}^{-1} J^\top a\bigr\lVert_2^2.
\end{aligned}
\end{equation}
Similarly,
\begin{equation}
\begin{aligned}
    &\mathbf{u}_{t,\infty}^*=m_t C^{-1} J^\top a \\
\Rightarrow &\bigl\lVert\mathbf{u}_{t,\infty}^*\bigr\lVert_2^2=m_t^2\bigl\lVert C^{-1} J^\top a\bigr\lVert_2^2.
\end{aligned}
\end{equation}
Since $\int_0^T m_t^2 dt=e^{2\bar f_T}\bar{g}_T^{\,2}$, we obtain
\begin{equation}
\begin{aligned}
&\dfrac{1}{2}\int_0^T(\bigl\lVert\mathbf{u}_{t,\infty}^*\bigr\lVert_2^2-\bigl\lVert\mathbf{u}_{t,\gamma}^*\bigr\lVert_2^2)dt\\
&= \dfrac{1}{2}\int_0^T m_t^2(\bigl\lVert\gamma d_{0,\gamma}^{-1} J^\top a\bigr\lVert_2^2 - \bigl\lVert C^{-1} J^\top a\bigr\lVert_2^2) dt, \\
&=\dfrac{1}{2}e^{2\bar f_T}\bar{g}_T^{\,2} \cdot (J^\top a)^{\top}(C^{-2}-\gamma^2(I + \gamma C)^{-2})v. \\
\end{aligned}
\end{equation}
Assume $C \succeq 0$ with eigendecomposition $C=U\Lambda U^\top$ where $\Lambda=\mathrm{diag}(\lambda_i)$ and $\lambda_i \ge 0$, and set $w:=U^\top v$. Then
\begin{equation}
\begin{aligned}
&\dfrac{1}{2}\int_0^T(\bigl\lVert\mathbf{u}_{t,\infty}^*\bigr\lVert_2^2-\bigl\lVert\mathbf{u}_{t,\gamma}^*\bigr\lVert_2^2)dt\\
&=\dfrac{1}{2}e^{2\bar f_T}\bar{g}_T^{\,2} \sum_i(\dfrac{1}{\lambda_i^2}-\dfrac{\gamma^2}{(1+\gamma \lambda_i)^2})w_i^2. 
\end{aligned}\label{eq:res_left}
\end{equation}
On the other hand, by Woodbury under the same GN model,
\begin{equation}
y-\mathcal{H}(x_T) \approx \bigl(I + \gamma e^{2\bar f_T}\bar{g}_T^{\,2}J J^\top\bigr)^{-1}a.
\end{equation}
Let $J=P\Sigma U^\top$ be the SVD of $J$ and set $\lambda_i=e^{2\bar f_T}\bar{g}_T^{\,2}\sigma_i^2$. Writing $a = P\alpha$ and using $w=U^\top v=U^\top J^\top a=\Sigma \alpha$ (i.e.,$\alpha_i=w_i/\sigma_i$), we obtain
\begin{equation}
\begin{aligned}
\dfrac{\gamma}{2}\bigl\lVert y-\mathcal{H}(x_T)\bigr\lVert_2^2&=\dfrac{\gamma}{2}\sum_i\dfrac{\alpha_i^2}{(1+\gamma e^{2\bar f_T}\bar{g}_T^{\,2}\sigma_i^2)^2}, \\
&=\dfrac{1}{2}e^{2\bar f_T}\bar{g}_T^{\,2} \sum_i \dfrac{\gamma}{\lambda_i(1+\gamma \lambda_i)^2}w_i^2.
\end{aligned}\label{eq:res_right}
\end{equation}
Comparing the coefficients along each eigen-direction in Eq.~\eqref{eq:res_left} and Eq.~\eqref{eq:res_right},
\begin{equation}
\dfrac{1}{\lambda_i^2}-\dfrac{\gamma^2}{(1+\gamma \lambda_i)^2} -\dfrac{\gamma}{\lambda_i(1+\gamma \lambda_i)^2}=\dfrac{1}{\lambda_i^2(1+\gamma \lambda_i)} \ge 0.
\end{equation}
which implies
\begin{equation}
\begin{aligned}
\dfrac{1}{2}&\int_0^T(\bigl\lVert\mathbf{u}_{t,\infty}^*\bigr\lVert_2^2-\bigl\lVert\mathbf{u}_{t,\gamma}^*\bigr\lVert_2^2)dt \\
&= \dfrac{1}{2}e^{2\bar f_T}\bar{g}_T^{\,2} \sum_i(\dfrac{1}{\lambda_i^2}-\dfrac{\gamma^2}{(1+\gamma \lambda_i)^2})w_i^2 \\
&\ge \dfrac{\gamma}{2} e^{2\bar f_T}\bar{g}_T^{\,2} \sum_i \dfrac{w_i^2}{\lambda_i(1+\gamma \lambda_i)^2} \\
&=\dfrac{\gamma}{2} \bigl\lVert y-\mathcal{H}(x_T)\bigr\lVert_2^2\\
&=\dfrac{\gamma}{2} \bigl\lVert y-\mathcal{H}(x_T)\bigr\lVert_2^2 - \lim_{\gamma \to \infty}\dfrac{\gamma}{2}\bigl\lVert y-\mathcal{H}(x_T)\bigr\lVert_2^2.
\end{aligned}
\end{equation}
Therefore,
\begin{equation}
\begin{aligned}
&\dfrac{\gamma}{2} \bigl\lVert y-\mathcal{H}(x_T)\bigr\lVert_2^2 - \lim_{\gamma \to \infty}\dfrac{\gamma}{2}\bigl\lVert y-\mathcal{H}(x_T)\bigr\lVert_2^2 \\
&\le \dfrac{1}{2}\int_0^T(\bigl\lVert\mathbf{u}_{t,\infty}^*\bigr\lVert_2^2-\bigl\lVert\mathbf{u}_{t,\gamma}^*\bigr\lVert_2^2)dt, \\
\iff &\int_0^T \dfrac{1}{2}\bigl\lVert\mathbf{u}_{t,\gamma}^*\bigr\lVert_2^2\,\mathrm dt+ \dfrac{\gamma}{2}\bigl\lVert y-\mathcal{H}(x_T)\bigr\lVert_2^2 \\
&\le \int_0^T \dfrac{1}{2}\bigl\lVert\mathbf{u}_{t,\infty}^*\bigr\lVert_2^2\,\mathrm dt+ \lim_{\gamma \to \infty}\dfrac{\gamma}{2}\bigl\lVert y-\mathcal{H}(x_T)\bigr\lVert_2^2, \\
\iff &J(\mathbf{u}_{t,\gamma}^*,\gamma) \le J(\mathbf{u}_{t,\infty}^*,\infty).
\end{aligned}
\end{equation}
This completes the proof of optimality inequality. The above argument shows that the optimal total cost of the soft-constrained problem with a finite penalty weight cannot exceed that of the hard-constrained formulation that enforces exact measurement consistency; in particular, its optimal value provides a lower bound on the minimum cost of the hard-constrained problem. This highlights the advantage of keeping the terminal penalty finite.

\subsection{Proof of Convergence Guarantees}\label{sec:Proof_Proposition 3.3}
Denote the initial state distribution $x_0$, the terminal distribution $x_T$ by the controller, the predefined terminal distribution $y$ and $J=J_\mathcal{H}(x_T^u)$. Under a linear/linearized measurement model, we have
\begin{equation}
\begin{aligned}
\bigl\|y-\mathcal H(x_T)\bigr\|_2^2
&=
\bigl(y-e^{\bar f_T}\mathcal H(x_0)\bigr)^\top
\Bigl(I+\gamma e^{2\bar f_T}\bar g_T^{2}JJ^\top\Bigr)^{-2}
\\
&\quad
\bigl(y-e^{\bar f_T}\mathcal H(x_0)\bigr).
\end{aligned}
\label{eq:r-gamma-quadratic}
\end{equation}
Proof. Recall \cref{sec:Proof of Theorem 3.1} that for the VP-ODE $\mathrm dx_t=(f_t x_t+g_t \mathbf{u}_t)\mathrm dt$ the optimal controlled trajectory reads
\begin{equation}
\begin{aligned}
x_t &= e^{\bar{f}_t}x_0 + \left(\gamma e^{\bar{f}_t} e^{\bar{f}_T} \bar{g}_t^2d_{0,\gamma}^{-1} J_\mathcal{H}(x_T^u)^\top  \right)\big(y-e^{\bar{f}_T}\mathcal H(x_0)\big).
\end{aligned}
\end{equation}
where $d_{0,\gamma}^{-1}:=(I+\gamma C)^{-1}$,with $C:=e^{2\bar{f}_T}\bar{g}_T^{\,2}$, we donate $J:=J_\mathcal{H}(x_T^u)$, then have $J^\top J\succeq0$.
Taking $t=T$ gives
\begin{equation}\label{eq:xT-closed}
x_T=e^{\bar f_T}x_0 + \big(e^{2\bar f_T}\bar g_T^{\,2}\big)\,\gamma(I+\gamma C)^{-1}J^\top\,\big(y-e^{\bar f_T}\mathcal H(x_0)\big).
\end{equation}
Under the linear (or Gauss--Newton linearized at $x_T^u$) measurement model
\begin{equation}
\mathcal H(x)\approx \mathcal H(x_T^u)+J(x-x_T^u),
\end{equation}
with $J=J_\mathcal{H}(x_T^u)$, Woodbury’s identity yields
\begin{equation}\label{eq:r-gamma}
y-\mathcal H(x_T)
=(I+\gamma\,e^{2\bar f_T}\bar g_T^{\,2}JJ^\top)^{-1}\big(y-e^{\bar f_T}\mathcal H(x_0)\big).
\end{equation}
Consequently, the squared norm of the residual admits the quadratic form
\begin{equation}
\begin{aligned}
\bigl\|y-\mathcal H(x_T)\bigr\|_2^2
&=
\bigl(y-e^{\bar f_T}\mathcal H(x_0)\bigr)^\top
\Bigl(I+\gamma e^{2\bar f_T}\bar g_T^{2}JJ^\top\Bigr)^{-2}
\\
&\quad
\bigl(y-e^{\bar f_T}\mathcal H(x_0)\bigr).
\end{aligned}
\label{eq:residual-quadratic-gamma}
\end{equation}
Let $JJ^\top = P\Lambda P^\top$ be the eigenvalue decomposition with
$\Lambda=\mathrm{diag}(\lambda_i)$ and $\lambda_i\ge 0$, and write
$a = P^\top\bigl(y-e^{\bar f_T}\mathcal H(x_0)\bigr)$.
Then \eqref{eq:residual-quadratic-gamma} becomes
\begin{equation}
\bigl\|y-\mathcal H(x_T)\bigr\|_2^2
=
\sum_{i}
\frac{a_i^2}{\bigl(1+\gamma e^{2\bar f_T}\bar g_T^{2}\lambda_i\bigr)^2}.
\end{equation}
which concludes the proof of convergence guarantees. This expression indicates that as $\gamma \to \infty$ the terminal constraint is enforced exactly, but the intermediate control injection becomes unconstrained, potentially driving the dynamics far away from the native denoising behavior of the diffusion model. In this sense, a finite $\gamma$ helps strike a more favorable balance between fidelity to the measurements and the overall generation quality.

\subsection{Proof of the Interpretation (prior proximity via control energy)}
\label{app:bounded_energy_prior_proximity} 
Consider the native diffusion:
$\mathrm{d}x_t = f(x_t,t)\,\mathrm{d}t + g_t\,\mathrm{d}W_t$
and the controlled process in \cref{eq:SOC_construct_SDE}:
$\mathrm{d}x_t = \big(f(x_t,t)+ g_t \mathbf u_{t,\gamma}\big)\,\mathrm{d}t + g_t\,\mathrm{d}W_t$
where $x_0 \sim \mu_0$ and $\mathbf u_{t,\gamma}$ is progressively measurable w.r.t.\ the filtration generated by $(x_s)_{s\le t}$.
Let $P_{\mathrm{path}}$ and $Q_{\mathrm{path}}$ denote the path measures on $x_{0:T}$ induced by the native and controlled SDEs, respectively,
and let $P_T,Q_T$ be the corresponding terminal marginals.

In our setting, the (standard) Novikov condition is satisfied by construction: it coincides with the integrability requirement imposed by the SOC objective in \cref{eq:SOC_construct_SDE},
\begin{equation}
\mathbb{E}_{P_{\mathrm{path}}}\!\left[\exp\!\left(\frac12\int_0^T \|\mathbf u_{t,\gamma}\|^2\,\mathrm{d}t\right)\right] < \infty.
\label{eq:novikov_app}
\end{equation}
Then $Q_{\mathrm{path}}$ is absolutely continuous w.r.t.\ $P_{\mathrm{path}}$, and the Radon--Nikodym derivative is given by
\begin{equation}
\frac{\mathrm{d}Q_{\mathrm{path}}}{\mathrm{d}P_{\mathrm{path}}}
= \exp\!\left(\int_0^T \langle u_{t,\gamma},\mathrm{d}W_t\rangle
-\frac12\int_0^T \|u_{t,\gamma}\|^2\,\mathrm{d}t\right).
\label{eq:RN_girsanov}
\end{equation}
Moreover, under $Q_{\mathrm{path}}$, the process $W_t^u := W_t-\int_0^t \mathbf u_{s,\gamma}\,\mathrm{d}s$ is a Brownian motion,
and the controlled drift indeed becomes $f(x_t,t)+g_t \mathbf u_{t,\gamma}$.

Using \cref{eq:RN_girsanov} and taking expectation under $Q_{\mathrm{path}}$,
\begin{align}
D_{\mathrm{KL}}(Q_{\mathrm{path}}\|P_{\mathrm{path}})
&= \mathbb{E}_{Q_{\mathrm{path}}}\!\left[\log\frac{\mathrm{d}Q_{\mathrm{path}}}{\mathrm{d}P_{\mathrm{path}}}\right] \nonumber\\
&= \mathbb{E}_{Q_{\mathrm{path}}}\!\left[\int_0^T \langle \mathbf u_{t,\gamma},\mathrm{d}W_t\rangle
-\frac12\int_0^T \|\mathbf u_{t,\gamma}\|^2\,\mathrm{d}t\right]. \label{eq:kl_path_mid}
\end{align}
Under $Q_{\mathrm{path}}$, we have $\mathrm{d}W_t=\mathrm{d}W_t^u + u_{t,\gamma}\,\mathrm{d}t$ and
$\mathbb{E}_{Q_{\mathrm{path}}}\!\big[\int_0^T \langle u_{t,\gamma},\mathrm{d}W_t^u\rangle\big]=0$,
hence \eqref{eq:kl_path_mid} simplifies to
\begin{equation}
D_{\mathrm{KL}}(Q_{\mathrm{path}}\|P_{\mathrm{path}})
= \frac12\,\mathbb{E}_{Q_{\mathrm{path}}}\!\left[\int_0^T \|\mathbf u_{t,\gamma}\|^2\,\mathrm{d}t\right].
\label{eq:kl_path_energy}
\end{equation}

Let $\pi_T:x_{0:T}\mapsto x_T$ be the terminal projection. Since $Q_T=\mathcal{L}_{Q_{\mathrm{path}}}(x_T) $ and $P_T=\mathcal{L}_{P_{\mathrm{path}}}(x_T) $,
the KL divergence contracts under measurable maps, giving
\begin{equation}
D_{\mathrm{KL}}(Q_T\|P_T)\le D_{\mathrm{KL}}(Q_{\mathrm{path}}\|P_{\mathrm{path}})
= \frac12\,\mathbb{E}_{Q_{\mathrm{path}}}\!\left[\int_0^T \|\mathbf u_{t,\gamma}\|^2\,\mathrm{d}t\right].
\label{eq:kl_marg_app}
\end{equation}
Therefore, bounded control energy directly bounds the deviation of the terminal distribution from the diffusion prior.
In particular, if $\mathbb{E}_{Q_{\mathrm{path}}}\!\left[\int_0^T \|\mathbf u_{t,\gamma}\|^2\,\mathrm{d}t\right]\le 2\varepsilon$,
then $D_{\mathrm{KL}}(Q_T\|P_T)\le \varepsilon$.

\subsection{Proof of the Interpretation (posterior sampling)}
\label{app:socs_prior_posterior}

At each step, SOCS constructs a control $\mathbf u_{t,\gamma}$ by solving \cref{eq:SOC_construct_ODE}, i.e., minimizing the control energy
$\int_0^T \frac12\|\mathbf u_{t,\gamma}\|_2^2\,\mathrm dt$
together with the terminal measurement penalty
$\phi(x_T)=\frac{\gamma}{2}\|y-\mathcal H(x_T)\|_2^2$.
To interpret SOCS in terms of measure change and posterior sampling, we consider the stochastic counterpart with the same drift/control injection:
\begin{equation}
\mathrm d x_t = (f_t x_t + g_t \mathbf u_{t,\gamma})\,\mathrm dt + g_t\,\mathrm dW_t,
\qquad x_0\sim \mu_0,
\end{equation}
and the native diffusion obtained by setting $\mathbf u_{t,\gamma}\equiv 0$.
Denote the induced path measures on $x_{0:T}$ by $Q_{\mathrm{path}}$ and $P_{\mathrm{path}}$, and the terminal marginals by $Q_T$ and $P_T$.
By \cref{eq:kl_marg_app}, bounded control energy implies a bounded terminal KL deviation $D_{\mathrm{KL}}(Q_T\|P_T)$, hence the controlled flow cannot drift arbitrarily far from the diffusion prior in distribution.

Moreover, the SOC objective admits an equivalent variational form over path measures,
whose minimizer is a Gibbs reweighting of the diffusion prior:
\begin{equation}
\frac{\mathrm d Q^*_{\mathrm{path}}}{\mathrm d P_{\mathrm{path}}}
=\frac{1}{Z}\exp\!\big(-\phi(x_T)\big),
\qquad
Z:=\mathbb E_{P_{\mathrm{path}}}\!\big[\exp(-\phi(x_T))\big].
\label{eq:gibbs_path_socs_app}
\end{equation}
Taking the terminal projection yields the optimal terminal marginal
\begin{equation}
q_T^*(x)=\frac{1}{Z}\,p_T(x)\exp\!\big(-\phi(x)\big)
\;\propto\;
p_T(x)\exp\!\Big(-\frac{\gamma}{2}\|y-\mathcal H(x)\|_2^2\Big).
\label{eq:gibbs_terminal_socs_app}
\end{equation}
If the measurement noise is Gaussian, $y=\mathcal H(x_T)+\varepsilon$ with $\varepsilon\sim \mathcal N(0,\sigma^2 I)$, then
$p(y|x)\propto \exp\!\big(-\frac{1}{2\sigma^2}\|y-\mathcal H(x)\|_2^2\big)$.
Choosing $\gamma=1/\sigma^2$ (or absorbing constants into $\gamma$) makes the SOC reweighting $\exp(-\phi(x))$ exactly match the Gaussian likelihood, hence
$q_T^*(x)\propto p_T(x)\,p(y|x)=p(x_T\!\mid y)$.
In practice, $\gamma$ can be interpreted as an effective noise precision that controls the prior--likelihood trade-off. 
More generally, when $\gamma\neq 1/\sigma^2$, the optimal terminal marginal corresponds to a tempered posterior:
\begin{equation}
q_{T,\gamma}^*(x)\propto p_T(x)\exp\!\Big(-\frac{\gamma}{2}\|y-\mathcal H(x)\|_2^2\Big)
= p_T(x)\,p(y|x)^{\beta},\qquad \beta:=\gamma\sigma^2.
\end{equation}
Equivalently, this is the exact Bayesian posterior under an \emph{effective} Gaussian noise level
$\sigma_{\mathrm{eff}}^2 = 1/\gamma$ (up to constant factors absorbed into $\gamma$).
When $\gamma > 1/\sigma^2$ (i.e., $\beta>1$), the likelihood is sharpened and samples concentrate more tightly around the measurement-consistent set, potentially reducing diversity and over-emphasizing data fidelity under model mismatch.
When $\gamma < 1/\sigma^2$ (i.e., $\beta<1$), the likelihood is tempered and samples stay closer to the diffusion prior, which increases diversity but weakens measurement consistency.
In practice, the MCMC refinement is implemented on a preconditioned/normal-equation energy and uses an explicit likelihood “temperature” (e.g., $r_t$ in our Langevin update), so $\gamma$ should be interpreted as an \emph{effective} precision $\gamma_{\mathrm{eff}}$ after absorbing such implementation-dependent scalings; this helps explain why a broad range of numerical $\gamma$ values yields similar performance.
\clearpage

\section{Stochastic Optimal Control}
Stochastic optimal control has been widely applied in robot control~\cite{todorov2005generalized}, financial engineering~\cite{merton1975optimum}, epidemiology~\cite{behncke2000optimal},  communication networks~\cite{neely2010stochastic} and so on. A common formulation is:
\begin{equation}
\begin{aligned}
& \min _{u_t \in \mathcal{U}} \mathbb{E}\left[\int_0^T L\left(x_t, u_t, t\right) \mathrm{d} t+\Phi\left(x_T\right)\right] \\
& \text { s.t. } \mathrm{d} x_t=\big(f\left(x_t, t\right)+G\left(x_t, t\right) u_t\big)\mathrm{~d} t+\Sigma\left(x_t, t\right) \mathrm{d} w_t.
\end{aligned}
\end{equation}
Here $x_t$ denotes the system state and $u_t\in\mathcal U$ the control input. The function $f(x_t,t)$ specifies the drift of the uncontrolled dynamics, $G(x_t,t)u_t$ maps the control to its effect on the state, and $\Sigma(x_t,t) \mathrm dw_t$ injects stochastic disturbances with $w_t$ a standard Wiener process. The running cost $L(x_t,u_t,t)$ balances tracking performance and control effort over time, while the terminal cost $\Phi(x_T)$ encodes the final objective at time $T$. The expectation $\mathbb E[\cdot]$ averages over process noise and any randomness induced by the policy.

The core idea of SOC is to seek a feedback policy $u_t=\mu(t,x_t)$ that minimizes the expected cumulative cost while remaining robust to uncertainty. This principle admits a value function $V(t,x)$ governed by a dynamic programming equation, which yields a pointwise optimal feedback law. In the linear–quadratic case the value is quadratic and the feedback is given by a Riccati equation, where the state penalty shapes accuracy and the control penalty regularizes input magnitude and ensures well posedness.

In practice, SOC problems typically relies on a few classical paradigms. For low-dimensional systems, dynamic programming leads to the Hamilton–Jacobi–Bellman (HJB) equation for the value function, but solving this nonlinear PDE quickly becomes intractable in high dimensions. This motivates approximate schemes such as iterative linear–quadratic regulators (iLQR) and model predictive control, which repeatedly solve local LQ subproblems along a nominal trajectory. Another line of work instead optimizes a parametrized control sequence or policy by gradient-based trajectory optimization or reinforcement learning, trading exact optimality for scalability.

In this work, we encode the inverse problem directly in the terminal cost $\Phi(x_T)$ and, in parallel, regulate the amplitude of information injection along the trajectory. We approximate the degradation operator $\mathcal H$ either as linear or via a Gauss–Newton linearization around a clean-image estimate, so that $\mathbb{E}\lVert y-\mathcal H(x_T)\rVert^2$ becomes quadratic. Under this quadratic structure, the certainty-equivalence principle permits reducing the SDE formulation to an ODE.

\section{SOCS-VP-SDE Details}\label{sec:Experimental_details}
\begin{table}[htbp]
\centering
\caption{Sampling hyperparameters for different tasks on FFHQ and ImageNet.}
\label{tab:hparams_tasks}
\renewcommand{\arraystretch}{1.05}
\setlength{\tabcolsep}{4pt}
\small
\resizebox{\linewidth}{!}{%
\begin{tabular}{@{}l l | ccccccccc @{}}
\midrule
\multicolumn{2}{c|}{} &
SR & \shortstack[c]{Inpaint\\(Box)} & \shortstack[c]{Inpaint\\(Random)} & \shortstack[c]{Inpaint\\ \&SR}
& \shortstack[c]{Gaussian\\deblurring} & \shortstack[c]{Motion\\deblurring}
& \shortstack[c]{Phase\\retrieval} & \shortstack[c]{Nonlinear\\deblurring}
& \shortstack[c]{High dynamic\\range} \\
\midrule
\multirow{3}{*}{$\eta_0$}
  & nonlinear-$\gamma$(1e5) &5e-6  &5e-7  &5e-7 &5e-6  &5e-6  &5e-6  &5e-6  &5e-7  &2e-8  \\
  & nonlinear-$\gamma$(1e7) &5e-8  &5e-9  &5e-9 &5e-8  &5e-8  &5e-8  &5e-8  &5e-9  &2e-10  \\
  & nonlinear-$\gamma$(1e9) &5e-10  &5e-11  &5e-11 &5e-10  &5e-10  &5e-10  &5e-11  &3e-10  &2e-12  \\
  & nonlinear-$\infty$ &5e-5  &5e-5  &5e-5 &5e-5  &5e-5  &5e-5  &5e-5  &5e-5  &7e-5  \\
\multicolumn{2}{c|}{$\delta$} & 1e-2 & 1e-2 & 1e-2 & 1e-2 & 1e-2 & 1e-2 & 1e-2 & 1e-2 & 1e-2 \\
\multicolumn{2}{c|}{$N$}      & 150  & 100  & 100  & 100  & 100  & 100  & 150  & 500  & 300  \\
\multicolumn{2}{c|}{$p$}      & 2    & 2    & 2    & 2    & 2    & 2    & 2    & 2    & 2    \\
\midrule
\end{tabular}%
}
\end{table}
\subsection{Inverse Problem Setup}\label{sec:inverse_problem_setup}
Our nonlinear variant relies only on the forward degradation operator $\mathcal{H}(\cdot)$, following the implementation of~\cite{DPS}. The linear variant additionally requires a pseudoinverse $\mathcal{H}^{\dagger}(\cdot)$. For super-resolution, we implement $\mathcal{H}_{\text{SR}}^{\dagger}$ as an upsampling operator that restores the original resolution. For inpainting, the pseudoinverse coincides with the forward operator. For the combined inpainting and super-resolution setting, we use the composition $\mathcal{H}^{\dagger}_{\text{Inpaint}}\big(\mathcal{H}^{\dagger}_{\text{SR}}(\cdot)\big)$. Phase retrieval does not admit a corresponding inverse that maps magnitude measurements back to the pixel domain, which is why the linear variant is not applicable to this task. For the remaining tasks, we treat the input as the output of the inverse operator.

\subsection{SOCS Implementation Details}
SOCS addresses inverse problems in pixel space. Its linear-SDE pre-trained model learns at each step to estimate the score $s_{\theta}(x_t,\sigma_t) \approx \nabla_x \log p(x_t;\sigma_t)$, where $\sigma_t$ denotes a predefined noise level ranging from $\sigma_0$ to $\sigma_{\max}$. We further detail the procedure in Algorithm~\ref{algo:SOCS_algorithmic}. The time grid $(t_i)_{i=0}^{N}$ depends on the denoising schedule, and in our experiments we adopt a schedule that allows us to flexibly control the sampling density of the denoising time steps.

\noindent\textbf{Euler ODE Solver.} To propagate samples under the unconditional diffusion prior, we adopt the VE formulation and parameterize a forward noising process by the SDE
$\mathrm{d}x_t = \sqrt{2 \dot{\sigma}_t \sigma_t} \,\mathrm{d}w_t$, where $\sigma_t$ is a monotonically increasing noise scale, $\dot{\sigma}_t$ denotes its time derivative, and $w_t$ is a standard Wiener process. This SDE transports the clean data distribution $p(x_0)$ into a family of noisy marginals $p(x_t; \sigma_t)$ indexed by the noise level $\sigma_t$. As established in the score-based generative modeling literature~\cite{EDM,song2021scorebasedgenerativemodelingstochastic}, there exists a deterministic probability-flow ODE whose solution shares the same time-marginals as the SDE:
\begin{equation}
    \mathrm{d}x_t=-\dot{\sigma}_t\sigma_t\nabla_{x_t}\log p(x_t;\sigma_t)dt. 
\end{equation}

Following the preconditioning strategy in~\cite{FFHQ}, the pre-trained diffusion model with parameters $\theta$ is interpreted as providing a noise-conditioned score approximation $s_\theta(x_t, \sigma_t) \approx \nabla_{x_t} \log p(x_t; \sigma_t)$, which allows us to approximate the probability-flow vector field without explicitly modeling $p(x_t; \sigma_t)$. Within SOCS, we use this score field to define a deterministic predictor for the clean latent, and obtain $\hat{x}_T(x_t)$ by numerically integrating the above ODE with initial condition $(x_t, t)$.

In the practical implementation, at each denoising step we first predict $x_0$ using Eq.~\eqref{eq:pre_x0_when_xt} and then add noise back so that the result matches the initial noise level required by the ODE sampler:
\begin{equation}
\begin{aligned}
    x_{0 \mid t}=\frac{x_t-\sqrt{1-\alpha_t} \epsilon_\theta\left(x_t, t\right)}{\sqrt{\alpha_t}}.
\end{aligned}\label{eq:pre_x0_when_xt}
\end{equation}
We then set the schedule to $\sigma_t = t$ and apply a few-step Euler solver~\cite{EDM}, which evaluates $\frac{\mathrm{d}x_t}{dt}$ at $N_{ode}$ discretized time steps in the interval $[0,t]$ and updates $x_t$ according to the discretized ODE. The time steps $t_j, j=1,\cdots,N_{ode}$, and the value of $N_{ode}$ are chosen by a polynomial interpolation between $t$ and $t_{\min}$:
\begin{equation}
    t_j=\left(t^{\frac{1}{\rho}}+\dfrac{j}{N-1}\big(t_{min}^{\frac{1}{\rho}}-t^{\frac{1}{\rho}}\big)\right)^{\rho}.
\end{equation}

We use $\rho=7$ and $t_{min}=0.02$ throughout all experiments. 

\noindent\textbf{Denoising scheduler.} We propose the following denoising schedule to generate the noise levels $\sigma_t = t_k$ required by the ODE sampler:
\begin{equation}
    \begin{aligned}
        t_k = \text{round}\big((1-\frac{k}{N-1})^p(T-1)\big),\quad k=0,\cdots,N-1.
    \end{aligned}\label{eq:denoising_scheduler}
\end{equation}
Here $N$ denotes the target number of denoising steps, $T$ is the total number of training steps, and $p$ acts as a tunable hyperparameter that governs the step allocation along the trajectory. When $p=1$, the schedule collapses to a linear assignment of steps. For $p>1$, the discretization places a larger fraction of steps in the low-noise regime, enabling denser refinement of high-frequency details. Moreover, by exploiting a closed-form solution derived from stochastic optimal control, we map the predicted clean image back onto the noise manifold. This projection is agnostic to the specific discretization and therefore applies to any integer-valued denoising schedule.

\noindent\textbf{Hyperparameters Overview.} The hyperparameters of SOCS can be grouped into the following three categories, and the complete configuration is summarized in Table~\ref{tab:hparams_tasks}:

\begin{enumerate}[label=(\arabic*)]
  \item The ODE solver steps $N_{\text{ode}}$ and denoising steps $N_D$, these two quantities determine the total NFE of SOCS and therefore govern the trade-off between computational cost and reconstruction quality. In our experiments, we set $N_{\text{ode}} = 4$ and $N_D = 250$ for linear tasks, while for nonlinear tasks we use $N_{\text{ode}} = 8$ and $N_D = 500$.
  \item The step size $\eta_t$ and the total number of Langevin steps $N$ together control how we sample from the denoising process. In our implementation, we use a linearly decaying step-size schedule, $\eta_t = \eta_0\left(\delta + \frac{t}{T}(1-\delta)\right)$, where $\delta$ specifies the decay rate and $T$ denotes the starting time index. We implement Eq.~\eqref{eq:MCMC-sampling} in practice as
\begin{equation}
x_T^{j+1}
= x_T^{j}-\eta\nabla_{x_T^{j}}\left(\frac{\big\lVert \mathcal{H}(x_T^{j})-y\big\rVert^{2}}{2r_t^{2}}\right)+\sqrt{2\eta}\epsilon_j.
\label{eq:MCMC-sampling(Appendix)}
\end{equation}
Here $r_t$ is the annealed measurement noise level at step $t$. It plays the role of a likelihood “temperature,” scaling the data-fidelity term and thus controlling the balance between fitting the measurement and following the prior. In our experiments we set $r_t=0.01$ for the nonlinear-$\infty$ setting and $r_t=1$ for the nonlinear-$\gamma$ tasks. Consequently, we decrease the initial step size $\eta_0$ as $\gamma$ grows so that the optimization remains stable and more effective under the stronger data term.
  \item The hyperparameter $p$ controls the denoising density along the sampling trajectory, and we set $p=2$ by default in all experiments.
\end{enumerate}

\subsection{Baseline Details}
\noindent\textbf{Baselines on VP-SDE backbones.}
\begin{itemize}
  \item \textbf{DPS} All experiments are conducted with the original code and default settings as specified in \cite{DPS}. For the high dynamic range reconstruction task, we set $\epsilon_i = 1/\lVert y-H(\hat{x}_0(x_t))\rVert$.
  \item \textbf{DDNM} We adopt the default setting of $\eta_B=1.0$ and $\eta=0.85$ with 100 DDIM steps as specified in \cite{DDNM}.
  \item \textbf{DDRM} We use the default setting of $\eta_B=1.0$ and $\eta=0.85$ with 20 DDIM steps following \cite{DDRM}.
  \item \textbf{DCDP} We follow the default setting in \cite{DCDP} and directly use the open-sourced code for all results.
  \item \textbf{FPS-SMC} We adopt the default setting of $M=20$ and $N=1000$ as specified in \cite{FPS-SMC}.
  \item \textbf{Diff-PIR} We use the hyperparameters from \cite{DiffPIR} and set $\mathrm{NFE}=20$ for both FFHQ and ImageNet experiments.
  \item \textbf{RED-diff} For all experiments, we set $\lambda=0.25$ and $lr=0.5$ and use the other default settings in \cite{RED-Diff}.
  \item \textbf{DAPS} Following \cite{DAPS}, we use $N_{\mathrm{ODE}}=4$ and an annealing scheduler with $N_A=250$ for linear tasks, while for nonlinear tasks we use $N_{\mathrm{ODE}}=8$ and $N_A=500$.
  \item \textbf{DOC} We set $T=50$ for all experiments. For all tasks except SR$\times$4, we use $\text{num\_iters}=100$; for SR$\times$4 we use $\text{num\_iters}=50$. We directly employ the open-source implementation of \cite{solveIPviaDOC}.
\end{itemize}

\noindent\textbf{Baselines on SD3 (flow-matching) backbones.}
\begin{itemize}
  \item \textbf{PSLD} Following \cite{rout2023solving}, we set $\eta=1.0$ and $\gamma=0.1$, and run 200 NFEs.
  \item \textbf{LatentDAPS} We set $N_{\mathrm{ODE}}=5$ and $N_A=28$, resulting in a total NFE of 140. We further set the total number of Langevin steps to $N=50$, consistent with the default setting in \cite{DAPS}.
  \item \textbf{FlowChef} We adopt the protocol in \cite{Flowdps}, which selects hyperparameters via a grid search on 100 images. Accordingly, we set the step size to 200 for super-resolution tasks and 50 for deblurring tasks.
  \item \textbf{FlowDPS} For data-consistency optimization, we follow \cite{Flowdps} and apply 3 steps of gradient descent with step size 15 for all tasks.
\end{itemize}

\subsection{SOC-based inverse problem methods}
Among SOC-based approaches, DOC~\cite{solveIPviaDOC} is the most directly related work as it also tackles inverse problems in a training-free manner by casting diffusion sampling as a stochastic optimal control problem. In contrast, other SOC or diffusion-bridge formulations primarily address unconditional or conditional generation, rather than measurement-constrained inverse problems, and thus are complementary rather than directly comparable to our setting. Compared to DOC, which optimizes a trajectory-level control using iLQR and requires computing second-order information along the whole path, our method derives a closed-form, stepwise controller under a linear SDE with a quadratic terminal cost and applies it directly to the underlying sampler. This leads to a substantially simpler algorithm that can be plugged into existing diffusion samplers without second-order optimization, while still achieving competitive or better performance across a broad range of inverse problems.

\subsection{Approximations and Limitations}
In Sec.~3.2 we formulate a linear controlled VP-SDE with additive, control-independent Gaussian noise and a quadratic terminal cost, which places our derivation in the classical linear--quadratic--Gaussian (LQG) framework. Under these assumptions, the certainty-equivalence principle~\cite{chen2024generativemodelingphasestochastic,rout2024rbmodulationtrainingfreepersonalizationdiffusion} guarantees that the optimal feedback controller can be obtained by solving the corresponding deterministic ODE control problem; the Brownian noise does not change the form of the optimal policy. In practice, however, our diffusion prior is implemented by a highly nonlinear neural network and the measurement operator $\mathcal{H}$ can also be strongly nonlinear. We therefore treat the LQG formulation as a local surrogate model and apply the resulting closed-form controller to the actual VP-ODE used for sampling, after a Gauss--Newton-type linearization of $\mathcal{H}$ around the current iterate and a replacement of $\mathcal{H}$ by its Jacobian in the controller. Moreover, part of our analysis leverages $\gamma \to \infty$ limits and assumes that the associated matrices are well conditioned and invertible. Taken together, these modeling and linearization steps constitute the main approximations of SOCS, and the resulting algorithm should be viewed as an SOC-inspired approximate controller rather than an exact optimizer of the original stochastic control problem.
\begin{table}[!h]
    \centering
    \caption{\textbf{FID stability under different test-set sizes.} }
    \label{tab:fid_nums}
    \begin{tabular}{c|c|c|c|c}
        \hline
        Test images num & 100 & 200 & 500 & 1000 \\
        \hline
        Inpaint(Box) FID & 44.05$_{\pm 1.08}$ & 43.75$_{\pm 0.93}$ & 43.82$_{\pm 0.78}$ & 43.76$_{\pm 0.63}$ \\
        \hline
    \end{tabular}
\end{table}

Moreover, following DAPS~\cite{DAPS} and related work, we report FID scores computed on 100 validation images. This protocol enables a fair comparison with prior training-free methods, but FID estimates at such a small sample size are inherently noisy and should be interpreted with caution. To assess the stability of this evaluation, \cref{tab:fid_nums} reports the mean and standard deviation of FID across three random seeds when varying the number of test images from 100 to 1000, showing that the results are largely stable. In our experiments, the relative trends of PSNR/SSIM/LPIPS and FID are consistent, and our main conclusions on reconstruction quality are supported primarily by the reconstruction metrics rather than by FID alone.
\clearpage

\section{Experiments on Synthetic Data Distributions}
\label{sec:Toy_example}
Specifically, we consider a two-dimensional bimodal Gaussian mixture prior $p(x_0)=\frac{1}{2}\Big(\mathcal{N}(x_0;c_1,\Sigma_1)+\mathcal{N}(x_0;c_2,\Sigma_2)\Big),$
where \(c_1=(-0.3,-0.4)\) and \(c_2=(0.6,0.5)\), and \(\Sigma_1=\Sigma_2=\mathrm{diag}(0.01,0.04)\).
We randomly draw 2000 samples from the prior to construct a smoothed prior that varies with the noise level \(\sigma\).
This allows us to compute the Stein score function for any \(\sigma\), which plays a role similar to the pretrained score model \(s_\theta(x,\sigma)\) in practical diffusion models.
\begin{figure*}[htbp]
  \centering
  \includegraphics[width=\linewidth]{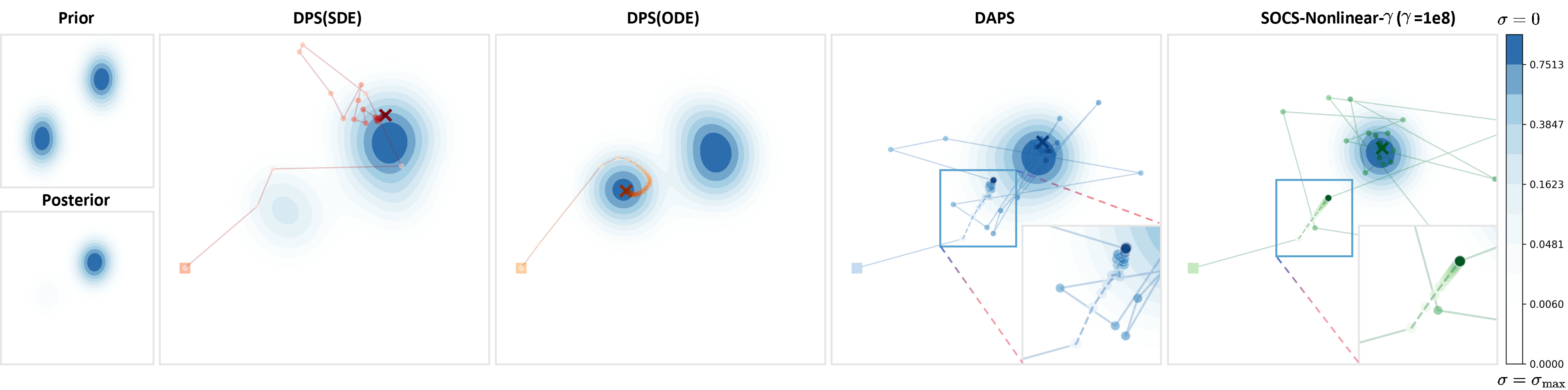}
  \caption{SOCS, DAPS, and DPS (both SDE and ODE variants) on two-dimensional synthetic data.}
  \label{fig:SOCS_toy_example}
\end{figure*}

We further design a simple nonlinear measurement model $
y=\exp\!\Big(-\tfrac{\|x\|^2}{0.5}\Big)+\exp\!\Big(-\tfrac{\|x-(0.5,0.5)\|^2}{0.5}\Big)+n,$ where \(n\sim\mathcal{N}(0,\beta_y^2)\) with \(\beta_y=0.3\), to match our inverse problem setting.
When \(y=1.50\), the prior is bimodal, yet the likelihood induced by the nonlinear measurement places most of its mass near \(m_2=(0.5,0.5)\).
As a result, the other prior mode is suppressed, and the posterior becomes effectively unimodal and is dominated by the mode near \((0.6,0.5)\).

We run all methods, including SOCS, DAPS, and DPS with both the SDE and ODE variants, for 1000 denoising steps and draw 100 synthetic samples for evaluation.
As shown in \cref{fig:SOCS_toy_example}, DPS (ODE) converges to an incorrect bimodal posterior in this setting.
DAPS allows larger per-step corrections through inner data-consistency refinement, which approximates the reference posterior better.
Building upon DAPS, SOCS further introduces a stochastic optimal control objective.
In \cref{fig:SOCS_toy_example}, we visualize the inner Langevin refinement during the first outer step for both DAPS and SOCS, where dashed segments connect intermediate iterates.
Compared to DAPS, the SOCS refinement path is less zigzagging and more direct, which better preserves the prior geometry.
\begin{figure}[t]
\centering
\begin{minipage}[t]{0.32\linewidth}
  \centering
  \includegraphics[width=0.82\linewidth]{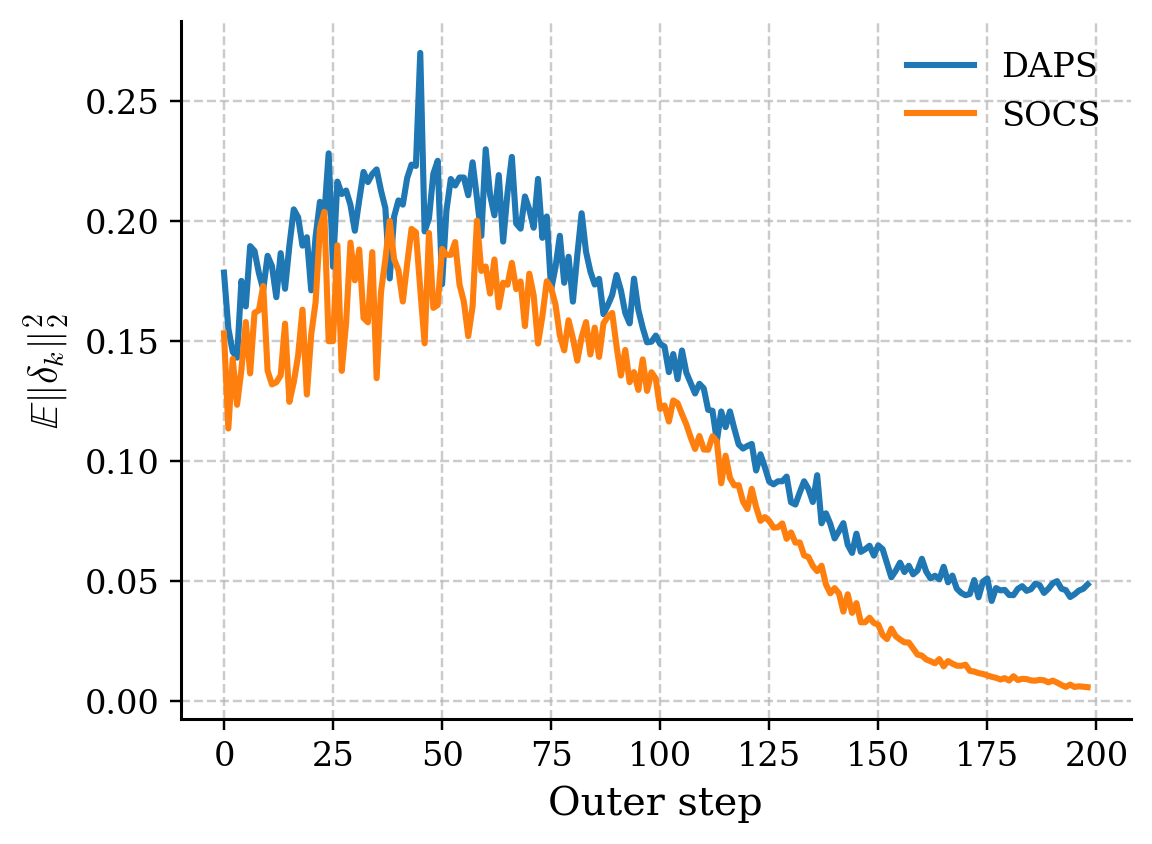}
  \captionof{figure}{Per-step correction magnitude over steps.}
  \label{fig:toy_steps}
\end{minipage}\hfill
\begin{minipage}[t]{0.66\linewidth}
  \centering
  \includegraphics[width=\linewidth]{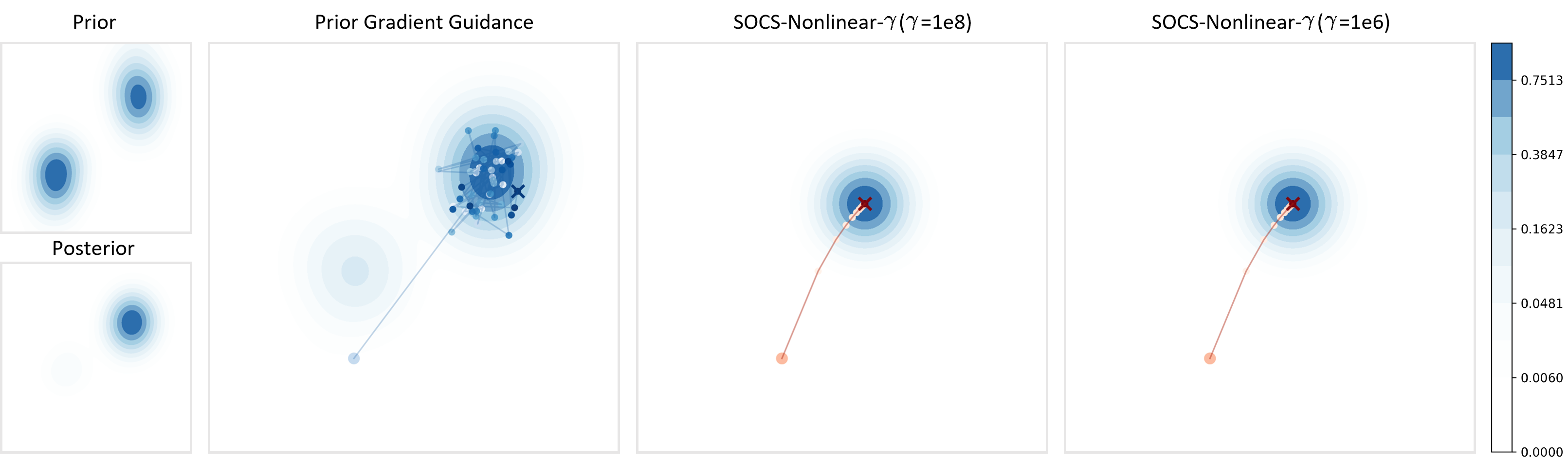}
  \captionof{figure}{\textbf{SOCS vs.\ prior-gradient guidance in Langevin refinement.} SOCS follows a straighter, lower-energy path to reach data consistency.}
  \label{fig:SOCS_mcmc_until_convergence}
\end{minipage}
\end{figure}
Figure~\ref{fig:control_pareto_sec3} reports the trade-off between data consistency and the deviation from the prior trajectory. To ensure a fair comparison and eliminate the influence of the Langevin step size $\eta_0$, we tune $\eta_0$ for SOCS so that the data error after the first Langevin refinement matches that of DAPS. SOCS consistently achieves a lower data error with a smaller deviation budget than DAPS. \cref{fig:toy_steps} also indicates that SOCS injects less control at each step. These results indicate that SOCS enforces measurement consistency more efficiently while perturbing the prior-generated trajectory less, which is consistent with the stochastic optimal control principle of achieving the target with minimal control energy.

In addition, we investigate the difference between using the standard measurement gradient $\nabla_{x_t}\lVert \mathcal{H}(x_t)-y\rVert_2^2$ in the Langevin refinement (\cref{eq:MCMC-sampling}) and using the $\gamma$-weighted objective $\bigl\lVert\bigl(I + \gamma(1-e^{2\bar{f}_T}) J_\mathcal{H}(x_T^u)^\top\bigr)\circ\mathcal{H}(\hat{x}_T^{j})
- \gamma J_\mathcal{H}(x_T^u)^\top\bigl(y-e^{\bar{f}_T}\mathcal{H}(x_0)\bigr)\bigr\rVert^2$ in \cref{eq:MCMC-sampling-gamma}.
Using the same 2D synthetic setup, we start from random samples drawn from the prior and perform direct data-consistency optimization via Langevin dynamics until convergence, once with each of the two gradient terms above. In \cref{fig:SOCS_mcmc_until_convergence}, we visualize the refinement trajectories of both methods. SOCS achieves measurement consistency with a smaller correction energy, resulting in a more direct refinement path. Moreover, the trajectory is largely insensitive to $\gamma$, which corroborates the robustness and wide-plateau behavior of $\gamma$ reported in \cref{sec:Additional_Discussion}.
\clearpage

\section{SOCS-VE-SDE}
\label{sec:SOCS-VE-SDE}
In the VE-SDE, the forward noising has zero drift and diffusion $g(t)$, i.e. $\mathrm{d}x_t = g(t)\mathrm{d}w_t $. As discussed in Sec.~\ref{sec:A Case: SOCS-VP-SDE}, we can likewise incorporate a terminal penalty within the VE–SDE formulation. Take the drift and diffusion term into Eq.~\eqref{eq:final_x_t} and Eq.~\eqref{eq:final_u_t} and donate $d_{0,\gamma}^{-1} = \left(I + \gamma \bar{g}_T^2J_\mathcal{H}(x_T^u)^\top \circ H\right)^{-1}$ :
    \begin{equation}
        \begin{aligned}
        &\Rightarrow x_t =x_0 + \gamma \bar{g}_t^2 d_{0,\gamma}^{-1}J_\mathcal{H}(x_T^u)^\top (y-H(x_0)), \\
        &\Rightarrow \mathbf{u}_{t,\gamma}^{\star} =-g_t\gamma d_{0,\gamma}^{-1}J_\mathcal{H}(x_T^u)^\top (y-H(x_0)).
        \end{aligned}\label{eq:VE-finite}
    \end{equation}
\noindent\textbf{SOCS-VE-Nonlinear-$\infty$.} When $\gamma \to \infty$, we simplify Eq.~\eqref{eq:VE-finite}:
\begin{equation}
\begingroup
\thinmuskip=2mu \medmuskip=2mu plus 1mu minus 1mu
\thickmuskip=3mu plus 1mu minus 1mu
\begin{aligned}
&\Rightarrow\; x_t
= x_0
  + \frac{\bar g_t^{\,2}}{\bar g_T^{\,2}}
    \bigl(J_\mathcal{H}(x_T^u)^{\top}\circ H\bigr)^{-1}
    J_\mathcal{H}(x_T^u)^{\top}\!\bigl(y-H(x_0)\bigr),
\\
&\Rightarrow\; \mathbf u_{t,\gamma}^{\star}
= -\,\frac{g_t^{\,2}}{\bar g_T^{\,2}}
    \bigl(J_\mathcal{H}(x_T^u)^{\top}\circ H\bigr)^{-1}
    J_\mathcal{H}(x_T^u)^{\top}\!\bigl(y-H(x_0)\bigr),
\\
&\Rightarrow\; \mathrm{d}x_t
= -\,\frac{g_t^{\,2}}{\bar g_T^{\,2}-\bar g_t^{\,2}}
    \bigl(J_\mathcal{H}(x_T^u)^{\top}\circ H\bigr)^{-1}
    J_\mathcal{H}(x_T^u)^{\top}\!\bigl(H(x_t)-y\bigr)\,dt
\\
&\quad\quad\quad\;\;= \frac{g_t^{\,2}}{\bar g_T^{\,2}}
   \bigl(J_\mathcal{H}(x_T^u)^{\top}\circ H\bigr)^{-1}
   J_\mathcal{H}(x_T^u)^{\top}\!\bigl(H(x_0)-y\bigr)\,dt.
\end{aligned}
\endgroup
\label{eq:VE-infinite}
\end{equation}

Starting from the VE–SDE, we model the forward perturbation of data by a zero-drift diffusion $\mathrm{d}x_t = g(t),\mathrm{d}w_t$, which induces a family of noisy marginals $p(x_t\mid \sigma_t)$ and, in reverse time (or equivalently via the probability-flow ODE), requires access to the score $\nabla_x \log p(x_t\mid \sigma_t)$ at each noise level. NCSN++~\cite{song2019generative} then serves as the noise-conditioned U-Net architecture that parameterizes this score, i.e., $s_\theta(x,\sigma)\approx \nabla_x \log p(x\mid \sigma).$ During training under the VE schedule, clean data samples are first corrupted by additive Gaussian noise with variance $\sigma^2$ drawn from a predefined noise ladder, and NCSN++ is optimized via denoising score matching to predict the score at each $\sigma$, thereby learning a multiscale approximation to the entire score field. At inference time, the same network provides $s_\theta(\cdot,\sigma_t)$ as a score oracle to numerically integrate either the VE reverse SDE or its probability-flow ODE, yielding a flexible unconditional diffusion prior that can subsequently be adapted to downstream inverse problems.

In the experiments of VE-SDE, we use publicly released VE–SDE checkpoints from \cite{song2021scorebasedgenerativemodelingstochastic}, trained with the NCSN++ architecture on FFHQ (256 px) and LSUN Church (256 px), respectively. We load the weights as provided, without any fine-tuning, and employ them solely as unconditional diffusion priors.
\begin{table}[t]
\captionsetup{font=small}
\centering
\begin{minipage}{0.75\textwidth}
\centering
\caption{Comparison of SOCS-VE-Nonlinear with previous methods on LSUN Church inpainting tasks.}
\label{tab:ve_linear_SOCSnonlinear}
\small
\setlength{\tabcolsep}{4.5pt}
\renewcommand{\arraystretch}{0.92}
\resizebox{\linewidth}{!}{
\begin{tabular}{@{}l l cccc@{}}
\toprule
\multicolumn{1}{c}{Task} & \multicolumn{1}{c}{Method} & PSNR $(\uparrow)$ & SSIM $(\uparrow)$ & LPIPS $(\downarrow)$ & FID $(\downarrow)$ \\
\midrule
\multirow{3}{*}{\shortstack[l]{Inpaint\\(Box)}}
  & Ours & \bfseries 23.17 & \bfseries 0.694 & \bfseries 0.276 & \bfseries 67.60 \\
  & MCG  & 21.14 & 0.532 & 0.356 & 84.91 \\
  & Score\-SDE & 22.34 & 0.545 & 0.384 & 106.28 \\
\cmidrule(lr){1-6}
\multirow{3}{*}{\shortstack[l]{Inpaint\\(Random)}}
  & Ours & \bfseries 28.44 & \bfseries 0.776 & \bfseries 0.245 & \bfseries 69.81 \\
  & MCG  & 25.37 & 0.543 & 0.357 & 89.59 \\
  & Score\-SDE & 26.28 & 0.584 & 0.332 & 83.14 \\
\bottomrule
\end{tabular}
}
\end{minipage}
\end{table}

When we adopt a linear approximation of the degradation operator, i.e., assume $J_\mathcal{H}(x_T^u) \approx H$, the expression for $x_t$ in Eq.~\eqref{eq:VE-infinite} can be written as follows:
\begin{equation}
\begin{aligned}
x_t = (I-\dfrac{\bar{g}_t^2}{\bar{g}_T^2})x_0 + \dfrac{\bar{g}_t^2}{\bar{g}_T^2}H^{\dagger}y, 
\end{aligned}\label{eq:VE-Linear}
\end{equation}
where $H^{\dagger}(\cdot)=(H^\top H)^{-1}\circ H^\top (\cdot)$. Eq.~\eqref{eq:VE-Linear} realizes a time-scheduled interpolation between the prior term and the measurement-consistent reconstruction. Early in reverse diffusion (high noise, $t\approx 0$) the weight $\bar g_t^2/\bar g_T^2\approx1$ anchors the trajectory near $x_0$ for robustness, while later (low noise) the controller shifts mass to $H^\dagger y$ (or its estimate) to recover high-frequency details. This yields a principled, stability-friendly bias–variance trade-off guided by data consistency first and the learned prior thereafter.

In practice, to obtain the per-step estimate of $H^{\dagger}(y)$, we implement a dynamic-step ODE sampler under the EDM framework with the poly-7 noise schedule to generate an initial image $x_T(x_t)$. Next, we refine this estimate with an MCMC scheme in the spirit of Eq.~\eqref{eq:MCMC-sampling}, enforcing measurement consistency so that $H^{\dagger}(y) \approx \hat{x}_T(x_t)$. Finally, using Eq.~\eqref{eq:VE-Linear}, we map $\hat{x}_T(x_t)$ back into the denoising space at the noise level required for the subsequent time step.

\begin{table}[htbp]
\centering
\caption{\textbf{Quantitative evaluation on FFHQ $256\times256$ of SOCS with Stable Diffusion v1.5.} The value shows the mean over 100 images, and all tasks assume the measurement noise level $\beta_y = 0.01$. Numbers for PSLD~\cite{rout2023solving} are copied from the original paper.}
\label{tab:quant_eval_sd15_small}
\renewcommand{\arraystretch}{1.1}
\setlength{\tabcolsep}{4pt}
\resizebox{0.88\textwidth}{!}{%
\begin{tabular}{@{}l|cc|cc|cc|cc@{}}
\toprule
Method &
\multicolumn{2}{c|}{Super Resolution 4$\times$} &
\multicolumn{2}{c|}{High dynamic range} &
\multicolumn{2}{c|}{Inpaint (Random)} & 
\multicolumn{2}{c}{Nonlinear deblurring} \\
\cmidrule(lr){2-3}\cmidrule(lr){4-5}\cmidrule(lr){6-7}\cmidrule(l){8-9}
& LPIPS$\downarrow$ & PSNR$\uparrow$
& LPIPS$\downarrow$ & PSNR$\uparrow$
& LPIPS$\downarrow$ & PSNR$\uparrow$
& LPIPS$\downarrow$ & PSNR$\uparrow$ \\
\midrule
SOCS (SD-v1.5, ours) &\textbf{0.137}  &29.86  &\textbf{0.189}  &24.15  &\textbf{0.080}  &\textbf{31.94}  &\textbf{0.139} &29.83  \\
LatentDAPS (SD-v1.5) &0.142 &\textbf{30.21} &0.193 &\textbf{24.40} &0.084 &31.80 &0.142 &\textbf{30.21} \\
PSLD (SD-v1.5)       & 0.201 & 30.73  &   --  &   --   & 0.096 & 30.31 &   --  &   --  \\
\bottomrule
\end{tabular}%
}
\end{table}
\section{SOCS with SD v1.5 Latent Diffusion Model}\label{app:SOCS with LDMs}
Stable Diffusion v1.5 (SD v1.5) is a large-scale text-conditioned latent diffusion model built upon the standard VP-SDE or DDPM-style formulation, where denoising is performed in a learned latent space instead of pixel space.
Latent Diffusion Models (LDMs)~\cite{rombach2022high} erform denoising in a learned, low-dimensional latent space rather than in pixel space. Working in this space reduces computation and memory, stabilizes optimization, and directs capacity toward semantically meaningful structure instead of raw texture. It also typically affords larger effective receptive fields at the same cost and induces smoother, more structured priors, leading to more stable gradients, stronger perceptual quality under a fixed sampling budget, and improved robustness to distribution shift. In this section, we focus on SD v1.5 as a representative LDM and show that our approach can extend naturally to LDMs.

\begin{figure*}[htbp]
  \centering
  \includegraphics[width=\textwidth]{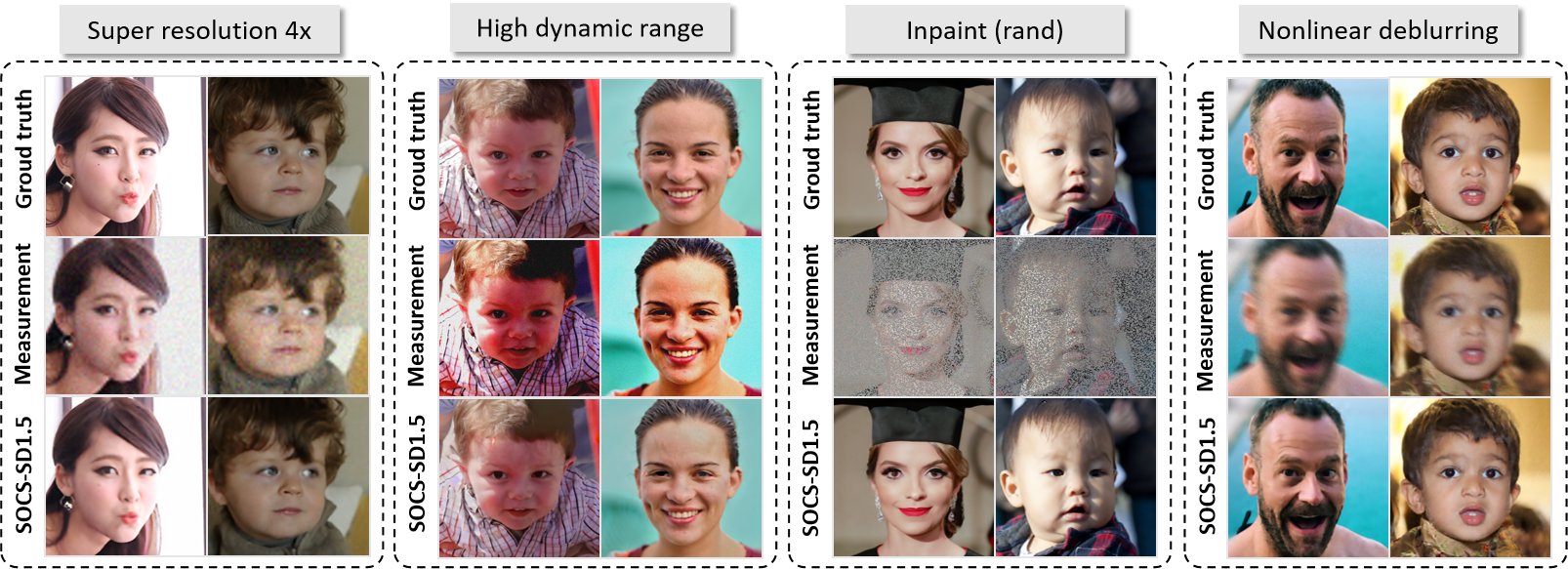}
  \caption{\textbf{Sampling results of SOCS-SD1.5 on FFHQ $\bf{256\times 256}$ images.} The sampling is enhanced with classifier-free guidance for text with guidance scale 7.5. The used text prompt is \text{"a natural looking human face"}.}
  \label{fig:SD1.5}
\end{figure*}

Let $\mathcal{E}:\mathbb{R}^n \rightarrow \mathbb{R}^k$ and $\mathcal{D}:\mathbb{R}^k \rightarrow \mathbb{R}^n$ denote the encoder and decoder of the LDM, respectively. We extend Eq.~\ref{eq:MCMC-sampling-gamma} to the latent space and perform the corresponding optimization over latent variables:
\begin{equation}
\begingroup
\thinmuskip=2mu \medmuskip=2mu plus 1mu minus 1mu
\thickmuskip=3mu plus 1mu minus 1mu
\begin{aligned}
z_T^{j+1}
&= z_T^{j}
   - \eta\,\nabla_{z_T^{j}}
   \bigl\|
      \bigl(I+\gamma(1-e^{2\bar f_T})\,J_{\mathcal H}(\mathcal D(z_T^u))^{\top}\bigr)
      \circ\,\mathcal H(\mathcal D(z_T^{j}))
\\[-1pt]
&\quad
   - \gamma\,J_{\mathcal H}(\mathcal D(z_T^u))^{\top}
     \bigl(y-e^{\bar f_T}\!\mathcal H(\mathcal D(z_0))\bigr)
   \bigr\|^{2}
   + \sqrt{2\eta}\,\epsilon_j.
\end{aligned}
\endgroup
\label{eq:MCMC-sampling-gamma(LDMs)}
\end{equation}
\begin{table}[htbp]
\centering
\caption{The text prompts used in SOCS (SD v1.5).}
\label{tab:text_prompts in SD1.5}
\renewcommand{\arraystretch}{1.15}
\begin{tabularx}{\linewidth}{@{}l X@{}}
\toprule
\multicolumn{2}{c}{\text{Text Prompts}}\\
\midrule
\text{FFHQ Evaluation} &
\textit{A natural looking human face.}\\
\addlinespace[2pt]
\text{\cref{fig:first_page}} &
\textit{A cupcake with white frosting and carrot shreds on a small plate, with rolled wafers beside it.}\\
& \textit{A yellow Labrador pants next to an orange frisbee on the grass.}\\
\text{\cref{fig:sr8_progress_sd_imagenet}} &
\textit{A pile of fresh vegetables on a kitchen counter}\\
\text{\cref{fig:hdr_progress_sd_imagenet}} &
\textit{polar bear on snow}\\
\bottomrule
\end{tabularx}
\end{table}
By focusing solely on the terminal loss with $\gamma \to \infty$, we obtain a simplified form of the above:
\begin{equation}
z_T^{j+1} = z_T^{j} - \eta  \nabla_{z_T^{j}} ||\mathcal{H}(\mathcal{D}(z_T^{j})) - y||^2 + \sqrt{2\eta} \epsilon_j.
\label{eq:MCMC-sampling(LDMs)}
\end{equation}
We summarize our method with LDMs in Algorithm~\ref{algo:SOCS_LDMs}.

Large-scale text-conditioned latent diffusion models (LDMs) guide sampling with a text prompt $c$ by encoding $c$ into embeddings that condition the denoiser throughout the reverse process. We adopt Stable Diffusion v1.5~\cite{rombach2022high} as the base model, whose VP-SDE formulation lets our method in Sec.~\ref{sec:A Case: SOCS-VP-SDE} apply directly following Algorithm~\ref{algo:SOCS_LDMs}. For evaluation we use 100 validation images from FFHQ at $256\times256$ and apply classifier-free guidance with scale 7.5 to strengthen conditioning. Qualitative results are shown in \cref{fig:SD1.5}, and quantitative metrics are reported in Table~\ref{tab:quant_eval_sd15_small}. Our method proves more effective at improving perceptual quality. Text-conditioned LDMs flexibly steer sampling via prompts, and Table~\ref{tab:text_prompts in SD1.5} lists all prompts used across tasks.
\begin{algorithm}[htbp]
\caption{SOCS (LDMs)}\label{algo:SOCS_LDMs}
\begin{algorithmic}[1]
\Require Latent space score model $s_\theta$, measurement $\mathbf y$, operator $\mathcal H$, time grid $(t_i)_{i=0}^{N}$, encoder $\mathcal{E}$ and decoder $\mathcal{D}$
\State Sample $\mathbf z_{t_0} \sim \mathcal N(\mathbf 0, I)$
\For{$i = 0$ \textbf{to} $N-1$}
  \State \begingroup
    \thinmuskip=2mu \medmuskip=2mu plus 1mu minus 1mu
    \thickmuskip=3mu plus 1mu minus 1mu
    $\hat{\mathbf{z}}_{T\mid t_i}\!\gets\!\mathbf z_{t_i}^{(T)}\!\bigl(\mathbf z_{t_i})\bigr)$
    \Comment{few step ODE sampling with $s_\theta$}
    \endgroup
 \For{$j = 0$ \textbf{to} $N_{\text{inner}}-1$}
     \State $\hat{\mathbf{z}}_{T|t_i}^{(j+1)} \gets \textsc{Refine}(\hat{\mathbf{z}}_{T|t_i}^{(j)})$   \Comment{Eq.~\eqref{eq:MCMC-sampling-gamma(LDMs)},Eq.~\eqref{eq:MCMC-sampling(LDMs)}}
 \EndFor
 \State $\hat{\mathbf{z}}_{T|t_i} \gets \hat{\mathbf{z}}_{T|t_i}^{(N_{\text{inner}})}$
  \State Project $\hat{\mathbf z}_{T|t_i}$ to SOC manifold, get $\hat{\mathbf{z}}_{t_{i+1}}$     \Comment{Eq.~\eqref{eq:SOC_x_t}}
\EndFor
\State $\mathbf z_N \gets \hat{\mathbf z}_N$
\State \Return  $\mathcal{D}(z_N)$
\end{algorithmic}
\end{algorithm}

\section{SOCS-FM: Extending SOCS to Flow-Matching Models}
\label{app:socs_fm}
Flow-matching (FM) models generate samples by integrating an ODE driven by a learned velocity field.
Let $z\in\mathbb{R}^k$ denote the latent variable and $D:\mathbb{R}^k\to\mathbb{R}^n$ the decoder.
Given a text prompt $c$, FM sampling starts from a Gaussian latent and follows the probability-flow ODE
\begin{equation}
    d z_t \;=\; v_\theta(z_t,t,c)d t, \qquad z_0 \sim \mathcal{N}(0,I), \quad t\in[0,T],
\end{equation}
and outputs the image $x_T = D(z_T)$. FM (rectified-flow) models can be interpreted through an idealized straight-line flow, where each trajectory is approximately linear and the corresponding velocity is (approximately) constant along that trajectory. Building on this structure, we formulate the same SOC problem as in LDMs and solve the inverse problem entirely in the latent space:
\begin{equation}
\begin{aligned}
\min_{\mathbf{u}_{t,\gamma}\in \mathcal{U}}\;& \int_0^T \tfrac{1}{2}\bigl\lVert\mathbf{u}_{t,\gamma}\bigr\rVert_2^2\,\mathrm dt
+ \tfrac{\gamma}{2}\bigl\lVert y-\mathcal{H}(D(z_T))\bigr\rVert_2^2 \\
\text{s.t.}\quad & \mathrm d z_t = (v_\theta(z_t,t,c) + \mathbf{u}_{t,\gamma})\,\mathrm dt.
\end{aligned}
\label{eq:SOC_construct_FM}
\end{equation}

Following the proof in Appendix~\ref{sec:Proof of Theorem 3.1}, we linearize the composite measurement map $h(z):=\mathcal{H}(D(z))$ around the nominal terminal prediction $\bar z_T = z_{\sigma} - \sigma\, v_\theta(z_{\sigma},\sigma,c)$:
\begin{equation}
h(z_T)\;\approx\; h(\bar z_T) + J\, (z_T-\bar z_T),
\qquad
J := \left.\frac{\partial h(z)}{\partial z}\right|_{z=\bar z_T}.
\end{equation}
With this local model, the optimal control is constant over $t\in[0,T]$ and admits a closed form:
\begin{equation}
u_{t,\gamma}^*
=\gamma\big(I+\gamma T\,J^\top J\big)^{-1}J^\top \big(y - h(\bar z_T)\big).
\label{eq:fm_u_finite}
\end{equation}
The corresponding controlled terminal correction is $\delta := z_T^*-\bar z_T = T u_{t,\gamma}^*$.
Using the FM straight-line geometry, this terminal correction induces a closed-form correction at intermediate time:
\begin{equation}
z_t^* \;=\; \bar z_t \;+\; t\,u_{t,\gamma}^*
\quad\Longleftrightarrow\quad
z_{\sigma}^* \;=\; z_{\sigma} \;+\; (1-\sigma)\,\delta,
\label{eq:fm_z_pullback}
\end{equation}
where $\bar z_t$ denotes the nominal FM state and in our implementation it is taken as the current iterate. In SD3 we use a normalized noise level $\sigma\in[0,1]$ and identify the time as $t=1-\sigma$ while the horizon-$T$ parametrization gives $t=(1-\sigma)T$. Finally, the reconstructed image is $x_t^* = D(z_t^)$ and $x_T^=D(z_T^*)$.

\noindent\textbf{SOCS-FM-Nonlinear-$\infty$.} Taking $\gamma\to\infty$ in Eq.~\eqref{eq:fm_u_finite} yields a minimum-norm solution enforcing the (linearized) terminal consistency:
\begin{equation}
u_{t,\infty}^*
=\frac{1}{T}\,(J^\top J)^{-1}J^\top \big(y - h(\bar z_T)\big)
\;=\;\frac{1}{T}\,J^\dagger \big(y - h(\bar z_T)\big),
\label{eq:fm_u_infty}
\end{equation}
where $J^\dagger$ is the pseudoinverse when $J^\top J$ is ill-conditioned.
The induced intermediate correction remains the same closed form:
\begin{equation}
    z_{\sigma}^* \;=\; z_{\sigma} + (1-\sigma)\,T\,u_{t,\infty}^*.
\end{equation}
In practice, $\gamma\to\infty$ corresponds to emphasizing data consistency and may pull samples away from the FM prior,
while a finite $\gamma$ provides a stable trade-off, consistent with our ablations. \cref{tab:text_prompts_in_SD3} lists the prompts used for our SD3 experiments.
\begin{table}[htbp]
\centering
\caption{The text prompts used in SOCS (SD3).}
\label{tab:text_prompts_in_SD3}
\renewcommand{\arraystretch}{1.15}
\begin{tabularx}{\linewidth}{@{}l X@{}}
\toprule
\multicolumn{2}{c}{\text{Text Prompts}}\\
\midrule
\text{FFHQ Evaluation} &
\textit{A photo of a closed face.}\\
\bottomrule
\end{tabularx}
\end{table}

\section{Additional Results} \label{sec:Additional_results}
\noindent\textbf{Sampling Efficiency.} Sampling efficiency in diffusion models has received considerable attention, as runtime is largely determined by the number of neural function evaluations (NFE). Here in \cref{tab:sampling_time} below reports the NFE used by several baselines, alongside our SOCS variants under different configurations and their corresponding per-image sampling time, peak memory and FLOPs. All methods that employ Langevin-dynamics refinement use the same optimization budget of 100 steps. Under the same NFE budget, our method runs in less than half the runtime and compute of \cite{solveIPviaDOC} while remaining comparable to sampling-based baselines.
\begin{table}[t]
\centering
\caption{\textbf{Sampling efficiency for FFHQ-256 phase retrieval.} We report non-parallel single-image runtime (sec/image), peak memory (GB), and FLOPs (TFLOP) measured on a single RTX 4090 24GB GPU. The number of function evaluations (NFE) and related solver settings are also listed. Runtimes may vary slightly across runs.}
\label{tab:sampling_time}
\small
\setlength{\tabcolsep}{2pt}
\resizebox{\linewidth}{!}{%
\begin{tabular}{@{}lcccccc@{}}
\toprule
Configuration & ODE Steps & Denoising Steps & NFE & Sec/Image & Peak Mem(GB) & FLOPs(TFLOP) \\
\midrule
DPS      & -- & --  & 1000 & 59 & 6.72 & 387.7\\ 
DAPS     &5   &200  & 1000 & 48 & 3.37 & 481.9\\  
RED-diff & -- & --  & 1000 & 47 & -- & --\\
\midrule
SOCS-Linear   &  4 &  100 &   400 & 10 & -- & -- \\
SOCS-Linear  &  4 &  250 &  1000 & 27 & 5.25 & 482.9\\
SOCS-Nonlinear  &  4 & 250 &  1000 & 60 & 3.37 & 481.4 \\
SOCS-Nonlinear  &  5 & 400 &  2000 & 103 & -- & --\\
SOCS-Nonlinear-$\gamma$   &  4 & 250 & 1000 & 68 & 3.94 & 512.4 \\
SOCS-Nonlinear-$\gamma$   &  8 & 250 & 2000 & 97 & -- & -- \\
SOCS-Nonlinear-$\gamma$   & 8 & 500 & 4000 & 192 & -- & --\\
\midrule
Configuration & T & Num Iters & NFE & Sec/Image & Peak Mem(GB) & FLOPs(TFLOP)\\
\midrule
DOC~\cite{solveIPviaDOC} & 20 & 50  & 1000 & 130 & 4.11 & 1147.2 \\
\bottomrule
\end{tabular}%
}
\end{table}

\begin{figure*}[htbp]
  \centering
  \includegraphics[width=\textwidth]{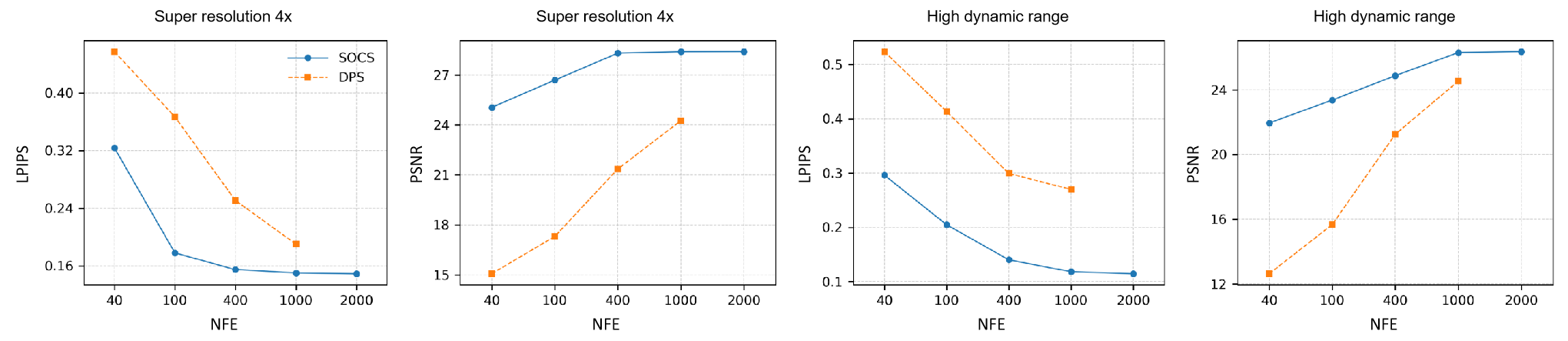}
  \caption{\textbf{Quantitative evaluations of image quality for different numbers of function evaluations (NFE).} Experiments are conducted on the FFHQ-256 dataset, covering one linear and one nonlinear task with SOCS-Nonlinear-$\infty$.}
  \label{fig:NFE_metrics}
\end{figure*}

\begin{figure*}[htbp]
  \centering
  \includegraphics[width=\textwidth]{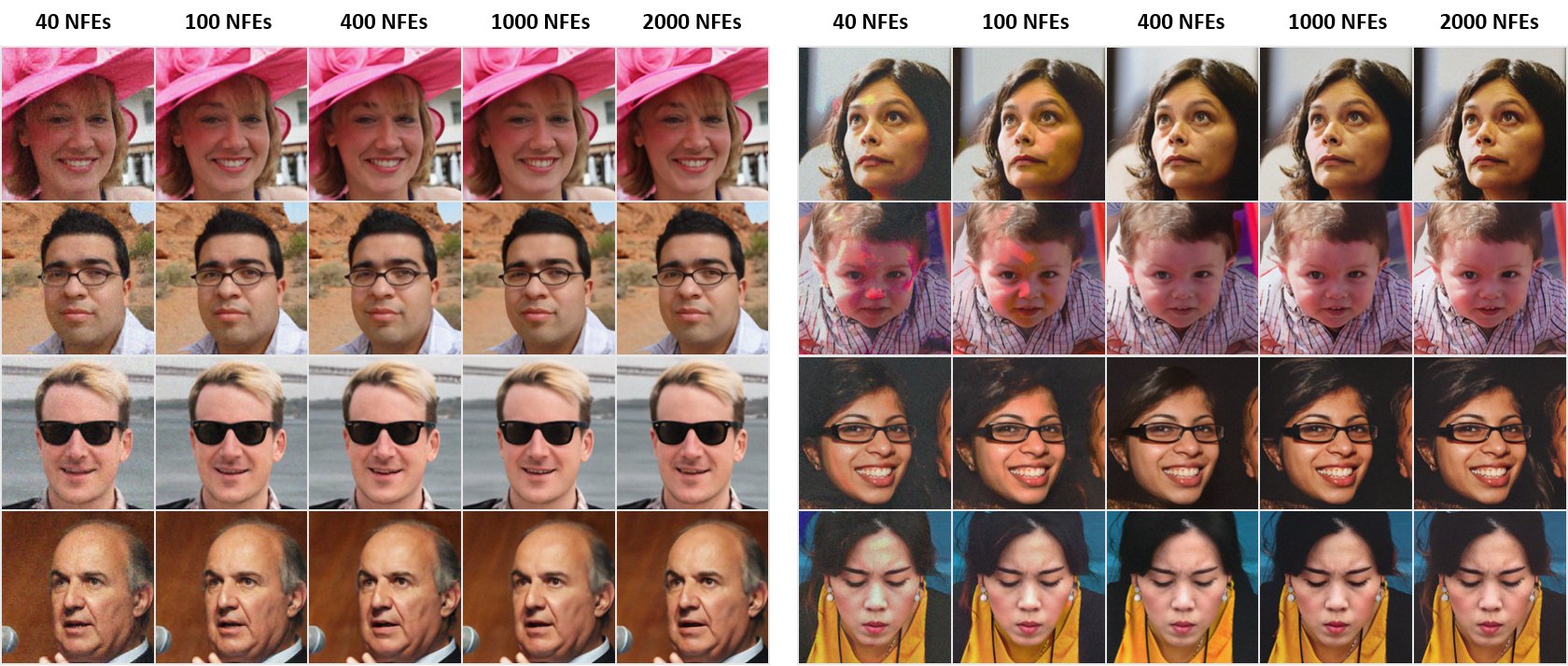}
  \caption{\textbf{Qualitative samples across different numbers of function evaluations (NFE).} The left panels show results on the $4\times$ super-resolution task, and the right panels correspond to the HDR reconstruction task. Image sharpness and perceptual details consistently improve as NFE increases.}
  \label{fig:NFE_show}
\end{figure*}

\noindent\textbf{Effectiveness of the number of function evaluations.}
To better assess the effect of the number of function evaluations (NFE) on generative capability, we conduct a systematic evaluation of SOCS under multiple configurations on one linear task and one nonlinear task, and compare it against the classical inverse-problem baseline DPS~\cite{DPS}. Recall that in the nonlinear method, each inner loop employs a few step ODE sampler to estimate $x_T(x_t)$. The total NFE equals the product of the discrete denoising steps and the ODE sampling steps. In our experiments, we fix the ODE sampling steps at 4 to isolate the influence of the denoising steps and vary NFE from 40 to 2000, covering both extremely few-step and moderately large-step regimes. As shown in \cref{fig:NFE_metrics}, quantitative image metrics improve rapidly at smaller numbers of denoising steps and exhibit diminishing returns once NFE reaches 1000. Compared to DPS, which is highly sensitive to the gradient step-size hyperparameter and can only extract limited information from the measurement $y$ at each time step, it typically requires a large number of steps to achieve good performance. 
In contrast, our method benefits from the efficient denoising scheduler and a decoupled denoising trajectory, which allows it to achieve strong results even with small NFE budget. We further provide qualitative samples in \cref{fig:NFE_show}, where image sharpness and perceptual quality improve progressively as the NFE increases, while SOCS remains relatively insensitive to exact NFE setting.
\begin{figure*}[htbp]
  \centering
  \includegraphics[width=\textwidth]{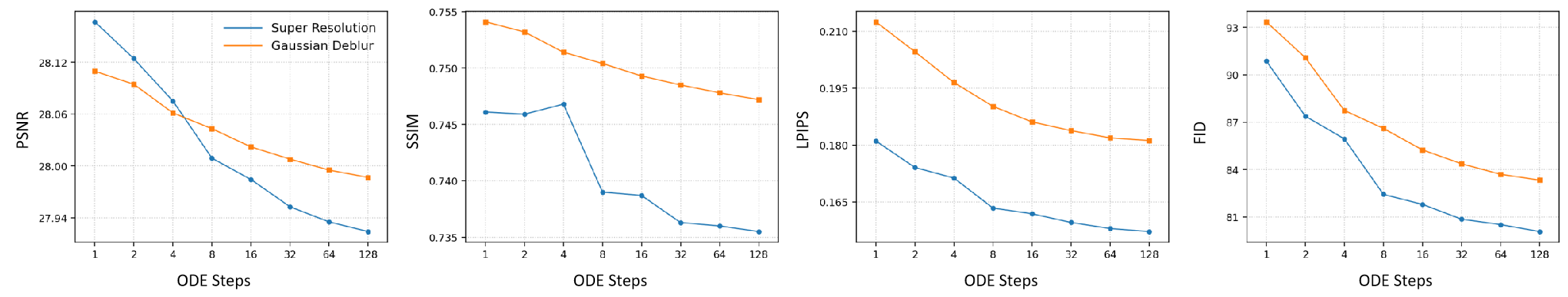}
  \caption{\textbf{Quantitative evaluations of image quality for different ODE steps.} We conduct experiments on the FFHQ-256 dataset across two tasks.}
  \label{fig:ODE_metric}
\end{figure*}
\begin{figure*}[htbp]
  \centering
  \includegraphics[width=\textwidth]{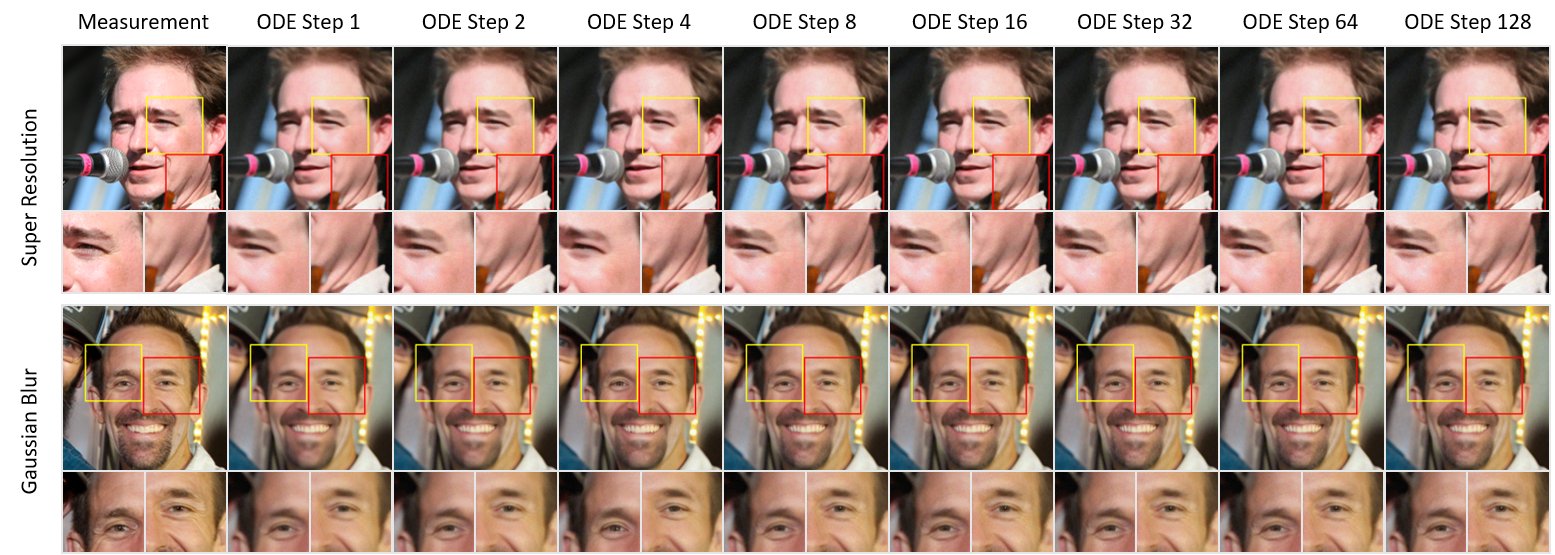}
  \caption{\textbf{Qualitative samples across different ODE steps.} With SOCS evaluated across multiple ODE sampling step counts, the results show progressively better detail restoration as the number of steps increases.}
  \label{fig:ODE_show}
\end{figure*}

\noindent\textbf{Effectiveness of the number of ODE steps.}
Recall that in Sec.~\ref{sec:method}, within the SOCS nonlinear method, we aim to better estimate:
\begingroup
\thinmuskip=2mu \medmuskip=2mu plus 1mu minus 1mu
\thickmuskip=3mu plus 1mu minus 1mu
\[
\gamma\bigl(I+\gamma(1-e^{2\bar f_T})J_\mathcal{H}(x_T^u)^{\top}H\bigr)^{-1}
J_\mathcal{H}(x_T^u)^{\top}\bigl(y-e^{\bar f_T}H(x_0)\bigr).
\]
\endgroup
We first employ a few-step ODE sampler to predict $x_T(x_t)$, which carries strong prior information from the model, and then enhance measurement consistency through Langevin dynamics. To understand how the number of function evaluations (NFE) in the ODE sampler affects both reconstruction fidelity and perceptual quality, we vary the ODE sampling steps while keeping the discrete denoising steps fixed. In particular, when $\text{NFE}=1$, the ODE sampler reduces to computing $\mathbb{E}[x_0 \mid x_t]$ via Tweedie’s formula. As shown in \cref{fig:ODE_metric}, increasing the number of ODE sampling steps consistently improves perceptual similarity, while PSNR and SSIM tend to degrade, which indicates a practical trade-off in selecting the step count. \cref{fig:ODE_show} presents qualitative samples for the two tasks, where we also observe a clear enhancement of fine details with larger numbers of steps.
\begin{figure*}[htbp]
  \centering
  \includegraphics[width=\textwidth]{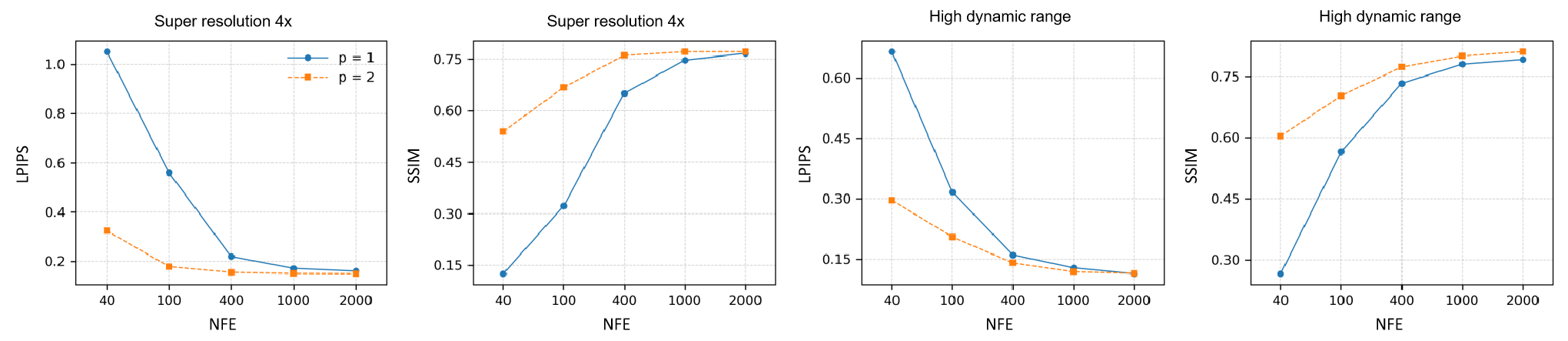}
  \caption{\textbf{Quantitative evaluations of image quality for different denoising schedulers.} At small NFE budgets, a denoising scheduler that concentrates more sampling steps in the low-noise regime can substantially improve generation quality.}
  \label{fig:power_compare}
\end{figure*}
\begin{figure}[t]
  \centering
  \includegraphics[width=0.90\linewidth]{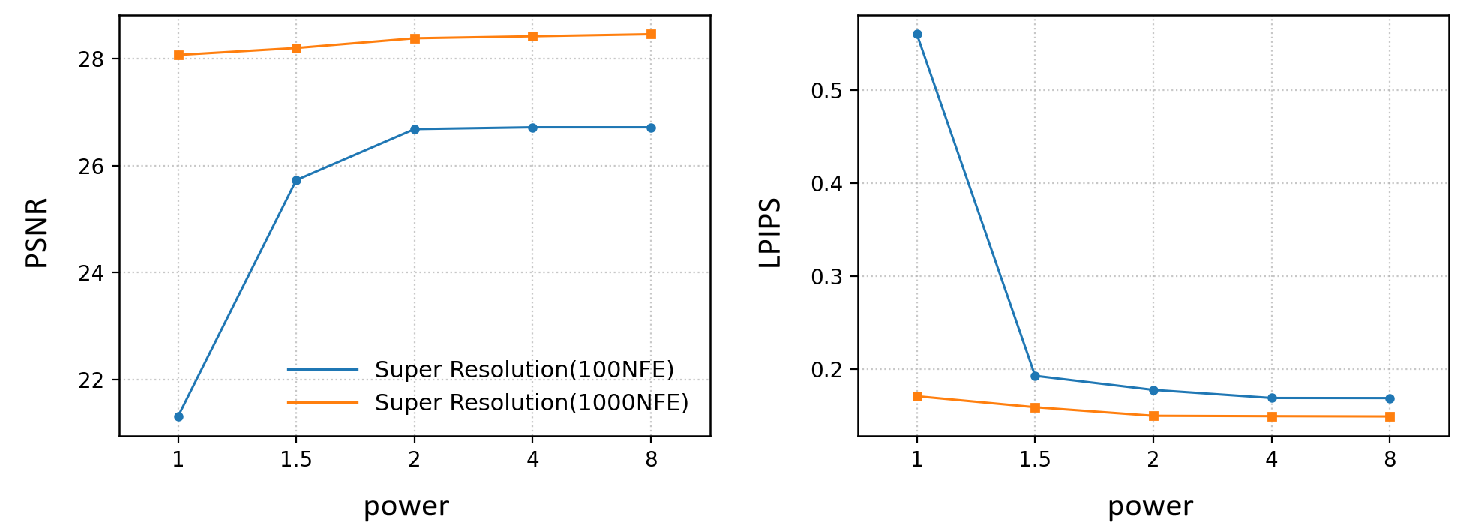}
  \caption{Image quality consistently improves with increasing $p$}
  \label{fig:Denoising_scheduler}
\end{figure}

\noindent\textbf{Effectiveness of the denoising schedule.}
We treat the parameter $p$ in the denoising scheduler of Eq.~\eqref{eq:denoising_scheduler} as a hyperparameter that controlsthe allocation and local density of denoising steps along the sampling trajectory. As shown quantitatively in \cref{fig:power_compare}, when the ODE sampling steps are fixed and the number of denoising steps is small with a linear denoising scheduler, generative performance degrades markedly. The underlying reason is that diffusion sampling follows a coarse-to-fine frequency progression in which the early high-noise stage reconstructs global structure and low-frequency content, while the later low-noise stage refines textures and high-frequency details. With very few denoising steps, the step sizes in the low-noise regime become too large to recover fine details, which results in visible noise artifacts in the images. 

We show in \cref{fig:Denoising_scheduler} that compared with a linear scheduler with $p=1$, densifying the denoising steps when the noise level is low (that is, using $p>1$) leads to a clear improvement in image quality and enables strong restoration even with small NFEs, and therefore we set $p=2$ in all experiments of this work to improve generative performance.

\subsection{More Qualitative Samples}\label{sec:more_Qualitative_Samples}
\begin{figure*}[t]  
  \centering
  \includegraphics[width=0.78\textwidth]{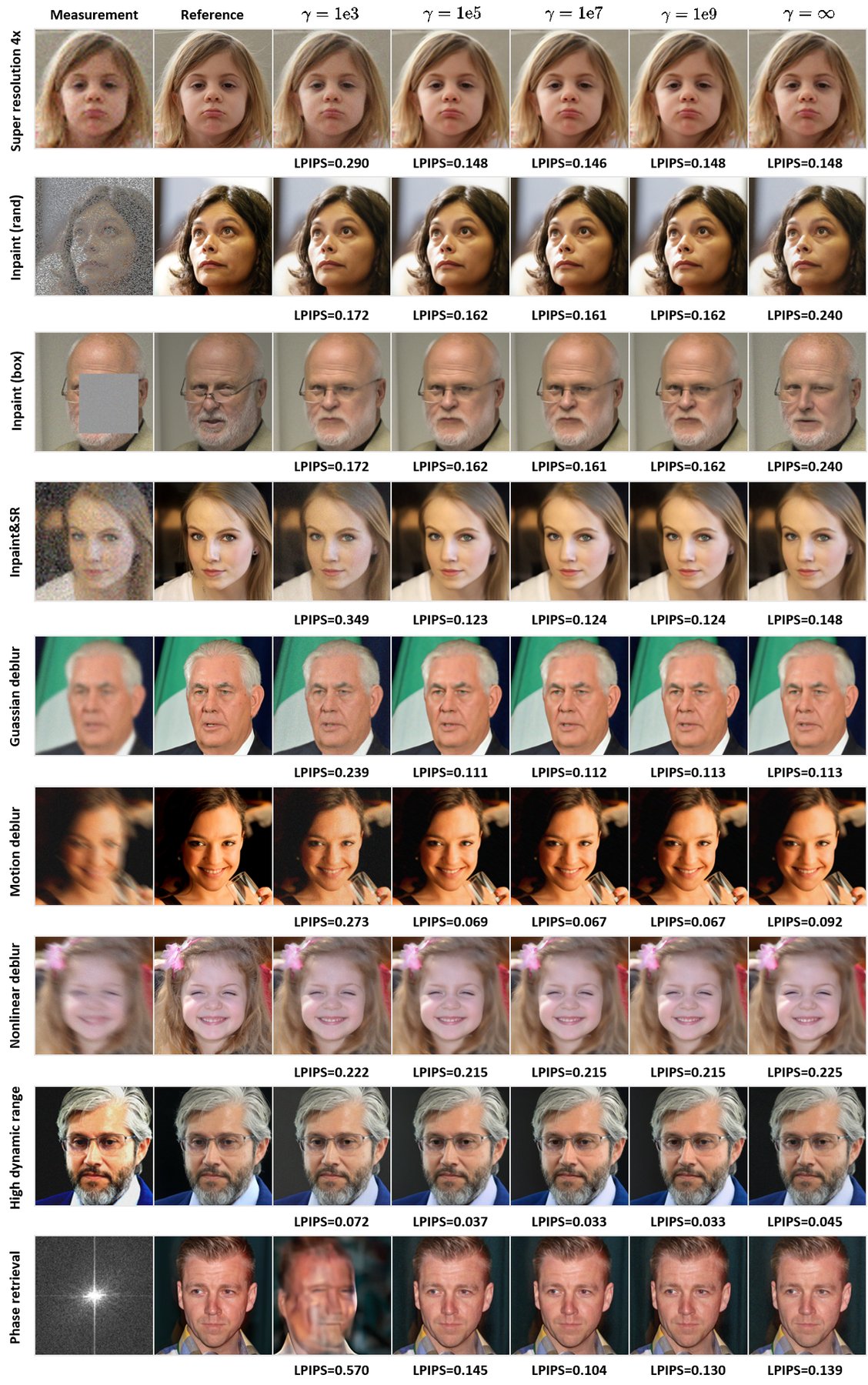}
  \caption{\textbf{Qualitative demonstration of the impact of $\gamma$ on image quality in all tasks.}}
  \label{fig:gamma_performance}
\end{figure*}
\begin{figure*}[t]
  \centering
  \includegraphics[width=0.90\textwidth]{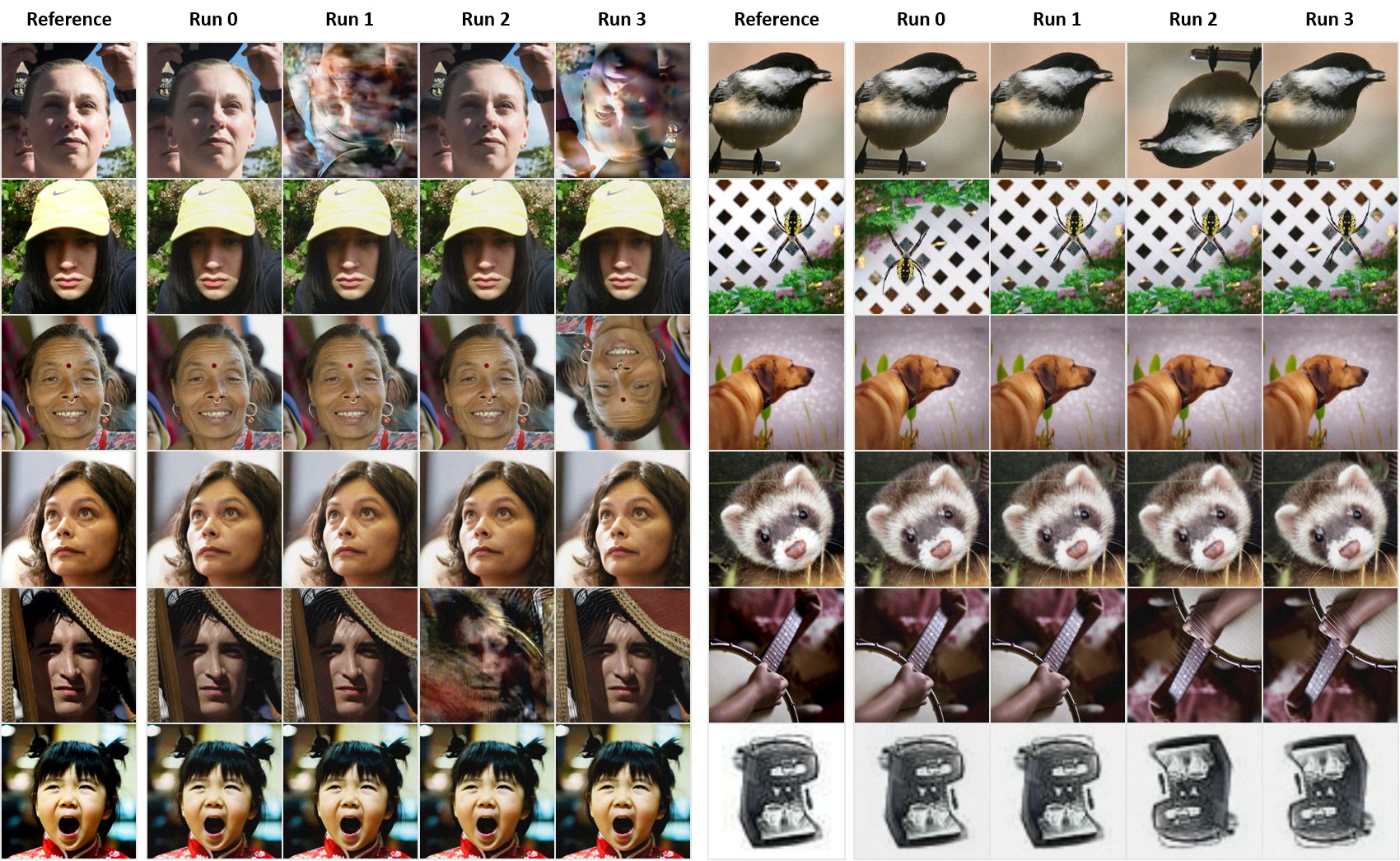}
  \caption{\textbf{Phase retrieval samples from SOCS across four independent runs on FFHQ (left) and ImageNet (right).} Image rotations by 180° appear in some samples on both datasets, and no manual selection was performed in order to more faithfully reflect success rates and sample quality.}
  \label{fig:phase_retrieval_show}
\end{figure*}
In \cref{fig:gamma_performance}, we show how the terminal penalty $\gamma$ affects the perceptual quality of a sampled FFHQ image across all tasks. We present selected phase retrieval samples on the FFHQ and ImageNet datasets in \cref{fig:phase_retrieval_show}. Some images appear rotated by 180 degrees on both datasets. This occurs because phase retrieval is more ill-posed than the other tasks. The measurement discards phase and retains only magnitude, which makes images with different rotation angles yield identical measurements under the degradation operator. The corresponding solutions are discrete and can differ completely, whereas the feasible solutions in other tasks vary continuously. 

\begin{figure*}[t]
  \centering
  \begin{subfigure}{0.85\textwidth}
    \centering
    \includegraphics[width=\linewidth]{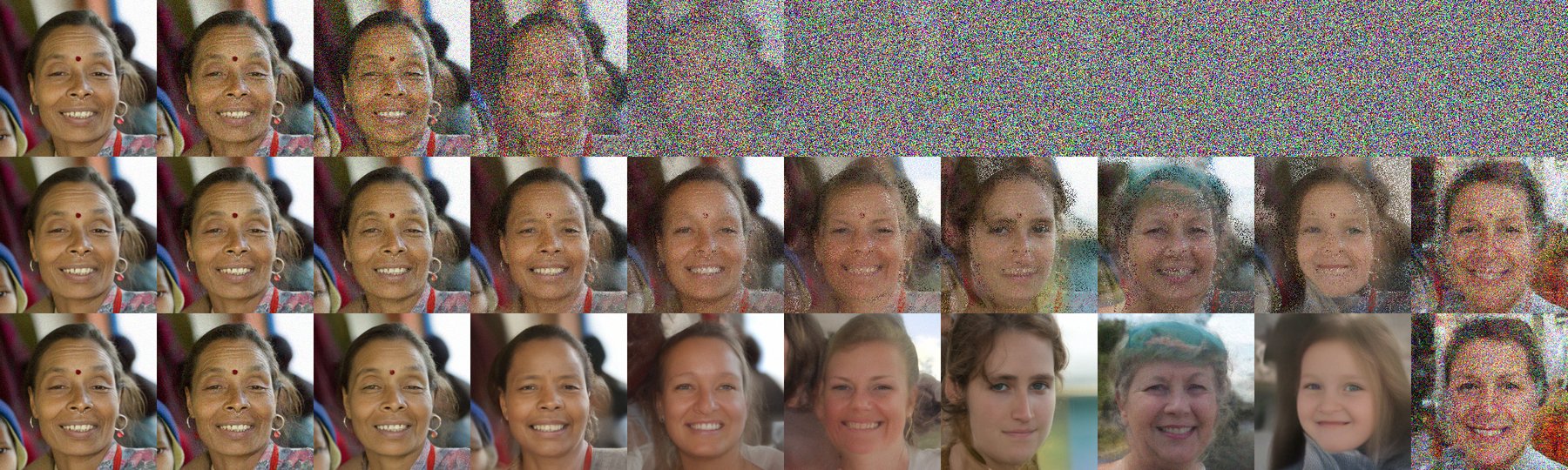}
    \caption{Denoising process for the Inpaint (Random) task with VE-SDE on the FFHQ dataset.}
    \label{fig:inpaintrandom_progress_ve_ffhq}
  \end{subfigure}\par\medskip

  \begin{subfigure}{0.85\textwidth}
    \centering
    \includegraphics[width=\linewidth]{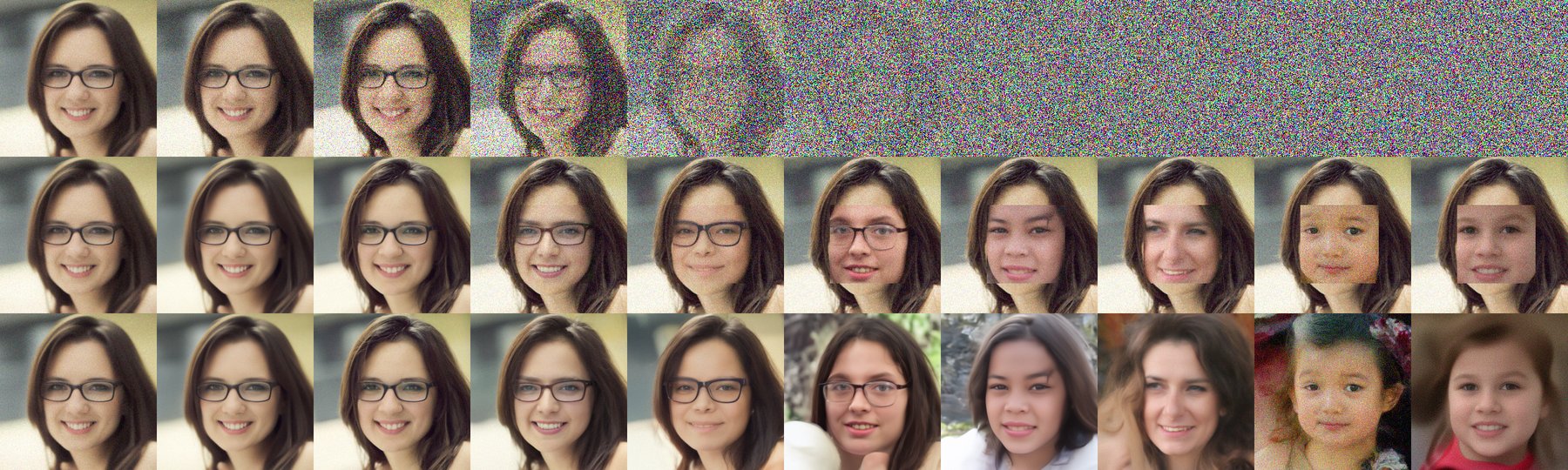}
    \caption{Denoising process for the Inpaint(Box) task with VE-SDE on the FFHQ dataset. }
    \label{fig:inpaintbox_progress_ve_ffhq}
  \end{subfigure}\par\medskip

  \begin{subfigure}{0.85\textwidth}
    \centering
    \includegraphics[width=\linewidth]{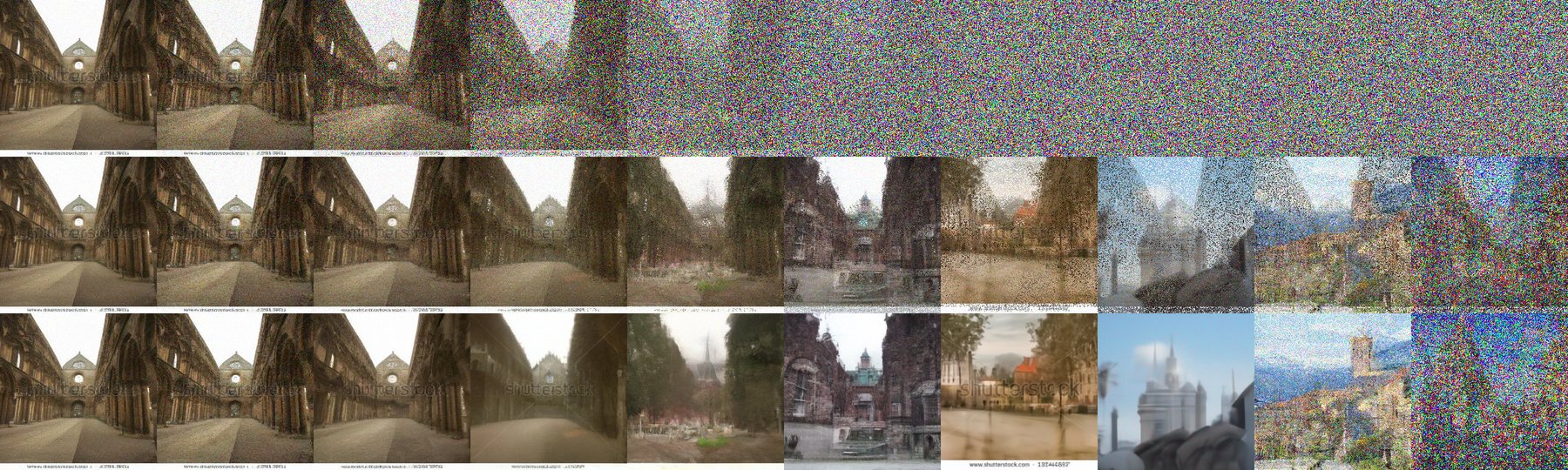}
    \caption{Denoising process for the Inpaint(Random) task with VE-SDE on the Church dataset.}
    \label{fig:inpaintrandom_progress_ve_church}
  \end{subfigure}\par\medskip

  \begin{subfigure}{0.85\textwidth}
    \centering
    \includegraphics[width=\linewidth]{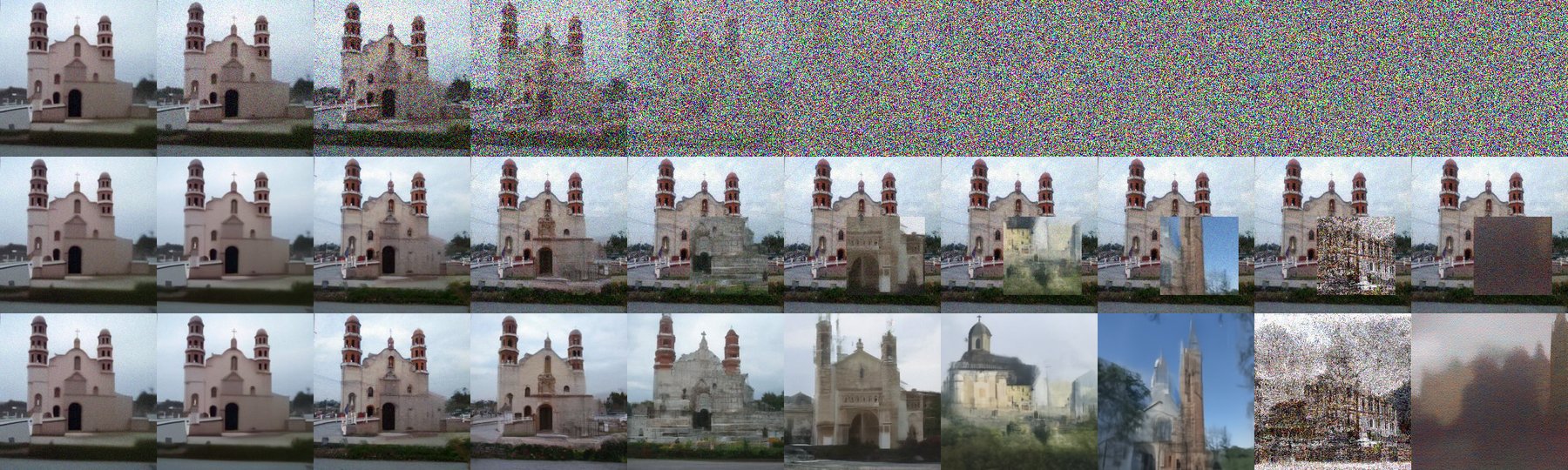}
    \caption{Denoising process for the Inpaint(Box) task with VE-SDE on the Church dataset. }
    \label{fig:inpaintbox_progress_ve_church}
  \end{subfigure}

  \caption{\textbf{Denoising processes for Inpaint tasks with VE-SDE}. For each task, we display a complete recovery trajectory starting from pure noise. From bottom to top, the rows show the few step ODE sampler estimate $x_{T|t}$, the Langevin dynamics refined $\hat{x}_{T|t}$, and the SOC controlled state $x_t$ pulled back onto the noise flow.}
  \label{fig:Denoising_processes_VE}
\end{figure*}

We visualize the sampling trajectories for selected examples across all tasks to better illustrate our method, presenting VP-SDE in \cref{fig:Denoising_processes_VP} and VE-SDE in \cref{fig:Denoising_processes_VE}. 

\begin{figure*}[t]
  \centering
  \begin{subfigure}{0.85\textwidth}
    \centering
    \includegraphics[width=\linewidth]{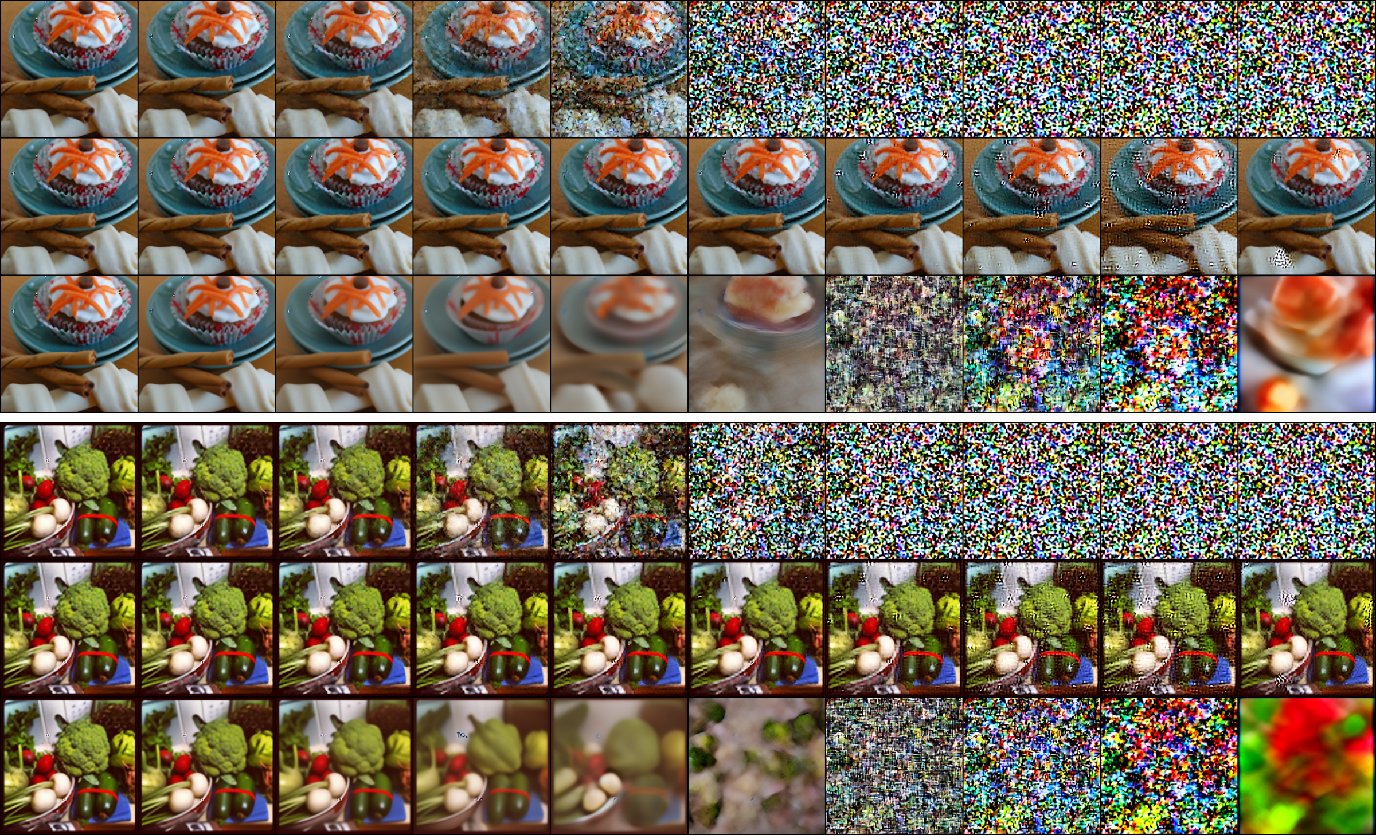}
    \caption{Denoising process for the Super resolution $8\times$ task with SD v1.5 at a resolution of 512.}
    \label{fig:sr8_progress_sd_imagenet}
  \end{subfigure}\par\medskip
  
  \begin{subfigure}{0.85\textwidth}
    \centering
    \includegraphics[width=\linewidth]{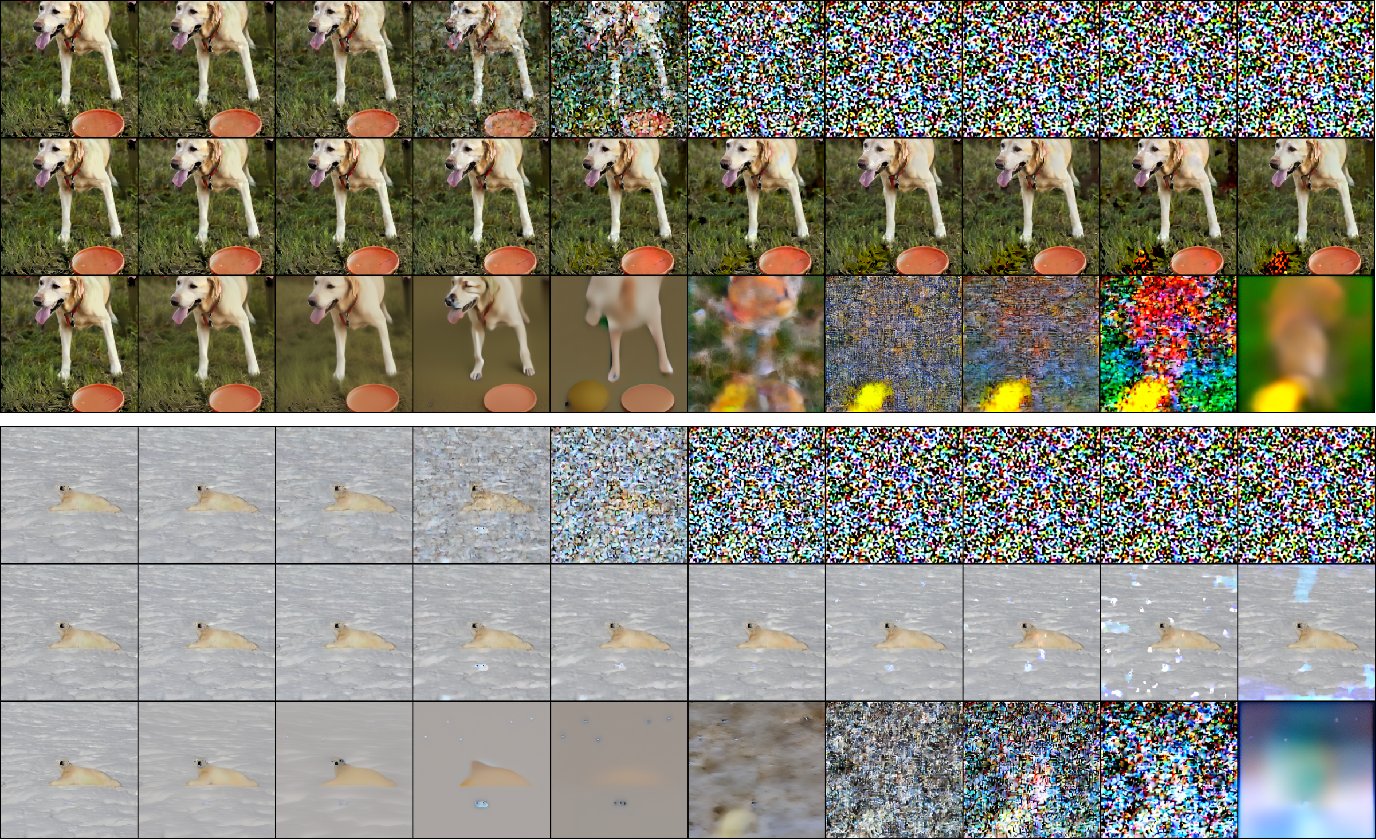}
    \caption{Denoising process for the High dynamic range task with SD v1.5 at a resolution of 512.}
    \label{fig:hdr_progress_sd_imagenet}
  \end{subfigure}

  \caption{\textbf{Denoising processes for SD v1.5 on natural images.} For each task, we display two complete recovery trajectories that start from pure noise. All denoising takes place in the latent space of the LDM. In each panel, from bottom to top, the rows show the few-step ODE sampler estimate $z_{T\mid t}$, the Langevin-refined $\hat{z}_{T\mid t}$, and the SOC-controlled state $z_t$ pulled back onto the noise flow.
}
  \label{fig:Denoising_processes_SDv1.5}
\end{figure*}

\begin{figure*}[t]
  \centering
  \begin{subfigure}{0.80\textwidth}
    \centering
    \includegraphics[width=\linewidth]{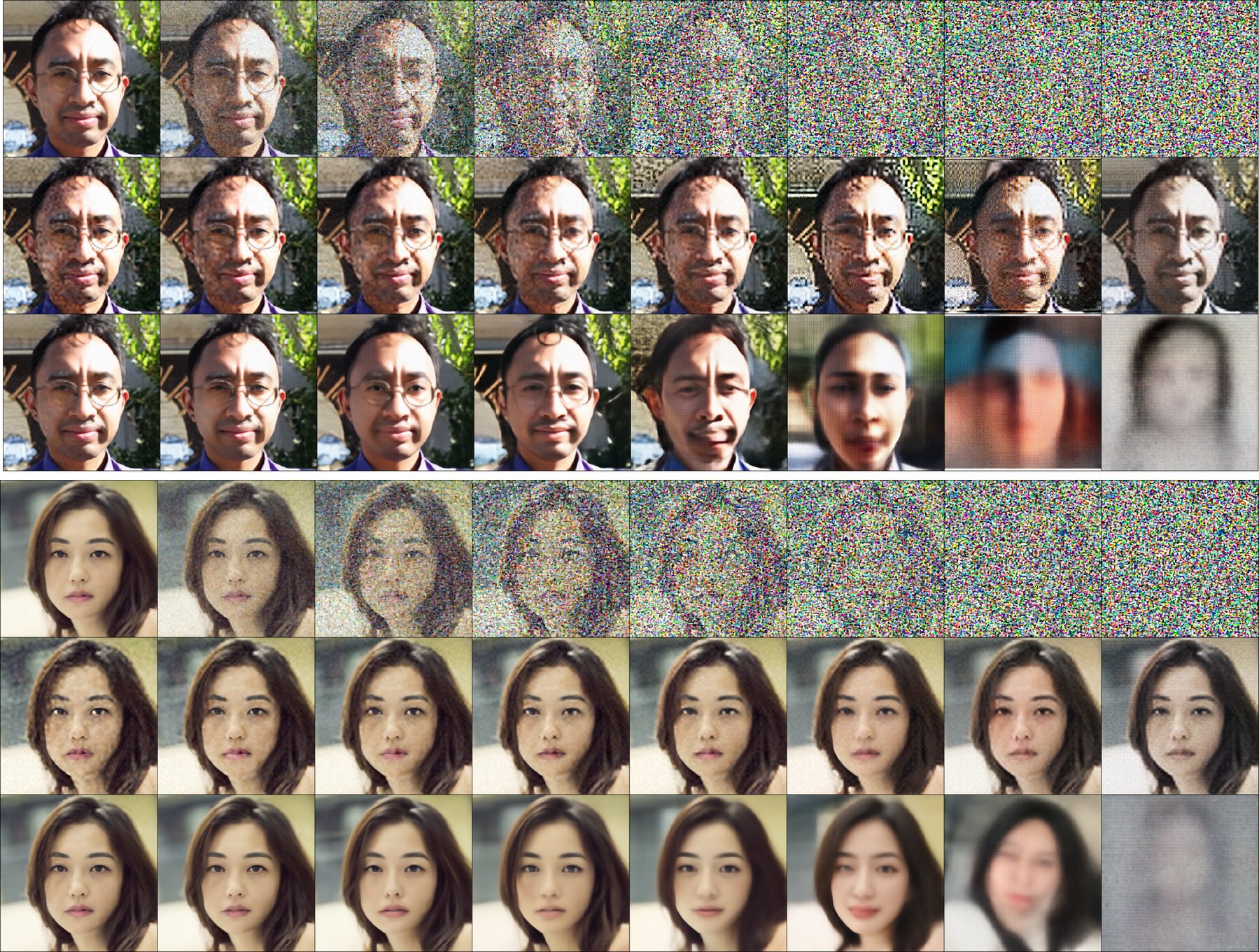}
    \caption{Denoising process for the bicubic super-resolution ($12\times$) task with SD3-medium at a resolution of 768.}
    \label{fig:sr_bicubic_x12_sd3_ffhq}
  \end{subfigure}\par\medskip
  
  \begin{subfigure}{0.80\textwidth}
    \centering
    \includegraphics[width=\linewidth]{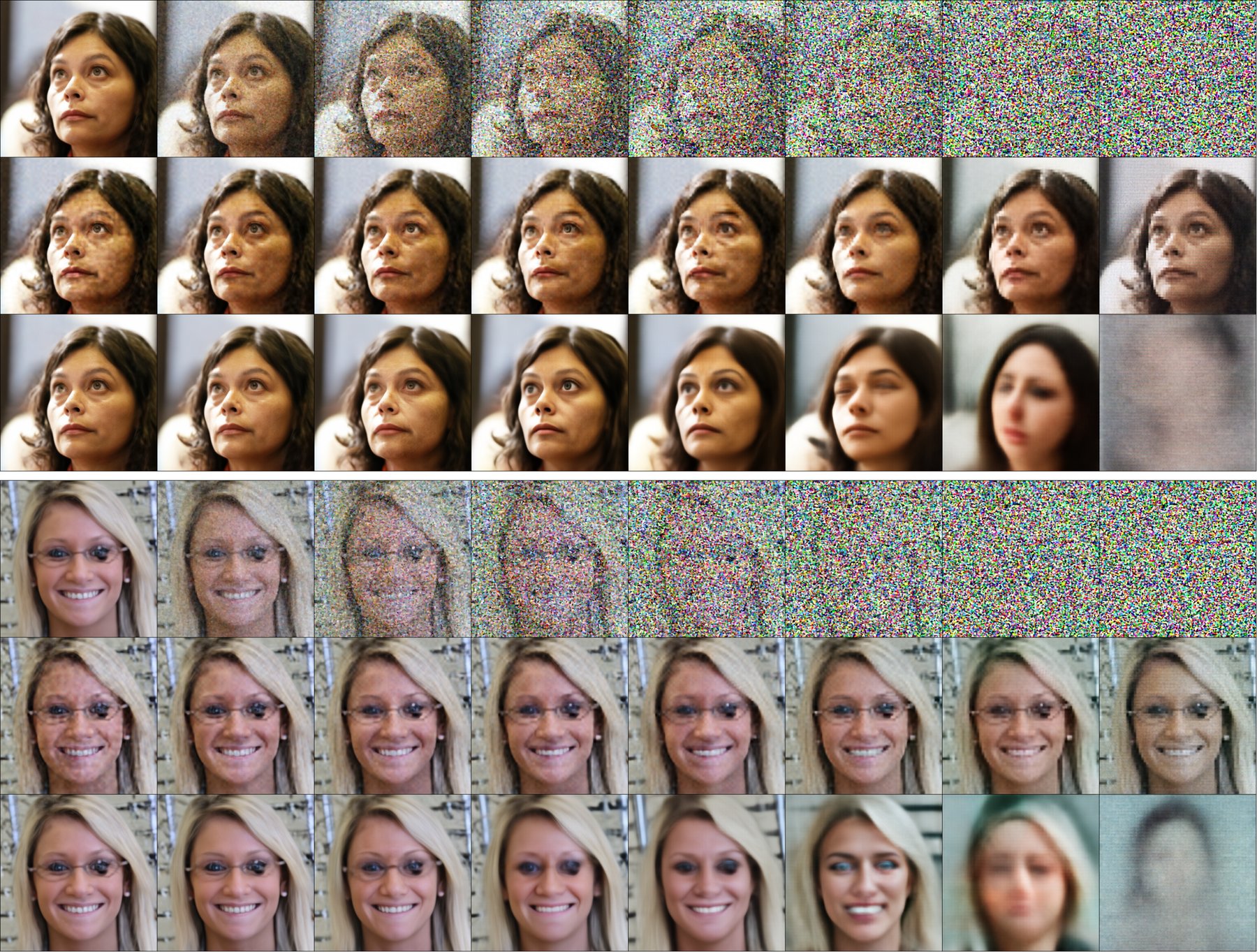}
    \caption{Denoising process for the avgpool super-resolution ($12\times$) task with SD3-medium at a resolution of 768.}
    \label{fig:sr_avgpool_x12_sd3_ffhq}
  \end{subfigure}

  \caption{\textbf{Denoising processes for SD3-medium on FFHQ images.} For each task, we display two complete recovery trajectories that start from pure noise. All denoising takes place in the latent space of the SD3. In each panel, from bottom to top, the rows show the few-step ODE sampler estimate $z_{T\mid t}$, the Langevin-refined $\hat{z}_{T\mid t}$, and the SOC-controlled state $z_t$ pulled back onto the noise flow.
}
  \label{fig:Denoising_processes_SD3}
\end{figure*}

We present additional results for SD v1.5 in \cref{fig:Denoising_processes_SDv1.5} and for SD3-medium in \cref{fig:Denoising_processes_SD3}. All operations are performed in the latent space, and the displayed images are obtained by decoding the intermediate variables with the decoder $\mathcal{D}$.

\begin{figure*}[t]
  \centering
  \begin{subfigure}{0.87\textwidth}
    \centering
    \includegraphics[width=\linewidth]{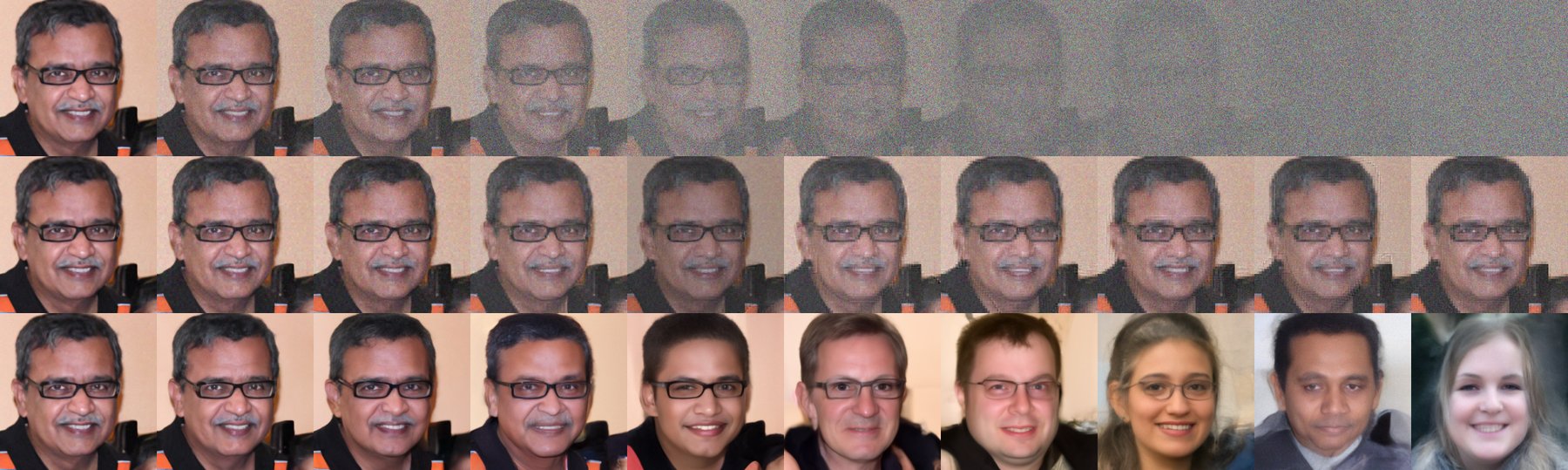}
    \caption{Denoising process for the super resolution task with VP-SDE.}
    \label{fig:sr_progress}
  \end{subfigure}

  \caption{\textbf{Denoising processes for all tasks with VP-SDE}. For each task, we display a complete recovery trajectory starting from pure noise. From bottom to top, the rows show the few step ODE sampler estimate $x_{T|t}$, the Langevin dynamics refined $\hat{x}_{T|t}$, and the SOC-controlled state $x_t$ pulled back onto the noise flow.}
\end{figure*}

\begin{figure*}[t]\ContinuedFloat
  \centering

  \begin{subfigure}{0.87\textwidth}
    \centering
    \includegraphics[width=\linewidth]{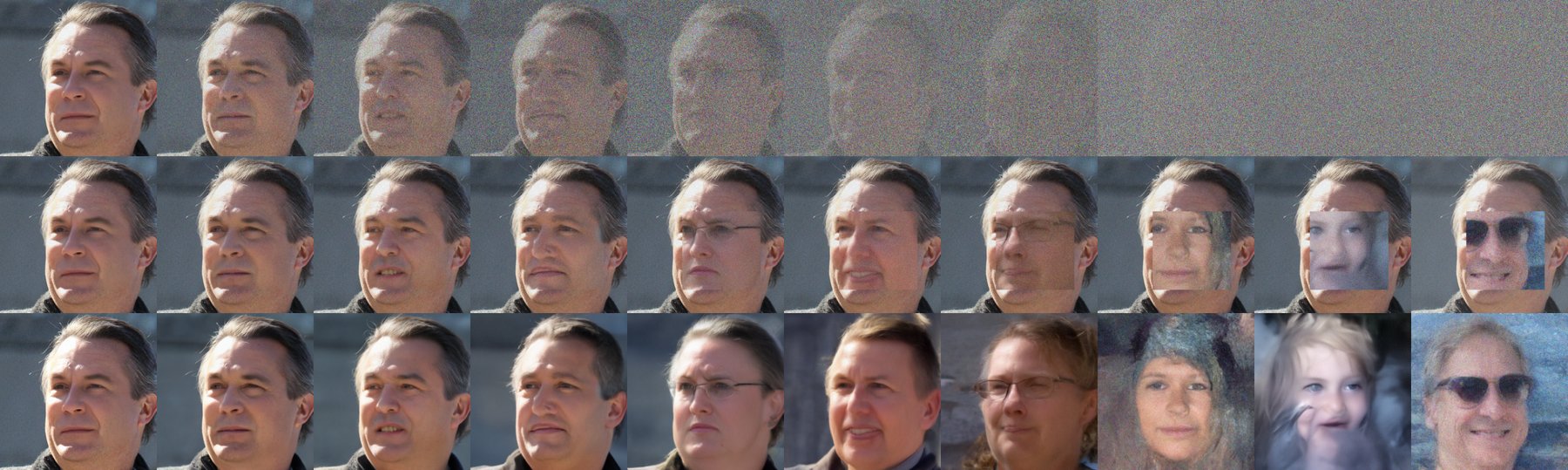}
    \caption{Denoising process for the Inpaint(Box) task with VP-SDE.}
    \label{fig:inpaintbox_progress}
  \end{subfigure}\par\medskip

  \begin{subfigure}{0.87\textwidth}
    \centering
    \includegraphics[width=\linewidth]{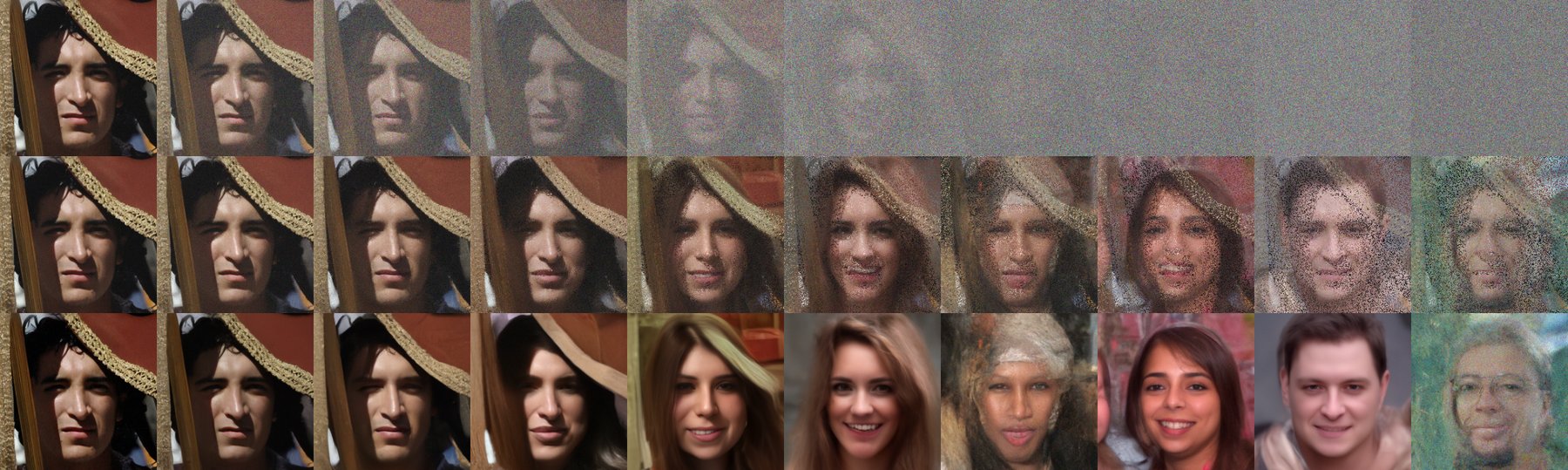}
    \caption{Denoising process for the Inpaint(Random) task with VP-SDE.}
    \label{fig:inpaintrandom_progress}
  \end{subfigure}\par\medskip

  \begin{subfigure}{0.87\textwidth}
    \centering
    \includegraphics[width=\linewidth]{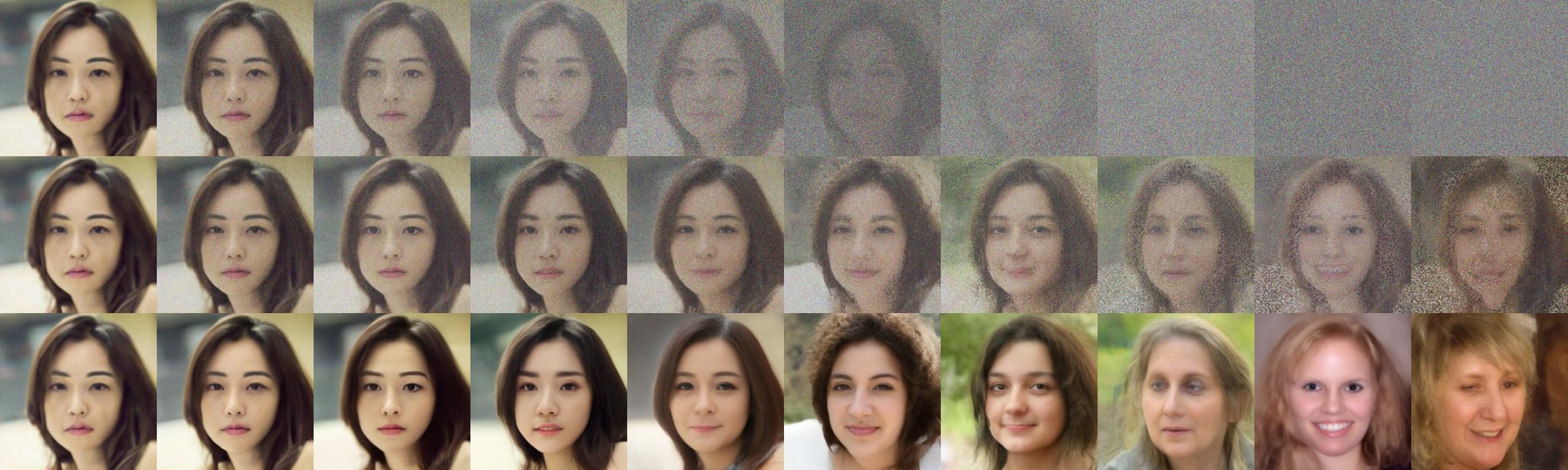}
    \caption{Denoising process for the Inpaint\&SR task with VP-SDE.}
    \label{fig:sr_inpaint_progress}
  \end{subfigure}

  \begin{subfigure}{0.87\textwidth}
    \centering
    \includegraphics[width=\linewidth]{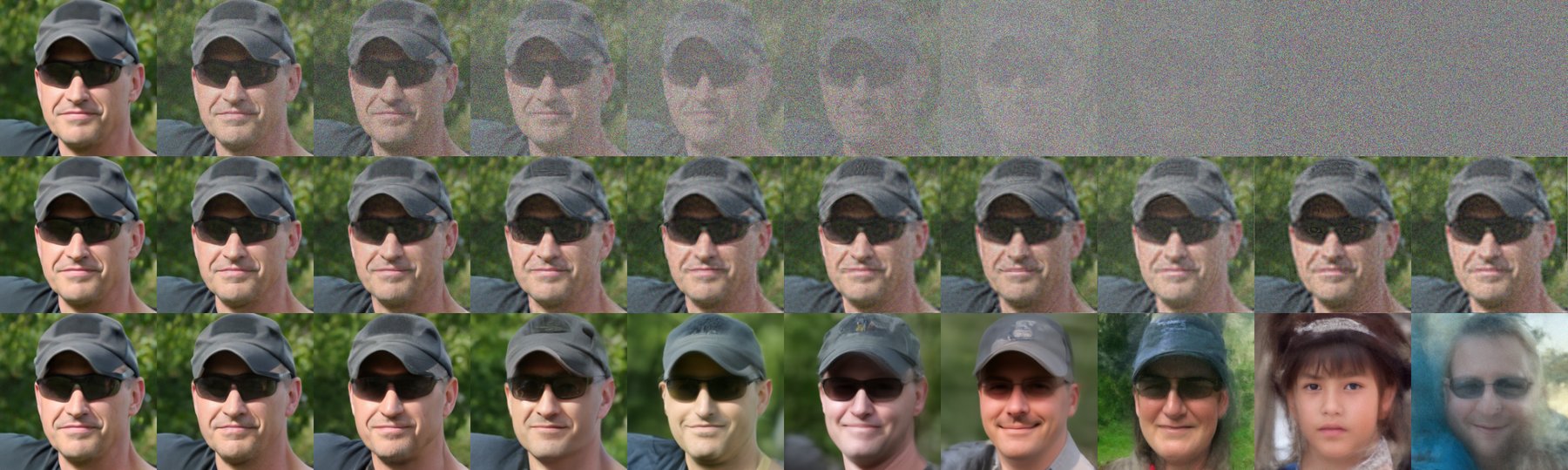}
    \caption{Denoising process for the Gaussian Blur task with VP-SDE.}
    \label{fig:gaussian_blur_progress}
  \end{subfigure}\par\medskip

  \caption{\textbf{Denoising processes for all tasks (continued)}.}
  \label{fig:Denoising_processes_VP}
\end{figure*}

\begin{figure*}[t]\ContinuedFloat
  \centering

  \begin{subfigure}{0.87\textwidth}
    \centering
    \includegraphics[width=\linewidth]{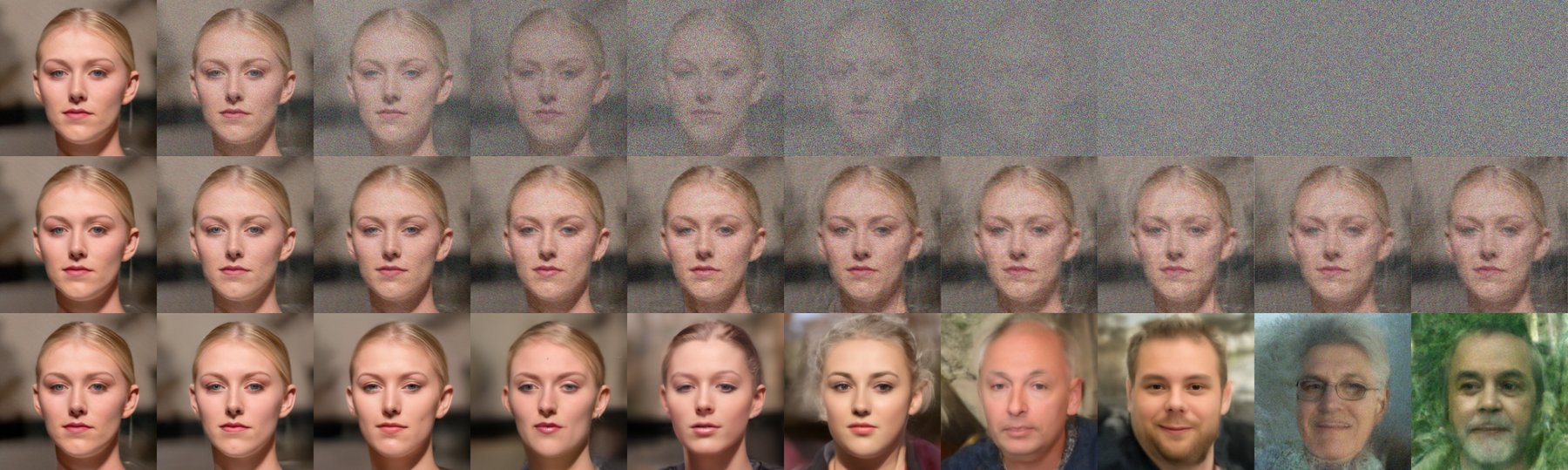}
    \caption{Denoising process for the Motion Blur task with VP-SDE.}
    \label{fig:motion_blur_progress}
  \end{subfigure}\par\medskip

  \begin{subfigure}{0.87\textwidth}
    \centering
    \includegraphics[width=\linewidth]{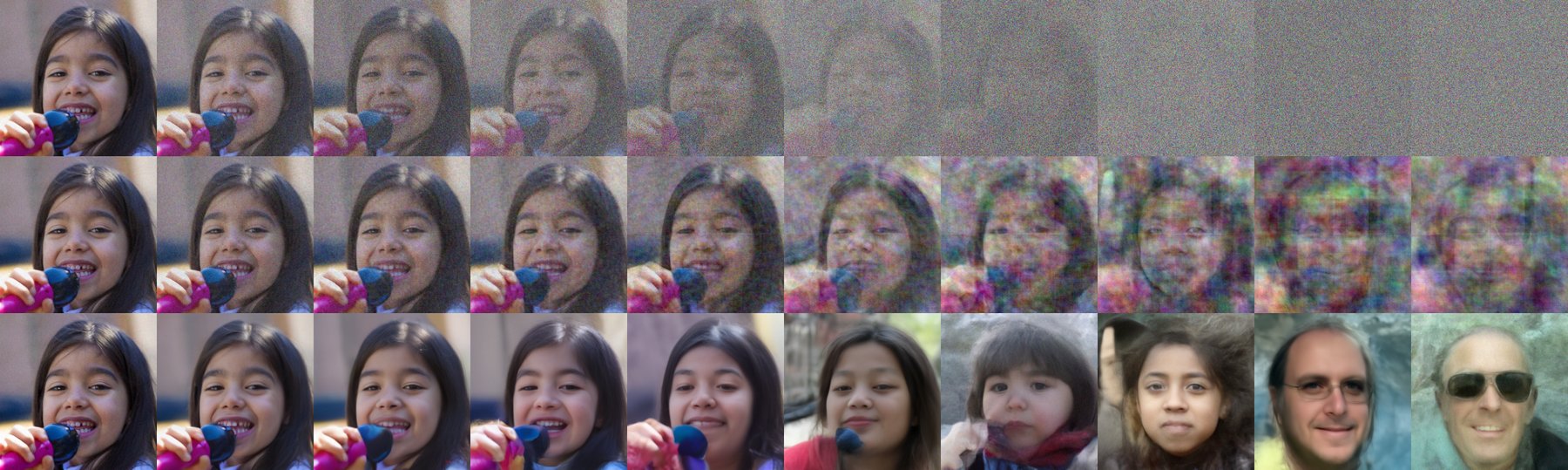}
    \caption{Denoising process for the Phase retrieval task with VP-SDE.}
    \label{fig:phaseretrieval_progress}
  \end{subfigure}\par\medskip

  \begin{subfigure}{0.87\textwidth}
    \centering
    \includegraphics[width=\linewidth]{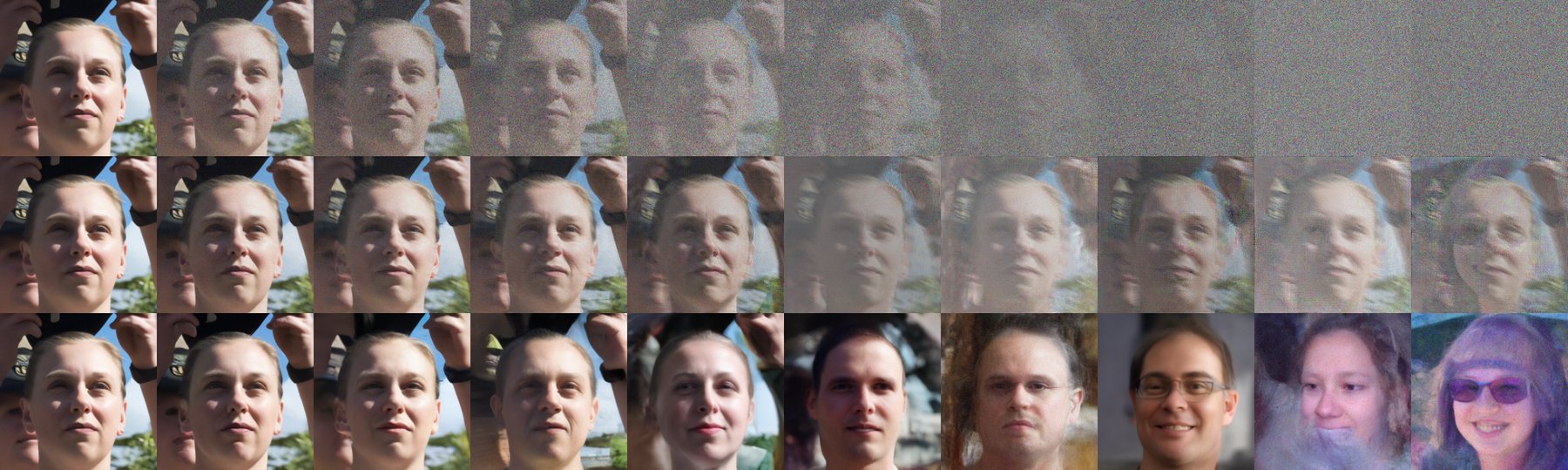}
    \caption{Denoising process for the Nonlinear Deblur task with VP-SDE.}
    \label{fig:Nonlinear_blur_progress}
  \end{subfigure}\par\medskip

  \begin{subfigure}{0.87\textwidth}
    \centering
    \includegraphics[width=\linewidth]{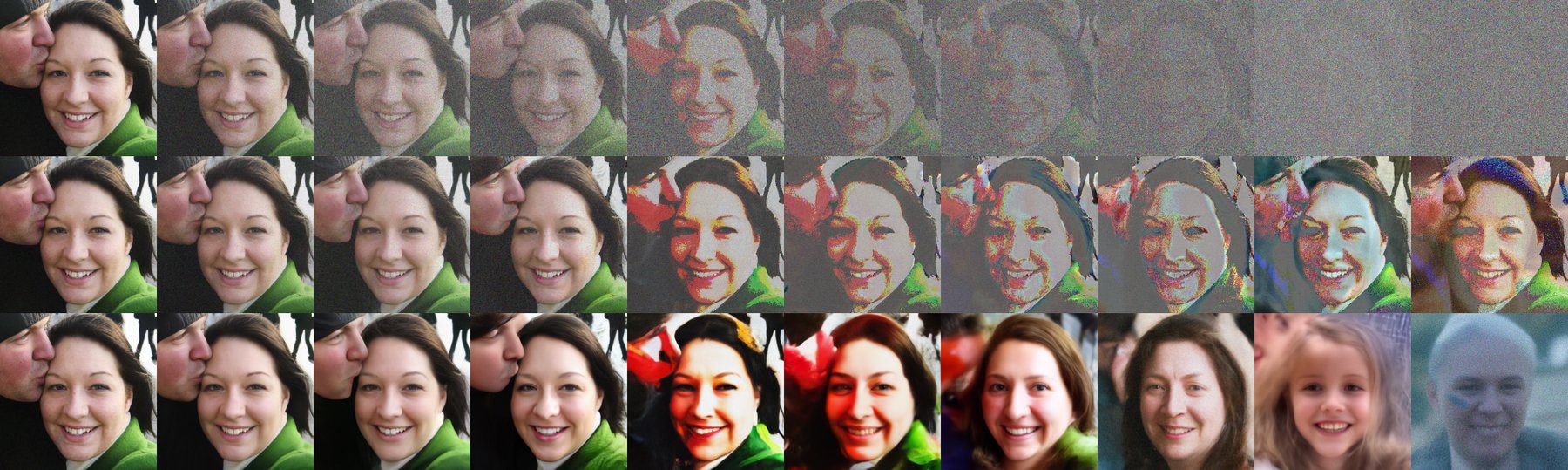}
    \caption{Denoising process for the High Dynamic Range task with VP-SDE.}
    \label{fig:hdr_progress}
  \end{subfigure}

  \caption{\textbf{Denoising processes for all tasks (continued)}.}
\end{figure*}

\end{document}